\PassOptionsToPackage{table,xcdraw,prologue}{xcolor}

\documentclass[sigconf, nonacm]{acmart}

\newcommand\vldbdoi{XX.XX/XXX.XX}
\newcommand\vldbpages{XXX-XXX}
\newcommand\vldbvolume{17}
\newcommand\vldbissue{5}
\newcommand\vldbyear{2024}
\newcommand\vldbauthors{\authors}
\newcommand\vldbtitle{\shorttitle} 
\newcommand\vldbavailabilityurl{https://github.com/khtae8250/Falcon}
\newcommand\vldbpagestyle{empty} 

\usepackage[linesnumbered,ruled,noend]{algorithm2e}

\newboolean{techreport}
\setboolean{techreport}{true}

\usepackage{graphicx}
\usepackage{balance}
\usepackage{bbm}
\usepackage{url}
\usepackage{amsmath}
\usepackage{caption}
\usepackage{subcaption}

\usepackage[table,xcdraw,prologue]{xcolor}
\usepackage{cleveref}
\usepackage{multirow}
\usepackage{xcolor}

\DeclareMathOperator*{\argmin}{arg\,min}
\DeclareMathOperator*{\argmax}{arg\,max}

\usepackage{comment}
\usepackage{enumitem}
\usepackage{xspace}

\usepackage{amsthm}
\usepackage{thmtools,thm-restate}

\usepackage[english]{babel}

\newcommand{\stitle}[1]{\vspace{1ex}\noindent{\bf #1}}

\newcommand{\squishlist}{ 
   \begin{list}{$\bullet$}
    { \setlength{\itemsep}{0pt}      \setlength{\parsep}{3pt} 
      \setlength{\topsep}{3pt}       \setlength{\partopsep}{0pt}
      \setlength{\leftmargin}{1.5em} \setlength{\labelwidth}{1em}
      \setlength{\labelsep}{0.5em} } }

\newcommand{\squishend}{
    \end{list}  } 

\newtheorem{definition}{Definition}
\newtheorem{example}{Example}

\AtBeginDocument{%
  \providecommand\BibTeX{{%
    \normalfont B\kern-0.5em{\scshape i\kern-0.25em b}\kern-0.8em\TeX}}}

\newcommand{\method}{\textsc{Falcon}}
\newcommand{\random}{{\em Random}}
\newcommand{\entropy}{{\em Entropy}}
\newcommand{\fal}{{\em FAL}}
\newcommand{\decouple}{$D$-$FA^{2}L$}

\newcommand{\rev}[1]{\textcolor{black}{#1}}

\begin{document}
\title{\method{}: Fair Active Learning using Multi-armed Bandits}

\author{Ki Hyun Tae}
\email{kihyun.tae@kaist.ac.kr}
\orcid{0000-0002-9307-7757}
\affiliation{%
  \institution{KAIST}
  \city{Daejeon}
  \country{Republic of Korea}
}

\author{Hantian Zhang}
\email{hantian.zhang@gatech.edu}
\orcid{0000-0002-9862-8773}
\affiliation{
\institution{Georgia Institute of Technology}
\city{Atlanta}
\country{USA}
}

\author{Jaeyoung Park}
\email{jypark@kaist.ac.kr}
\orcid{0009-0003-7583-0586}
\affiliation{%
  \institution{KAIST}
  \city{Daejeon}
  \country{Republic of Korea}
}

\author{Kexin Rong}
\email{krong@gatech.edu}
\orcid{0009-0007-3202-3767}
\affiliation{
\institution{Georgia Institute of Technology}
\city{Atlanta}
\country{USA}
}

\author{Steven Euijong Whang}
\authornote{Corresponding author}
\email{swhang@kaist.ac.kr}
\orcid{0000-0001-6419-931X}
\affiliation{%
  \institution{KAIST}
  \city{Daejeon}
  \country{Republic of Korea}
}


\begin{abstract}
Biased data can lead to unfair machine learning models, highlighting the importance of embedding fairness at the beginning of data analysis, particularly during dataset curation and labeling.
In response, we propose \method{}, a scalable fair active learning framework. \rev{\method{} adopts a data-centric approach that improves machine learning model fairness via strategic sample selection.}
Given a user-specified group fairness measure, \method{} identifies samples from ``target groups'' (e.g., \texttt{(attribute=female, label=positive)}) that are the most informative for improving fairness. However, a challenge arises since these target groups are defined using ground truth labels that are not available during sample selection. To handle this, we propose a novel trial-and-error method, where we postpone using a sample if the predicted label is different from the expected one and falls outside the target group. 
We also observe the trade-off that selecting more informative samples results in higher likelihood of postponing due to undesired label prediction, and the optimal balance varies per dataset. We capture the trade-off between informativeness and postpone rate as policies and propose to automatically select the best policy using adversarial multi-armed bandit methods, given their computational efficiency and theoretical guarantees. Experiments show that \method{} significantly outperforms existing fair active learning approaches in terms of fairness and accuracy and is more efficient. In particular, only \method{} supports a proper trade-off between accuracy and fairness where its maximum fairness score is 1.8--4.5x higher than the second-best results.


\end{abstract}

\maketitle

\pagestyle{\vldbpagestyle}
\begingroup\small\noindent\raggedright\textbf{PVLDB Reference Format:}\\
\vldbauthors. \vldbtitle. PVLDB, \vldbvolume(\vldbissue): \vldbpages, \vldbyear.\\
\href{https://doi.org/\vldbdoi}{doi:\vldbdoi}
\endgroup
\begingroup
\renewcommand\thefootnote{}\footnote{\noindent
This work is licensed under the Creative Commons BY-NC-ND 4.0 International License. Visit \url{https://creativecommons.org/licenses/by-nc-nd/4.0/} to view a copy of this license. For any use beyond those covered by this license, obtain permission by emailing \href{mailto:info@vldb.org}{info@vldb.org}. Copyright is held by the owner/author(s). Publication rights licensed to the VLDB Endowment. \\
\raggedright Proceedings of the VLDB Endowment, Vol. \vldbvolume, No. \vldbissue\ %
ISSN 2150-8097. \\
\href{https://doi.org/\vldbdoi}{doi:\vldbdoi} \\
}\addtocounter{footnote}{-1}\endgroup

\ifdefempty{\vldbavailabilityurl}{}{
\vspace{.3cm}
\begingroup\small\noindent\raggedright\textbf{PVLDB Artifact Availability:}\\
The source code, data, and/or other artifacts have been made available at \url{\vldbavailabilityurl}.
\endgroup
}

\section{Introduction}
\label{sec:introduction}

AI fairness is becoming essential as AI is widely used and needs to be trustworthy. Critical applications of fairness include AI-based hiring, AI-based judging, self-driving cars, and more. \rev{Recognizing that biased data is often the source of unfairness or discrimination in the downstream machine learning model, this work adopts a data-centric approach to improve fairness, as supported by previous studies~\cite{whangdata2023,DBLP:journals/jdiq/IlyasR22,DBLP:journals/vldb/LiuZR22,zhangiflipper2023}. Specifically, the data-centric approach mitigates unfairness in machine learning models by improving dataset curation and labeling, rather than improving model training.}

Manual data labeling in supervised learning is an expensive process, so active learning frameworks\,\cite{Lewis:1994:SAT:188490.188495,Seung:1992:QC:130385.130417,DBLP:series/synthesis/2012Settles,Abe:1998:QLS:645527.657478,Settles:2008:AAL:1613715.1613855,Settles:2007:MAL:2981562.2981724,Roy:2001:TOA:645530.655646,DBLP:conf/aaai/HsuL15} have been introduced to reduce the cost of data annotation. Traditional active learning focuses on selecting samples that lead to a maximal increase in model accuracy under a fixed labeling budget. This process of sample selection could worsen model fairness if not done carefully. For example, suppose there are two demographic groups men and women where it is desirable to have similar model accuracies for fairness, but women are currently under-represented. If no women samples are included during active learning, the accuracy disparities across groups might worsen as a result of data labeling. This scenario highlights the importance of incorporating fairness constraints during active learning to mitigate potential biases in the resulting model.

\begin{figure}[t]
\centering
  \includegraphics[width=\columnwidth]{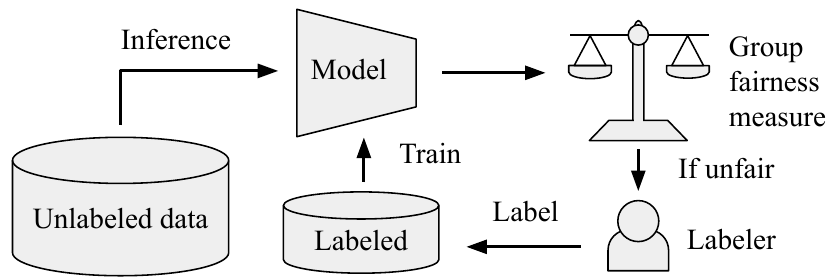}
  \vspace{-0.55cm}
  \caption{Fair active learning involves selecting samples that, when labeled, would enhance the fairness of a machine learning model according to a specific group fairness measure.}
  \label{fig:falcon}
  \vspace{-0.25cm}
\end{figure}

However, adding a fairness objective to the active learning framework is challenging for two reasons. First, while conventional fairness techniques can improve fairness by balancing training samples from different data subgroups\,\cite{DBLP:conf/sigmod/ZhangCAN21,DBLP:conf/iclr/Roh0WS21,DBLP:conf/nips/RohLWS21,DBLP:conf/cikm/IosifidisN19,DBLP:journals/kais/KamiranC11}, this labeled data is not available in the active learning setting. Hence, it is difficult to decide which samples are beneficial for fairness. Using pseudo-labels does not work well since active learning prioritizes uncertain samples, which tend to have inaccurate pseudo-labels. Second, as observed in prior works\,\cite{DBLP:conf/fat/MenonW18,DBLP:conf/nips/ChenJS18,DBLP:conf/nips/ZhaoG19,DBLP:conf/iclr/Roh0WS21}, the fairness objective is sometimes at odds with the accuracy objectives unless we are in ideal circumstances\,\cite{DBLP:conf/icml/DuttaWYC0V20}. For example, if fairness means having the same positive prediction rates among two ethnic groups (referred to as demographic parity\,\cite{DBLP:conf/kdd/FeldmanFMSV15}), then a perfectly-accurate classifier will not guarantee the same positive prediction rates if one ethnic group has positive samples more frequently than the other. \rev{The trade-off between accuracy and fairness is increasingly becoming a critical decision to make in practice. For example, an IT company recently scrapped its AI recruiting tool when it discriminated women~\cite{dastin2018amazon}. The problem was biased data where the majority of previous hires were men. A solution is to balance accuracy and fairness to reflect the increase of women in the workforce and ensure an unbiased evaluation, which may ultimately benefit the company.} 

In this work, we propose a fair active learning framework, \method{}, which accepts a group fairness measure and automatically learns policies for selecting samples that improve the fairness the most.  A user can first specify custom or well-known group fairness measures\,\cite{barocas-hardt-narayanan} including the prominent measures demographic parity\,\cite{DBLP:conf/kdd/FeldmanFMSV15}, equalized odds\,\cite{DBLP:conf/nips/HardtPNS16}, equal opportunity\,\cite{DBLP:conf/nips/HardtPNS16}, predictive parity\,\cite{DBLP:journals/bigdata/Chouldechova17}, and equalized error rate~\cite{DBLP:conf/sigmod/TaeW21}. Given the measure, \method{} identifies ``target groups'' defined using sensitive attribute values and labels. \method{} prioritizes getting labels for samples from the target groups to improve the trained model’s accuracy for that group, which in turn improves the overall fairness. As a running example, suppose that we use demographic parity as the fairness measure, where it is desirable to have similar positive prediction rates across demographic groups (e.g., men versus women). Suppose that the female group has a lower positive prediction rate at the moment. Note that the target group that requires more labels might change over the course of the training as we acquire more labels. \method{} would attempt to get more samples from the target group \texttt{(attribute=female, label=positive)} to improve the positive prediction rate. We can generalize this approach to find target groups for any group fairness measure specified by the user.


To address the first challenge of identifying target groups in fair active learning when ground truth labels are not available, \method{} proposes a {\em trial-and-error method} for handling unknown labels efficiently. Instead of solely relying on traditional active learning measures like entropy or confidence that determine a sample's informativeness by its proximity to the decision boundary, \method{} adds a fairness objective to select samples from specific groups (e.g., the \texttt{(attribute=female, label=positive)} group). However without ground truth labels, we can not always identify samples in the target group. Our key observation is that adding more labels does not necessarily lead to improved fairness, and it is better to delay using samples that would negatively impact fairness. We thus use a trial-and-error approach where we select samples in a sensitive group to label, but delay using them in model training if they have undesired labels for that group. For example, \method{} selects an informative sample in the female group to label, but only includes it in model training if it has a positive label since adding negative labels in the female group worsens demographic parity. 



In order to find the most informative samples for fairness, we introduce a policy selection framework on top of the trial-and-error method to optimize model fairness. Specifically, we observe that the more informative samples are, the more likely they are postponed due to undesired label predictions. We capture how aggressively we should select informative samples using a {\em policy}, and our solution is to learn the optimal policy using a multi-armed bandit (MAB) approach based on the rewards in terms of improved fairness. However, the policies are not independent of each other, and the rewards also vary as we label more data. As a result, the ideal policy depends on the dataset and the stage of the labeling process. In order to handle this complicated scenario, \method{} uses adversarial MAB methods to dynamically select the best sampling policy. Adversarial MABs provide a principled approach to selecting amongst competing strategies that share a limited set of resources and have strong theoretical bounds of regret. To further improve stability and performance, \method{} rewards the nearest policies to the chosen one, ensuring that the unknown best policy still receives rewards, even if it is not directly selected.


To address the second challenge of balancing accuracy and fairness objectives, we extend \method{} to improve accuracy by combining it with traditional active learning techniques where fair and accurate labeling are alternated probabilistically. Our approach does not require any modifications for other active learning methods and effectively controls an accuracy-fairness trade-off, as we show in the experiments. When there is no ambiguity, we refer to the combined version as \method{} as well.




In our experiments, we show that \method{} significantly outperforms various fair active learning baselines on real datasets in terms of model fairness and accuracy and is faster.


We summarize our contributions as follows:
\begin{itemize}[leftmargin=*]
\item We propose \method{}, a fair active learning framework that selects samples to improve fairness and accuracy. \method{} (1) uses a novel labeling strategy where it first selects subgroups to label and handles unknown ground truth labels using a trial-and-error strategy; (2) automatically selects the best sampling policy using adversarial MABs; and (3) balances fairness and accuracy by alternating its selection with traditional active learning.
\item \method{} is efficient by using MABs and requires much fewer model trainings than other fair active learning approaches.
\item We empirically show that \method{} drastically outperforms fair active learning baselines w.r.t.\@ fairness and accuracy and is faster.
\end{itemize}

\section{Background and Overview}
\label{sec:problem}


We focus on an active learning scenario where the labeling budget is limited, and we would like to improve both fairness and accuracy. We target any application where there are not enough labels and discrimination is a potential problem. In the following sections, we explain preliminaries (Section~\ref{sec:preliminaries}), define our problem (Section~\ref{sec:declarativelabeling}), and provide an overview of \method{} (Section~\ref{sec:falcon}). We focus on improving fairness for now and later discuss how to also improve accuracy in Section~\ref{sec:algorithm}. 


\subsection{Preliminaries}
\label{sec:preliminaries}
In this work, we focus on a binary classification setting and assume a training dataset $D_{train}=\{x_i, z_i, y_i\}_{i=1}^n$ where $x_i$ is a training sample, $z_i \in \mathbb{Z}$ is a sensitive attribute (e.g., gender), and $y_i$ is its label having a value of 0 or 1. We also denote the unlabeled, validation, and test datasets as $D_{un}$, $D_{val}$, and $D_{test}$, respectively. A classifier $h$ is trained on $D_{train}$, and its prediction on a test sample is $\hat{y}\in\{0,1\}$. 

\paragraph{Group fairness definitions}

To illustrate the fairness issues we aim to address, we begin by defining group fairness. Group fairness ensures that a trained model $h$ has equal or similar statistics across different sensitive groups. Here we list five representative fairness measures~\cite{10.1145/3194770.3194776} as follows: 
\begin{itemize}[leftmargin=*]
\item Demographic Parity (DP)~\cite{DBLP:conf/kdd/FeldmanFMSV15} is satisfied if a trained model has an equal positive prediction rate across sensitive groups.
\begin{equation} \label{eq:dp}
\forall z_i, z_j \in \mathbb{Z}, p(\hat{y}=1|z_i) \simeq p(\hat{y}=1|z_j)
\end{equation}

\item Equalized Odds (ED)~\cite{DBLP:conf/nips/HardtPNS16} is satisfied if a trained model has an equal accuracy across sensitive groups conditioned on the true label $\text{y} \in \{0, 1\}$, i.e., having equal false positive rate (FPR) and false negative rate (FNR). 
\begin{equation} \label{eq:ed}
\forall z_i, z_j \in \mathbb{Z}, \text{y} \in \{0, 1\}, p(\hat{y}=1|y=\text{y}, z_i) \simeq p(\hat{y}=1|y=\text{y}, z_j)
\end{equation}

\item Equal Opportunity (EO)~\cite{DBLP:conf/nips/HardtPNS16} is a relaxed version of ED that only consider conditioning on $y=1$, i.e., having equal FNR.
\begin{equation} \label{eq:eo}
\forall z_i, z_j \in \mathbb{Z}, p(\hat{y}=1|y=1, z_i) \simeq p(\hat{y}=1|y=1, z_j)
\end{equation}


\item Predictive Parity (PP)~\cite{DBLP:journals/bigdata/Chouldechova17} is satisfied if a trained model has an equal probability of having positive labels across sensitive groups conditioned on the model prediction $\hat{\text{y}} \in \{0, 1\}$, i.e., having equal false omission rate (FOR) and false discovery rate (FDR).
\begin{equation} \label{eq:pp}
\forall z_i, z_j \in \mathbb{Z}, \hat{\text{y}} \in \{0, 1\}, p(y=1|\hat{y}=\hat{\text{y}}, z_i) \simeq p(y=1|\hat{y}=\hat{\text{y}}, z_j)
\end{equation}


\item Equalized Error Rate (EER)~\cite{DBLP:conf/sigmod/TaeW21} is satisfied if a trained model has an equal classification error rate across sensitive groups.
\begin{equation} \label{eq:eer}
\forall z_i, z_j \in \mathbb{Z}, p(\hat{y}\ne y|z_i) \simeq p(\hat{y}\ne y|z_j)
\end{equation}

\end{itemize}

To evaluate the fairness of the trained model $h$, we define a fairness score as one minus the maximum fairness disparity~\cite{bird2020fairlearn} between any two sensitive groups on the unseen test set. In the extreme case, a fairness score of 1 indicates that the classifier is perfectly fair according to the given fairness measure.


\paragraph{Active learning}
The goal of active learning (AL) is to minimally label samples while maximizing model accuracy. A standard approach in AL is to choose samples that have the lowest confidence or highest entropy. Intuitively, we would like the labeler to label the most challenging samples. Within AL research, there are several approaches on how to evaluate each sample: uncertainty-based~\cite{margin2006roth, confidence2014wang, entropy2001shannon}, diversity-based~\cite{10.1145/1015330.1015349, sener2018active}, and hybrid approaches~\cite{6751135, ash2020deep}. In comparison, fair active learning adds one more dimension of difficulty where we would like to also select samples that improve fairness. As a default, we assume batch active learning where a set of samples are selected for labeling.

\paragraph{Informativeness for Fairness}
In traditional AL, one representative approach is to consider uncertain samples as informative, and choosing such samples improves overall accuracy the most. Informativeness is estimated using entropy or confidence. In our work, we define an analogous notion of {\em informativeness for fairness}. It is well-known that prominent group fairness measures can be expressed as a sum of the subgroup accuracies, where a subgroup is defined using a label and a sensitive attribute~\cite{DBLP:conf/iclr/Roh0WS21, DBLP:conf/sigmod/ZhangCAN21} like \texttt{(attribute=female, label=positive)} (details are in Section~\ref{sec:subgrouplabeling}). Based on this key insight, we define a sample to be informative for fairness if it can improve the accuracy of specific subgroups. However, estimating this information score accurately is challenging due to the lack of labels in unlabeled data. To address this problem, we propose a novel data-driven technique for identifying the most informative samples for fairness in Section~\ref{sec:informativesamples}.


\subsection{Problem Definition}
\label{sec:declarativelabeling}

Our goal is to select samples to label for the purpose of improving fairness of a trained model. Given an unlabeled dataset $D_{un}$, a train dataset $D_{train}$, a validation set $D_{val}$, a loss function $l_\theta$, a group fairness measure $F(\theta, D)$ that measures fairness when using the model parameters $\theta$ on the dataset $D$, a labeling process $H$ that receives unlabeled data and returns labeled data (e.g., human labeling), and a labeling budget $b$ at every round with a total budget of $B$, {\em fair active learning} (fair AL) solves the following optimization problem at each round of labeling: 
\begin{align*}
\argmax_{S \subseteq D_{un}} \quad & F(\argmin_{\theta} \ l_\theta(D_{train} \cup H(S)), D_{val}) \\
\text{s.t. } \quad & |S| \leq b.
\end{align*}

Here, in each step, we want to find a set $S$ with at most $b$ samples, such that, after labeling $S$ and adding them to the current training data, the model trained on this training data would achieve the highest fairness on the validation set.

In Section~\ref{sec:algorithm}, we describe how to extend the above problem formulation to jointly optimize accuracy and fairness by combining our approach with traditional AL methods.



\subsection{\method{} Overview}
\label{sec:falcon}

\begin{figure}[t]
\centering
  \includegraphics[width=\columnwidth]{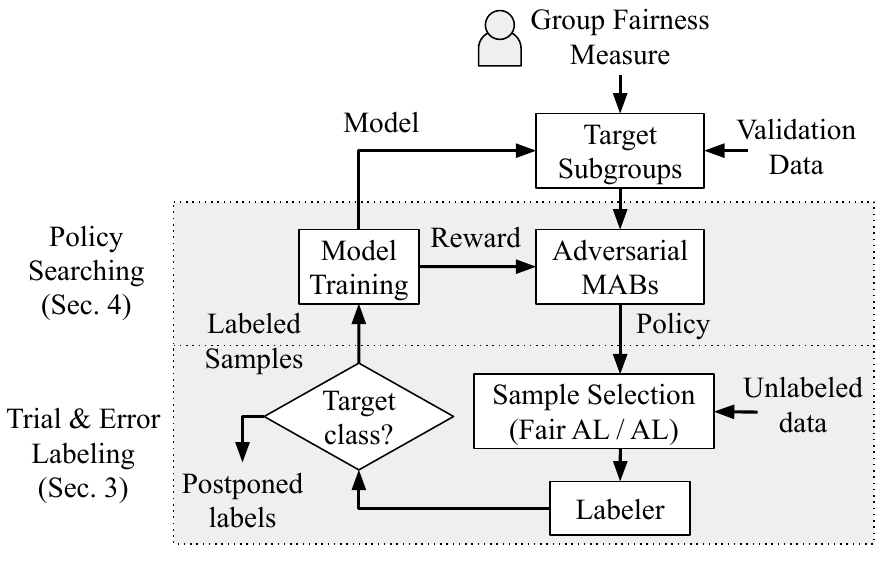}
  \vspace{-0.9cm}
  \caption{Overview of \method{} workflow.}
  \label{fig:overview}
  \vspace{-0.3cm}
\end{figure}

Figure~\ref{fig:overview} gives an overview of the \method{} framework. As an input, the user provides a group fairness measure. Then \method{} determines which ``target'' groups of samples need to be labeled first. This approach is more general than class imbalance works with fixed minority groups where the minority groups themselves may become majority groups as samples are labeled for the minority group. When selecting samples for a specific group, the selection itself may be undesired due to the lack of labels, so \method{} learns the right policy to select samples that are informative and yet are not likely to end up having undesired labels. Finally, we extend \method{} with traditional AL methods to improve model accuracy. When there is no ambiguity, we refer to the extended version as \method{} as well.

In the following sections, we introduce an effective trial-and-error labeling strategy (Section~\ref{sec:trialanderror}), propose automatic policy searching methods (Section~\ref{sec:policysearching}), and present \method{}'s algorithm (Section~\ref{sec:algorithm}).

\section{Trial-and–error Labeling Strategy}
\label{sec:trialanderror}
We discuss \method{}'s labeling strategy to improve fairness where it first selects subgroups to label (Section~\ref{sec:subgrouplabeling}) and handles unknown ground truth labels using a trial-and-error approach (Section~\ref{sec:handlingunknown}).

\subsection{Subgroup Labeling to Improve Fairness}
\label{sec:subgrouplabeling}

Our key strategy for improving fairness is to increase the labeling of specific subgroups of the data. We take inspiration from the minibatch selection approaches OmniFair\,\cite{DBLP:conf/sigmod/ZhangCAN21} and FairBatch\,\cite{DBLP:conf/iclr/Roh0WS21}, which adjust subgroup sampling ratios within a minibatch to improve fairness.

For illustration purposes, we assume two sensitive groups (i.e., $\mathbb{Z} = \{0, 1\}$) for now, and discuss extensions to multiple sensitive groups later. Suppose the fairness metric is EO, which requires the trained model to have similar accuracies on two sensitive groups for positive samples. Suppose that one of the groups currently has a lower accuracy than the other and is thus underrepresented. Then providing more samples to the underrepresented group can enhance the model's performance on that group because the model assigns more weight to the underrepresented group samples when optimizing its objective function, naturally boosting accuracy.

Using an analysis in FairBatch\,\cite{DBLP:conf/iclr/Roh0WS21}, optimizing for EO can be formulated as a quasi-convex optimization problem, which intuitively means that increasing the underrepresented group's samples first improves EO and then does not from some point. This setup justifies the approach of increasing the labeling of a subgroup if it is underrepresented. 

The subgroup decomposition analysis can extend to other popular group fairness metrics beyond EO. OmniFair~\cite{DBLP:conf/sigmod/ZhangCAN21} proposed a similar reweighting solution to improve group fairness as well. Based on this key observation, we can identify the target subgroups that need to be labeled in order to improve target fairness. We explain the subgroups to target when the fairness measures are DP and EO. The analysis for other group fairness metrics can be found in our technical report~\cite{techreport}.



\begin{example} \label{ex:dpgroup} {\bf Target Subgroups for DP.}
For DP (Equation~\ref{eq:dp}), let us assume that $p(\hat{y} = 1|z=0) < p(\hat{y}=1|z=1)$ without loss of generality. We also know that 
\begin{align*}
p(\hat{y} = 1) &= p(y=1, \hat{y}=1) + p(y=0, \hat{y}=1)\\
&=p(y=1)p(\hat{y}=1|y=1) + p(y=0)p(\hat{y}=1|y=0)\\ &=p(y=1)p(\hat{y}=1|y=1) + p(y=0)(1-p(\hat{y}=0|y=0)).
\end{align*}

Then $p(\hat{y} = 1|z=0) < p(\hat{y}=1|z=1)$ can be rewritten as 
\begin{align*}
& p(y=1|z=0)\underline{p(\hat{y}=1|y=1,z=0)} + \\
& p(y=0|z=0)(1-p(\hat{y}=0|y=0,z=0)) < \\ 
& p(y=1|z=0)p(\hat{y}=1|y=1,z=1) + \\
& p(y=0|z=1)(1-\underline{p(\hat{y}=0|y=0,z=1)}).
\end{align*}

Hence, we can see that improving $p(\hat{y}=1|y=1,z=0)$ results in increasing $p(\hat{y}=1|z=0)$, while improving $p(\hat{y}=0|y=0,z=1)$ results in decreasing $p(\hat{y}=1|z=1)$. Both strategies can reduce the disparity, so $(y=1, z=0)$ and $(y=0, z=1)$ subgroups should be targeted for additional labeling to improve the DP score.
\end{example}

\begin{example} \label{ex:edgroup} {\bf Target Subgroups for EO.}
For EO (Equation~\ref{eq:eo}), the goal is to close the gap between $p(\hat{y}=1|y=1,z=0)$ and $p(\hat{y}=1|y=1,z=1)$. Let us assume that the first term is smaller. Then, improving $p(\hat{y}=1|y=1,z=0)$ directly improves fairness. 
\end{example}

Table~\ref{tbl:targetgroups} provides a summary of the target subgroups for each fairness measure when a sensitive group $z^* \in \{0, 1\}$ has a lower fairness value. For example, for DP, the target groups are $(1, z^*)$ or $(0, 1-z^*)$ when $p(\hat{y} = 1|z=z^*) < p(\hat{y}=1|z=1-z^*)$. For ED and PP, the target groups are determined by the subgroups that have a larger disparity gap. 

\begin{table}[t]
\small
  \centering
  \begin{tabular}{ccc}
  \toprule
  {\bf Metric} & {\bf Target Subgroups} $\mathbf{(y, z)}$ & \\
  \midrule
  DP & $(1, z^*)$ or $(0, 1-z^*)$ \\
  \midrule
  EO & $(1, z^*)$ \\
  \midrule
  \multirow{2}{*}{ED} & $(0, 1-z^*)$, if FPR gap $\geq$ FNR gap\\
  & $(1, z^*)$, otherwise \\
  \midrule
  \multirow{2}{*}{PP} & $(0, 1-z^*)$ or $(1, 1-z^*)$, if FOR gap $\geq$ FDR gap \\
  & $(0, z^*)$ or $(1, z^*)$, otherwise\\
  \midrule
  EER & $(0, 1-z^*)$ or $(1, 1-z^*)$ \\
  \toprule
  \end{tabular}
  \caption{Target subgroups for each group fairness measure when a sensitive group $z^* \in \{0, 1\}$ has a lower fairness value.
  }
  \label{tbl:targetgroups}
  \vspace{-0.7cm}
\end{table}

\paragraph{Other Fairness Measures} 
In addition to the group fairness measures in Table~\ref{tbl:targetgroups}, \method{} can support any group fairness measure that can be expressed as the subgroup accuracies. Similar to the previous examples, one can identify the target groups that have low fairness values and then perform more labeling on those groups.


\paragraph{Dynamic Target Group Selection} 
As we continue labeling data, the groups requiring more labels might change. This is a key difference from existing works that only focus on solving class imbalance, where the underrepresented group is fixed and contains fewer samples. Another complication arises from the inter-group influence, where labeling certain groups might positively or negatively impact the accuracy of other groups. The direction of this influence depends on the data; similar groups might positively affect each other, while different ones may have a negative impact. Although we don't model this influence directly, it justifies our strategy of {\em dynamically} selecting the appropriate groups to improve.

A key challenge is that ground truth labels are not available in an AL setting, making it difficult to determine whether an unlabeled sample belongs to the target groups. In the next section, we propose a simple yet effective solution, which explicitly labels samples and then uses only those with the label of interest. 


\subsection{Handling Unknown Ground Truth}
\label{sec:handlingunknown}

The problem now shifts to selecting samples from the target groups when the labels are unavailable. A na\"ive approach for handling this issue is to generate pseudo labels~\cite{Lee2013PseudoLabelT} using model predictions and prioritize samples with higher informativeness. If the pseudo labels are perfect, then there would be no need to be concerned about using those samples at all. However, there is a fundamental limit to their correctness if they are informative for the trained model. Indeed, only samples with high confidence are likely to obtain correct pseudo labels.


Our solution is to select samples that are likely to be informative and label them, but postpone using them for training when they turn out to have undesirable labels. We refer to this strategy as trial-and-error labeling. This approach may sound counter-intuitive at first, given labeling is an expensive process. However, using samples with undesired labels can negatively affect fairness, so it is better not to use samples with undesired labels immediately. Later on, if we actually need these labels to further improve fairness or accuracy, we can use them. Here we show how trial-and-error labeling improves fairness when using DP and EO.

\begin{example} {\bf Trial-and-error labeling for DP.} Continuing from Example~\ref{ex:dpgroup} where $p(\hat{y} = 1|z=0) < p(\hat{y}=1|z=1)$, we should obtain samples from $(y=1, z=0)$ or $(y=0, z=1)$ to improve DP. However, some samples can potentially be acquired from $(\underline{y=0}, z=0)$ or $(\underline{y=1}, z=1)$ with undesired labels. These samples directly worsen the DP gap by decreasing $p(\hat{y} = 1|z=0)$ or increasing $p(\hat{y}=1|z=1)$. Hence, it is important to postpone using them to improve fairness.
\end{example}

\begin{example} {\bf Trial-and-error labeling for EO.} Continuing from Example~\ref{ex:edgroup}, suppose the targeted group is set to $(y=1, z=0)$ to address the EO disparity, assuming that $p(\hat{y}=1|y=1,z=0)$ < $p(\hat{y}=1|y=1,z=1)$. In this case, we can possibly obtain data from the $(\underline{y=0}, z=0)$ subgroup, which improves $p(\hat{y}=0|y=0, z=0)$. Although training a model on these samples does not directly decrease $p(\hat{y}=1|y=1,z=0)$, more training samples from $(y=1, z=0)$ can lead to overfitting the model's predictions for the $(z=0)$ data towards $\hat{y}=0$. As a result, there can be an ``indirect'' decrease in $p(\hat{y}=1|y=1,z=0)$ in the end. Hence, it is also better to delay the usage of these samples to improve fairness.
\end{example}

\paragraph{Informativeness for Fairness} \label{sec:informativesamples}
The remaining question is which sample is the most ``informative for fairness’’ when utilizing trial-and-error labeling. That is, the target sample should improve the specific subgroup's accuracy the most (and thus improve the overall fairness the most) and also have the label of interest. 

Informativeness can be measured in several ways, and we first propose a simple solution based on an information theoretic approach using Shannon’s entropy~\cite{entropy2001shannon}. The information obtained by a sample labeled as $x$, $I_{entropy}(x)$, can be expressed as follows:
\begin{equation*} \label{eq:entropy}
I_{entropy}(x) = p(+|x) \log \frac{1}{p(+|x)} + (1 - p(+|x)) \log \frac{1}{1-p(+|x)}
\end{equation*}
where $p(+|x) = p(\hat{y}=1|x)$ is the predicted probability of the sample $x$ being labeled as 1 by the trained model. The first term means the expected information score when $x$ is labeled 1, and the second term means the expected information score when the label is 0. $I_{entropy}$ is maximized when $p(+|x) = 0.5$.

Entropy is typically used to identify a sample that can improve overall accuracy, but it can also be adapted to improve fairness using the trial-and-error labeling strategy. Consider the example where the target subgroup is \texttt{(attribute=female, label=positive)}. In this case, we first select a sample that has the closest $p(+|x)$ value to $0.5$ from the \texttt{(attribute=female)} group. We then include this sample in the training set only if its true label is positive. If there are multiple target groups like DP, we randomly select one of these groups for each iteration and apply the same approach. In Section~\ref{sec:policiesforsampleselection}, we generalize the notion of selecting samples with a desired label value as a {\em policy} to capture how much ``risk'' we are willing to make for finding samples with desired labels.

\begin{figure}[t]
\centering
  \begin{subfigure}{0.49\columnwidth}
     \includegraphics[width=\columnwidth]{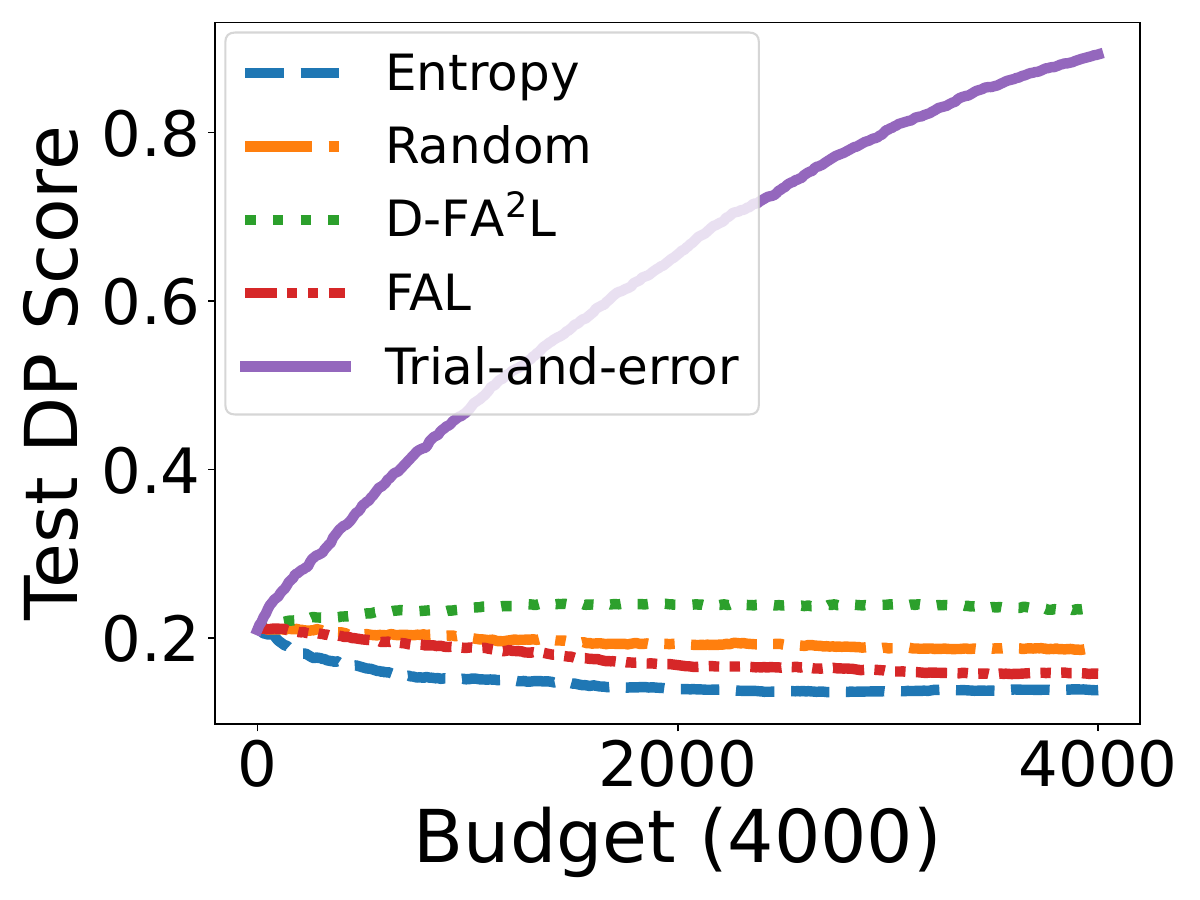}
     \vspace{-0.3cm}
     \caption{{\sf TravelTime (DP)}}
     \label{fig:simpletravel}
  \end{subfigure}
  \centering
    \begin{subfigure}{0.49\columnwidth}
     \includegraphics[width=\columnwidth]{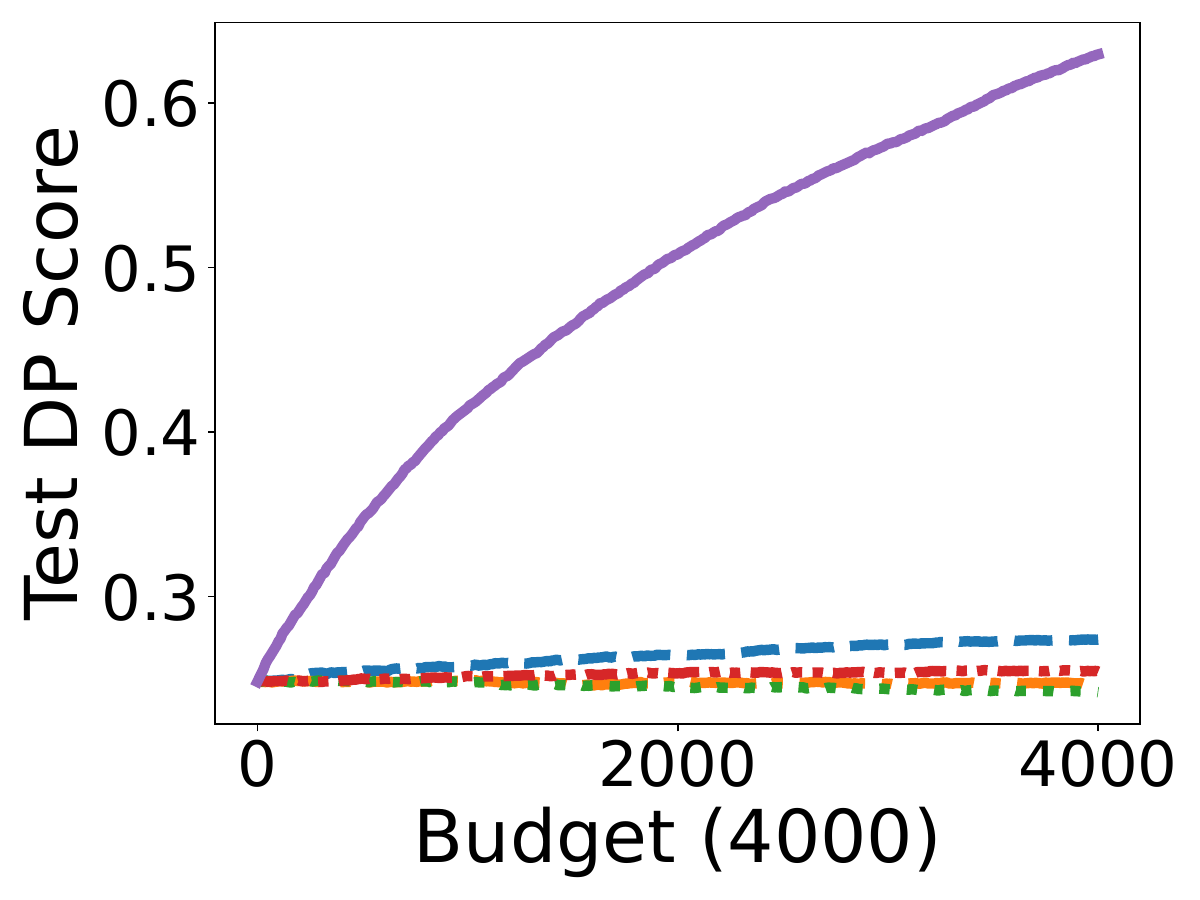}
     \vspace{-0.3cm}
     \caption{{\sf Employ (DP)}}
     \label{fig:simpleemploy}
  \end{subfigure}
  \vspace{-0.2cm}
  \caption{Comparing trial-and-error approach with the baselines on the TravelTime~\cite{ding2021retiring} and Employ~\cite{ding2021retiring} datasets where the target fairness is demographic parity (DP). Only the trial-and-error solution actually improves the DP score.
  }
  \label{fig:simplepolicy}
  \vspace{-0.5cm}
\end{figure}

We now show that this simple trial-and-error approach surprisingly outperforms all fair AL baselines including two state-of-the-art algorithms \fal~\cite{DBLP:journals/eswa/AnahidehAT22} and \decouple~\cite{cao2022decouple}. We train a logistic regression model on two fairness datasets, TravelTime~\cite{ding2021retiring} and Employ~\cite{ding2021retiring}. For the \fal{} and \decouple{} algorithms, we tune their hyperparameters to achieve the highest fairness score. Figure~\ref{fig:simplepolicy} shows the fairness results where the target fairness metric is DP. The x-axis is the labeling budget where 10 samples are selected for labeling in each iteration, and the y-axis is the fairness score on the test set. As a result, our simple solution significantly outperforms all the baselines and actually improves fairness while other baselines are not as effective, as we detail in Section~\ref{exp:accuracyandfairness}. This result clearly demonstrates the importance of postponing undesired samples for improving fairness. For the TravelTime experiment in Figure~\ref{fig:simpletravel}, it even delays about 2,900 samples out of the 4,000 labeling budget.

\paragraph{Trade-off between Informativeness and Postpone Rate} 

While the simple trial-and-error solution is effective, it is not general because it selects samples whose $p(+|x)$ is closest to a fixed threshold of 0.5. However, the bigger picture is that there are two competing factors: the more informative a sample is for improving the target group's accuracy, the less likely it has the target label. So depending on how we set this threshold, there is a trade-off between sample informativeness and postpone rate.

\begin{figure}[t]
\centering
  \includegraphics[width=0.75\columnwidth]{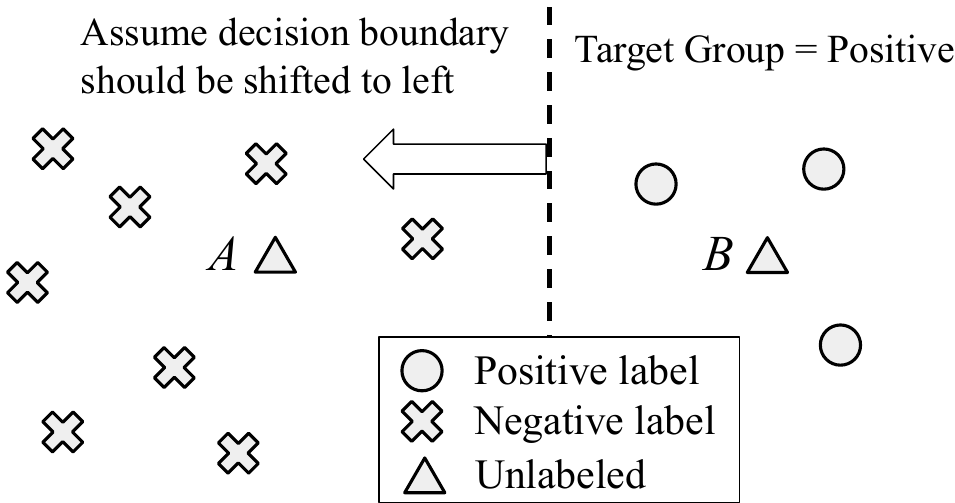}
  \vspace{-0.2cm}
  \caption{If the positive class is our target group, sample $A$ increases the target group accuracy more than $B$ if positively labeled and is thus more informative. However, $A$ is less likely to have the desired target label, leading to a higher postpone rate. It is non-trivial to balance between these two factors.}
  \label{fig:intuition}
  \vspace{-0.3cm}
\end{figure}

Consider the example in Figure~\ref{fig:intuition} where the training data is divided into two groups by their classes, which are either positive or negative. Suppose we want to reduce the accuracy gap between the groups for fairness where the positive class has fewer samples and lower accuracy. Hence, we set the positive class as the target group. The goal is to select a sample from the target group that can effectively shift the decision boundary towards the negative class, which improves the accuracy of the positive class and thus the overall fairness. Comparing the unlabeled samples $A$ and $B$, $A$ has a larger impact on the decision boundary if it turns out to have a positive label and thus more informative, but also has a higher chance to be labeled negatively.


The remaining challenge is how to balance the informativeness of the selected samples and the risk of finding undesired labels, which we cover in the next section. 

\section{Automatic Policy Searching}
\label{sec:policysearching}

We now discuss how to find a desirable policy that balances informativeness and postpone rate. Ideally, we would like to analytically estimate the informativeness and postpone rate, but this is just as hard as determining the labels themselves. A more practical solution, therefore, is to utilize a data-driven method to identify the most effective policy. In this section, we introduce a multi-armed bandit based solution for selecting the best policy.  


\subsection{Policies for Sample Selection}
\label{sec:policiesforsampleselection}
Depending on the relative importance we assign to selecting informative samples versus samples with desirable labels, we can have different policies for sample selection. The more ``risk'' we are willing to take for finding an informative sample, the less likely it has the desired label. \rev{For each target group, we capture this risk taking as a policy, which is defined as follows:}

\begin{definition} \textbf{\rev{Policy.}}
\label{def:policy}
\rev{Given a target group $(y, z)$ and a level of risk-taking $c$, a policy $r=c$ selects a sample $x$ from the $z$ group whose predicted probability for the label $y$, $p(\hat{y}=y|x)$, is closest to $1-c$.}
\end{definition}

\rev{For example, EO has one target group (e.g., \texttt{(attribute=female, label=positive)}), and we can use a policy set like $[r=0.3, r=0.5, r=0.7]$, where $r=0.7$ is a most risk-taking policy and selects a female sample whose predicted probability for a positive label is closest to 0.3}. If we use DP, there are two target groups (e.g., \texttt{(attribute=female, label=positive)} and \texttt{(attribute= male, label=negative)}), so we consider twice the number of policies, e.g., $[r=0.3, r=0.5, r=0.7]$ for each target group.



\begin{figure}[t]
  \centering
    \begin{subfigure}{0.495\columnwidth}
     \includegraphics[width=\columnwidth]{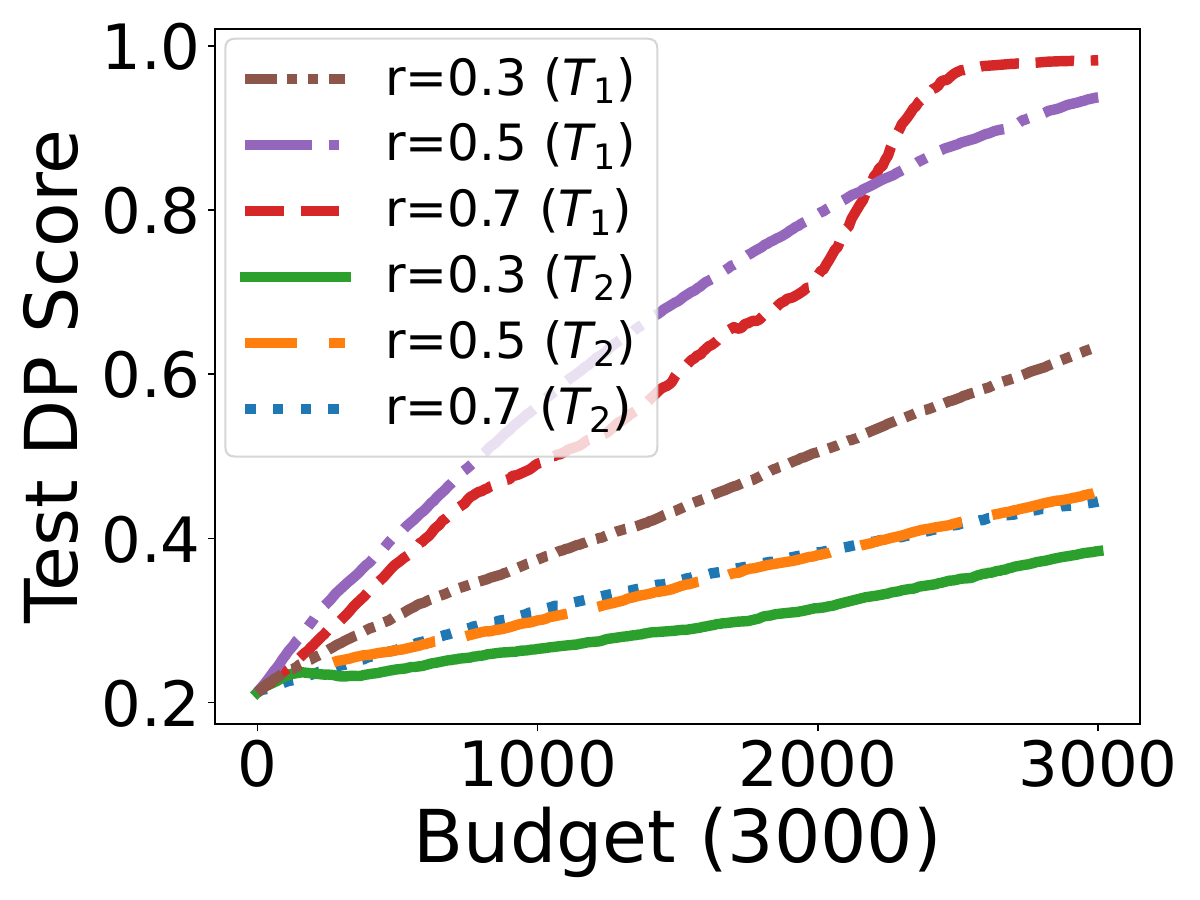}
     \vspace{-0.5cm}
     \caption{{\sf TravelTime-DP}}
     \label{fig:multiplepolicestime}
  \end{subfigure}
  \centering
    \begin{subfigure}{0.495\columnwidth}
     \includegraphics[width=\columnwidth]{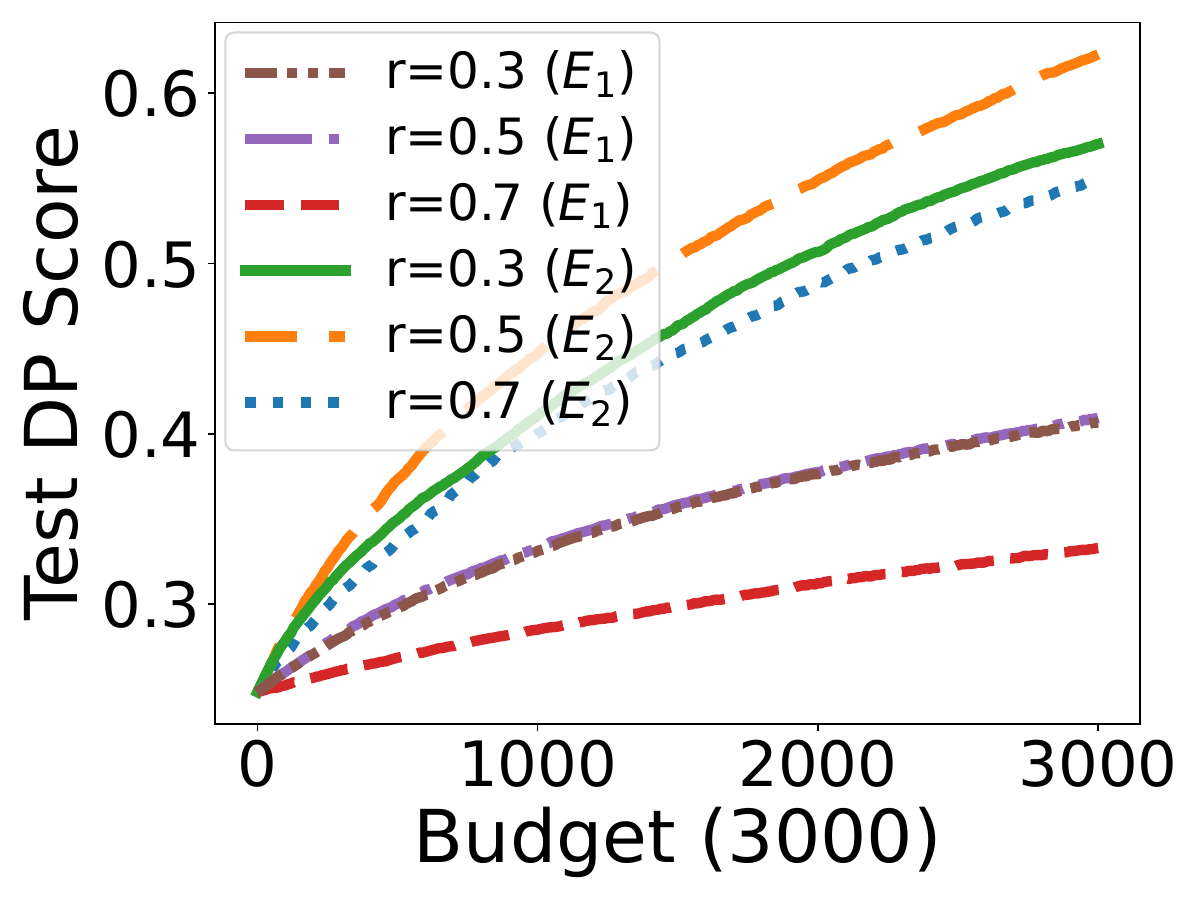}
     \vspace{-0.5cm}
     \caption{{\sf Employ-DP}}
     \label{fig:multiplepolicesemploy}
  \end{subfigure}
\vspace{-0.7cm}
  \caption{Policy comparison using the TravelTime and Employ datasets where the $i^{th}$ target groups are denoted as $T_i$ and $E_i$, respectively. DP fairness is used. The best policy depends on the dataset and how much labeling has been done. 
  }
  \label{fig:multiplepolices}
  \vspace{-0.5cm}
\end{figure}

Choosing the right policy is non-trivial because the decision boundary may shift as more samples get labeled, which means the most effective policy may change as well. Furthermore, this outcome varies by dataset. To demonstrate these points, we compare the performances of individual policies using a policy set of $[r=0.3, r=0.5, r=0.7]$ per target group. Figure~\ref{fig:multiplepolices} shows the fairness results on the TravelTime and Employ datasets where the target fairness is DP. For simplicity, we denote the $i^{th}$ target subgroup for TravelTime and Employ as $T_i$ and $E_i$, respectively. As a result, no single policy is always the best, and the best policy sometimes changes as we label more samples. For example, on the TravelTime dataset, $r=0.5$ for the target group $T_{1}$ performs well at the early stages, but as more samples are labeled, $r=0.7$ starts to perform better (Figure~\ref{fig:multiplepolicestime}). In comparison, $r=0.5$ for $E_2$ consistently performs the best for Employ (Figure~\ref{fig:multiplepolicesemploy}). Another observation is that there is a significant difference in fairness improvement between the target groups. For both datasets, even the worst policy in the best target group outperforms the best policy in the other group. Hence, we cannot rely on a fixed strategy for finding the best policy and need an adaptive approach instead.


Our solution is to use a multi-armed bandit (MAB) approach, which balances exploration and exploitation to find the optimal policy based on the rewards in terms of improved fairness. We discuss details of the MAB-based approach in the next section.

\subsection{Adversarial MAB for Policy Search}
\label{sec:adversarialmab}
We utilize and extend existing multi-armed bandit (MAB) techniques to choose the best policy based on previous rewards. Using an MAB is a standard approach for allocating limited resources to competing choices (i.e., pulling arms) when the resulting rewards are only partially known~\cite{10.1007/11564096_42}. In our setup, the competing choices are selecting the right policies where the reward is the fairness improvement, and the labeling effort is the limited resource. The key challenge is to balance exploration and exploitation where too much exploration of choices may lead to not utilizing the knowledge we already have, while too much exploitation may lead to missing opportunities of discovering better choices.

In particular, we use adversarial MABs~\cite{DBLP:journals/siamcomp/AuerCFS02, DBLP:journals/jmlr/BeygelzimerLLRS11}, which do not make assumptions about the reward distribution and only make choices to pull arms based on rewards. Traditional MABs~\cite{10.1023/A:1013689704352} assume that the rewards follow a fixed and time-invariant distribution and provide theoretical bounds on regret, which is the expected difference between the sum of rewards in an optimal offline strategy versus the actual rewards obtained. 
In our setup, however, the fairness improvement changes as we label more samples, as shown in Figure~\ref{fig:multiplepolices}. That is, the reward function does not follow statistical assumptions anymore. In addition, choosing one policy may have unpredictable effects on the rewards of using other policies. The reason is that using a policy results in actual data labeling, which improves the fairness of the current model and influences which samples to label for any other policy in the future. More specialized MABs like rotting bandits\,\cite{rottingbandit} make the assumption that rewards are independent and always decreasing, but this does not hold in our setup either. For example, if one subgroup is targeted and labeled more, then another subgroup's accuracy may actually decrease due to the influence. Then the next reward of the second subgroup can actually increase, as there is more opportunity to improve fairness. Since adversarial MABs make no assumption about the rewards, their behaviors are more conservative while having strong theoretical guarantees on regret bounds in the adversarial setup.

\rev{Although \method{} can be paired with any adversarial MAB, we choose the EXP3\,\cite{DBLP:journals/siamcomp/AuerCFS02} algorithm, which achieves an expected regret of $O(\sqrt{KT ln K})$ where $K$ is the number of arms and $T$ is the time horizon. Our choice is based on the fact that EXP3 is a representative method whose empirical performance is no worse than more recent MABs, which yield the same regret bounds with a smaller variance (referred to as high probability regret bound), as we detail in Section~\ref{exp:mabs}.} Algorithm~\ref{alg:mab} shows how EXP3 learns the optimal policy for given rewards. At each time step, EXP3 chooses an arm according to the selection probability (Lines 3--4). This probability is a combination of the uniform distribution and another distribution that assigns probabilities to each action proportional to the exponential of the cumulative rewards for that action. Since some arms may later be useful, mixing in the uniform distribution ensures that the algorithm keeps on giving each arm a chance to be selected. In addition, the estimated gain is calculated by dividing the actual gain by the selection probability, which compensates for the reward of actions that are unlikely to be chosen (Line 7). EXP3 then updates the policies based on the rewards (Line 8) and repeats these steps for $T$ iterations. Here the only assumption is that the reward value should be within the range of [0, 1] (Line 5).

\rev{For example, consider two policies $P_1$ and $P_2$, where $P_1$ initially yields high rewards, while $P_2$ starts with very low rewards but becomes more beneficial in later iterations. EXP3 begins increasing the probability of selecting $P_1$ but within a certain bound, mixing its probability with a uniform distribution. As a result, EXP3 still offers $P_2$ a chance to be chosen in later phases and will adapt the selection probabilities to changing rewards.}


\paragraph{Efficiency} 
One advantage of the MAB-based approach is its high efficiency, as updating the MAB does not require model training. In contrast, other baselines need multiple model trainings for sample selection, which is not as efficient as \method{} (see Section~\ref{exp:scalability}).

\subsection{Reward Design} \label{sec:rewarddesign}
The choice of the reward signal directly impacts the quality of MAB. We define the reward as the fairness improvement on the validation set. However, there are largely two challenges with the reward.

First, the reward obtained after labeling a few samples is usually very small, which has a negligible impact on updating policies. Hence, we propose a simple solution where we allocate the first $L$ iterations to run the algorithm and then use the initial rewards to normalize the upcoming reward values. This normalization ensures that the reward has a more significant impact on updating the selection probabilities while still being within the [0, 1] range. 

\SetKwComment{Comment}{// }{}
\begin{algorithm}[t]
    \SetKwInput{Input}{Input}
    \SetKwInOut{Output}{Output}
    \Input{Real $\gamma \in (0, 1]$, Arms $K$, Time horizon $T$}
    \Output{Updated probabilities}
    Initialize weights $w = 1$ \;
    \For{$t = 1,2, \ldots, T$}{
      Set $p_i(t) = (1-\gamma) \frac{w_i(t)}{\sum_{j=1}^K w_j(t)} + \frac{\gamma}{K}$\;
      Draw $i_t$ randomly accordingly to the probabilities $p_1(t), \ldots, p_K(t)$\;
      Receive reward $r_{i_t} \in [0, 1]$\;
      \For{$j=1, \ldots, K$}{
        $\hat{r_j}(t)$ =
        $
        \begin{cases}
            \frac{r_j(t)}{p_{i_t}} & \text{if } j = i_t \\
            0, & \text{otherwise} \\
        \end{cases}
        $
        \\
        $w_j(t+1) = w_j(t) exp(\gamma \hat{r_j}(t)/K)$ \;
      }
    }
    \caption{EXP3 algorithm\,\cite{DBLP:journals/siamcomp/AuerCFS02}.}
    \label{alg:mab}
\end{algorithm}

In addition, pulling one arm improves the fairness of the current model, but potentially limits the chances for other policies to be updated. The reason is that the total budget is limited, and the fairness improvement decreases as we label more data. If one sub-optimal policy is picked by chance in the early stages and receives a reasonable reward, the MAB may continue selecting that policy instead of searching for the optimal one. The dependency between policies makes it more challenging for the EXP3 algorithm to identify the best policy if it is not chosen sufficiently in previous iterations.




\rev{We thus propose a reward propagating scheme that distributes the reward of a selected arm to its neighbors. This approach is inspired by a previous MAB-based data acquisition framework~\cite{DBLP:journals/pvldb/ChaiLTLL22} that also assumes dependent arms.} Intuitively, if two policies are close to each other (i.e., similar $r$ values), their actual reward values tend to be similar. We thus propagate half of the obtained reward to the nearest policies, ensuring that the unknown best policy still receives some rewards even if it is not selected. Specifically, if policy $P_{i}$ is selected and obtains a reward value of $\hat{r_i}(t)$ at time step $t$, we compute rewards for the remaining policies $P_j$ as:
\begin{equation} \label{eq:rewardpropaggation}
    \hat{r_j}(t) =
    \begin{cases}
        \hat{r_i}(t) \times 0.5 & \text{if } P_{j} \in NN(P_{i}) \\
        0 & \text{otherwise} \\
    \end{cases}
\end{equation}
where $NN(P_i)$ denotes the nearest policies to $P_i$ that has the same target group. For example, suppose \method{} chooses $r=0.5$ for \texttt{(attribute=female, label=positive)} and gets a reward of 1.0. We then assign a reward of 0.5 each to the $r=0.4$ and $r=0.6$ for the female group. We use this simple design, although one can propose other reward functions that capture a similar intuition.

\section{\method{} Algorithm}
\label{sec:algorithm}

We now describe how \method{} can be extended with traditional AL for selecting samples for the purpose of improving both fairness and accuracy. We explain the extension of \method{} with AL and present the overall algorithm.

\paragraph{Improving Both Fairness and Accuracy}
Our approach is to alternate between fair and accurate labeling probabilistically. Specifically, we improve fairness with $\lambda$ probability and accuracy with $1-\lambda$ probability, where $\lambda \in [0, 1]$ is a hyperparameter that balances between fairness and accuracy. Here, a higher value of $\lambda$ means better fairness. We can thus naturally integrate \method{} with other AL methods without requiring significant modifications. A similar blending idea has also been proposed in a fair adaptive sampling work~\cite{pmlr-v162-abernethy22a}, where the goal is to sample ``labeled'' data to improve accuracy and a specialized fairness metric called min-max fairness. A full analysis on how our MAB approach converges and deriving the regret bound in this setup is interesting future work, but we show extensive empirical results on how one can indeed balance fairness and accuracy by adjusting the interleaving degrees in Section~\ref{exp:accuracyandfairness}.

\paragraph{\method{} Algorithm with AL} 
We now present the \method{} algorithm including AL in Algorithm~\ref{alg:falcon}. We have a fixed labeling budget of $B$ where $b$ samples are selected for labeling per iteration. Within each iteration, we first decide the labeling strategy to use based on the $\lambda$ probability (Line 3). If we perform labeling for the purpose of accuracy, we run the given AL algorithm (Lines 5--6). Otherwise, we improve fairness. We first determine which subgroups to label for improving the target fairness metric $F$ (Line 8) and construct an MAB for those target groups (Line 9). Then, the MAB selects a policy for labeling $b$ samples and obtains a reward (Section~\ref{sec:rewarddesign}), which measures the fairness improvement after labeling samples using the chosen policy (Lines 10--13). The reward is used to update the MAB (Line 14). We repeat these steps until we run out of budget.

\paragraph{\rev{Running Time Analysis}} 
\rev{The primary cost in \method{} is retraining the model with newly labeled data. This process requires $\frac{B}{b}$ number of model training. Such retraining is a standard requirement in any active learning technique. Other components in \method{} include identifying target subgroups, selecting samples based on policies, and updating the MAB, only incur a small overhead (see Section~\ref{exp:scalability}).}


\SetKwComment{Comment}{// }{}
\begin{algorithm}[t]
    \SetKwInput{Input}{Input}
    \SetKwInOut{Output}{Output}
    \Input{Train data $D_{train}$, Validation data $D_{val}$, Unlabeled data $D_{un}$, Labeling budget $B$, Batch size $b$, Set of policies $P$, Fairness measure $F$, AL method $A$, Blending parameter $\lambda \in [0, 1]$}
    \Output{Trained model $M$, Updated datasets $D_{train}$, $D_{un}$}

    $M$ = \textsc{TrainModel}($D_{train}$, $D_{val}$) \; 
    \For{$t = 1,2, \ldots, \frac{B}{b}$}{
        x $\sim$ \textsc{Bernoulli}($\lambda$) \;
        \Comment{Improve fairness ($\lambda$) or accuracy (1-$\lambda$)}
        \If{$x=0$}{
            \Comment{Improve accuracy using AL}
            $D_{train}, D_{un}$ = \textsc{LabelData}($D_{train}$, $D_{un}$, $M$, $A$, $b$) \;
            $M_{new}$ = \textsc{TrainModel}($D_{train}$, $D_{val}$) \;
        }
        \Else{
            \Comment{Improve fairness}
            $target\_groups$ = \textsc{GetTargetGroups}($M$, $D_{val}$, $F$)\;
            $target\_MAB$ = \textsc{GetMAB}($target\_groups$, $P$) \;
            $P_k$ = SelectPolicy($target\_MAB$) \; 
            $D_{train}, D_{un}$ = \textsc{LabelData}($D_{train}$, $D_{un}$, $M$, $P_k$, $b$) \;
            $M_{new}$ = \textsc{TrainModel}($D_{train}$, $D_{val}$) \; 
            $R$ = \textsc{GetReward}($M_{new}$, $M$, $D_{val}$, $F$, $k$) \;
            \textsc{UpdateMAB($target\_MAB$, $R$)} \;
        }
        $M$ = $M_{new}$
    }
    \caption{Overall \method{} algorithm.}
    \label{alg:falcon}
\end{algorithm}



\paragraph{Choice of Policy Set} 
We discuss how we select the policy set for \method{}. Overall, the more candidate policies, the more likely there is an optimal policy within them. \rev{However, using an infinite number of policies is not practical because the labeling budget is limited. Even if we do have an infinite budget, we would have to use infinitely many-armed bandits~\cite{NIPS2008_49ae49a2, pmlr-v162-kim22j} or Bayesian optimization~\cite{10.5555/2999325.2999464} for modeling continuous policies, but they are not designed for adversarial rewards.} We thus need to select a reasonable number of policies that are not extreme and use $[r=0.3, r=0.4, r=0.5, r=0.6, r=0.7]$ as a default policy set throughout the paper. We discuss the effect of the number of policies in Section~\ref{exp:numpolicies}.

\paragraph{Batch Size Effect} 
We discuss the effect of the batch size on \method{}'s performance. The larger the batch size, the better reward signals we can utilize. On the other hand, there are fewer chances to adjust the policies based on the rewards. If labeling a sample only has a small effect in improving the model performance, we prefer using a larger batch size. In our experiments, we use larger batch sizes on larger datasets for efficiency. 
There are also other MABs~\cite{batchbandit} that are designed for handling batches. However, a technical hurdle is that we need to know the rewards of individual actions for each batch. The rewards are not readily available in our setup, as we only have an aggregated reward per batch. Investigating how to utilize batch-specialized MABs is an interesting future work. 

\paragraph{Multiple Sensitive Attributes} 
While the previous examples assumed binary sensitive attributes, \method{} can be readily extended to support multiple sensitive groups (i.e., $\mathbb{Z} = \{0, 1, ..., n_{z}-1\}$). Again, the fairness score is computed as one minus the maximum disparity value among any sensitive groups, as we explain in Section~\ref{sec:preliminaries}. Hence, we can construct an MAB based on the two groups that currently show the highest disparity value and continue to follow the same procedure. If the target group pair changes during the labeling process, we can simply switch the target MAB accordingly.

\section{Experiments}
\label{sec:experiments}
In this section, we evaluate \method{} on real datasets and address the following key questions: (1) How does \method{} compare with the baselines in terms of model accuracy, fairness, and running time? (2) Does \method{} find the best policy and perform in various scenarios? and (3) How useful is each component of \method{}?

We implement \method{} in Python, use Scikit-learn~\cite{scikit-learn} for model training, and run all experiments on Intel Xeon Silver 4210R CPUs using ten different random seeds.


\subsection{Setting} \label{sec:setting}

\paragraph{Datasets} We use four popular datasets in the fairness literature. For the first three datasets, we use the same feature pre-processing as in the Folktables~\cite{ding2021retiring} package, and for the COMPAS dataset, we apply the method provided by IBM’s AI Fairness 360 toolkit~\cite{aif360}. For a detailed evaluation, we consider various sensitive attributes, including scenarios with multiple sensitive attributes.

\begin{table}[t]
  \centering
\small
  \begin{tabular}{ccccc}
  \toprule
  {\bf Dataset} & $\mathbf{|D_{train}|}$/$\mathbf{|D_{un}|}$/$\mathbf{|D_{test}|}$/$\mathbf{|D_{val}|}$ & {\bf Sen. Attr} & {\bf Batch} & $\mathbf{B}$\\
  \hline
  TravelTime & 2,446/48,940/24,470/2,446 & gender & 10 & 4K  \\
  Employ & 5,432/162,960/81,480/5,432 & disability & 10 & 4K \\
  Income & 3,188/63,760/31,880/3,188 & race & 10 & 4K \\
  COMPAS & 294/2,356/1,178/294 & gender & 1 & 200\\
  \hline
  \end{tabular}
  \caption{Parameters for the four datasets.}
  \label{tbl:parameters}
  \vspace{-0.8cm}
\end{table}

\begin{figure*}[t]
  \centering
  \begin{subfigure}{0.246\textwidth}
     \includegraphics[width=\columnwidth]{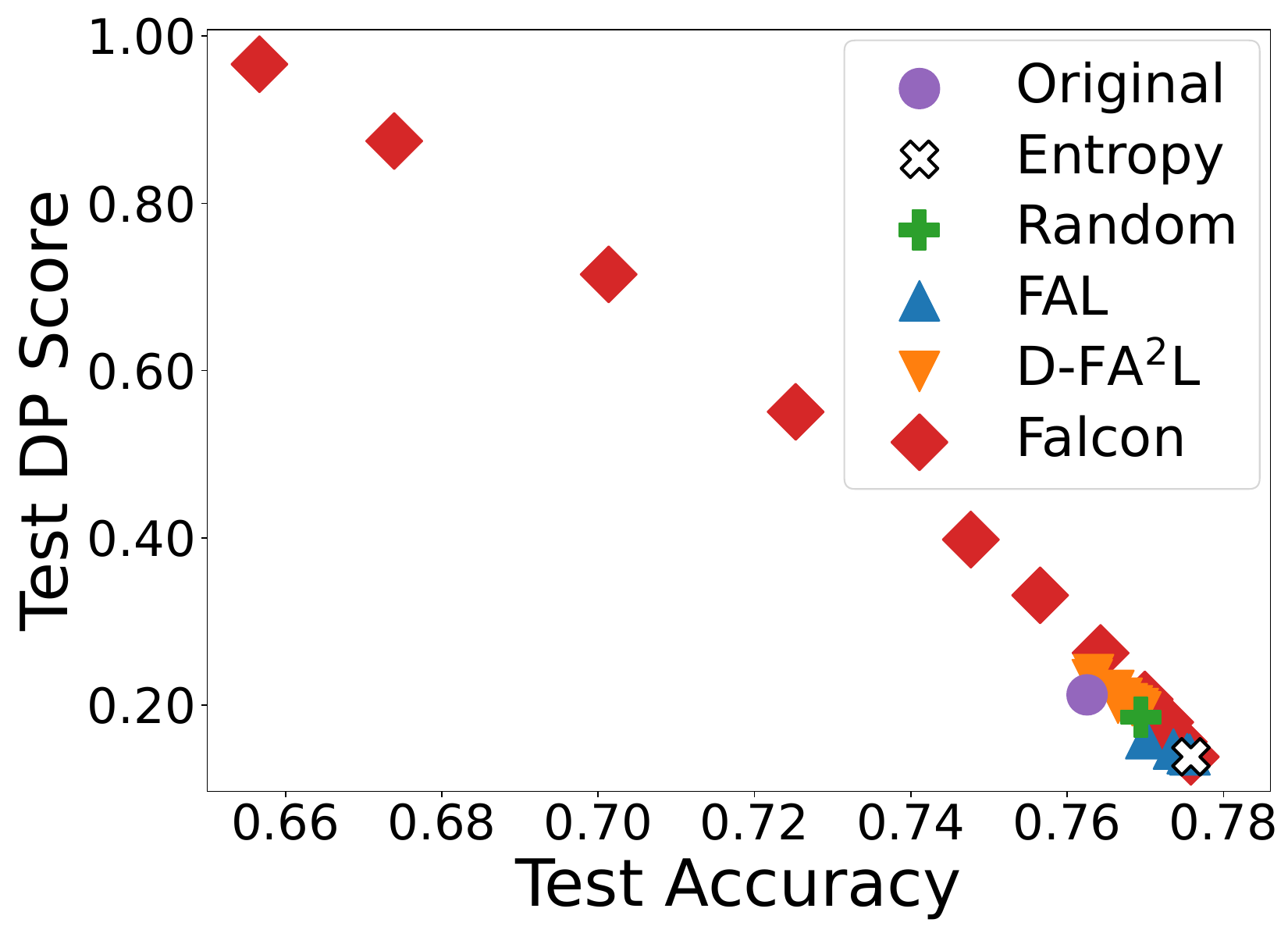}
     \vspace{-0.6cm}
     \caption{TravelTime (DP)}
     \label{fig:acsdp}
  \end{subfigure}
  \begin{subfigure}{0.246\textwidth}
    \includegraphics[width=\columnwidth]{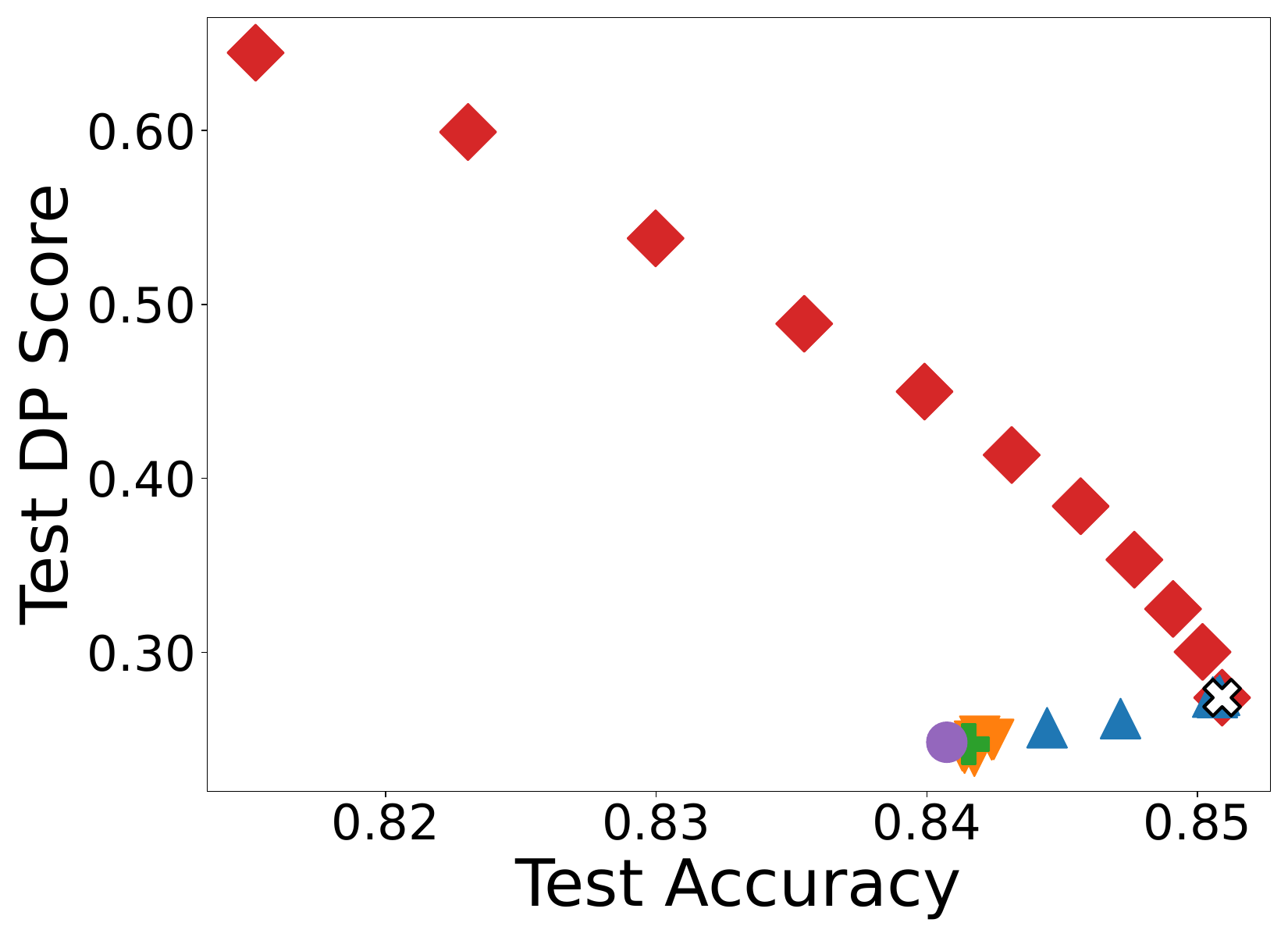}
    \vspace{-0.6cm}
     \caption{Employ (DP)}
     \label{fig:adultdp}
  \end{subfigure} 
  \begin{subfigure}{0.246\textwidth}
    \includegraphics[width=\columnwidth]{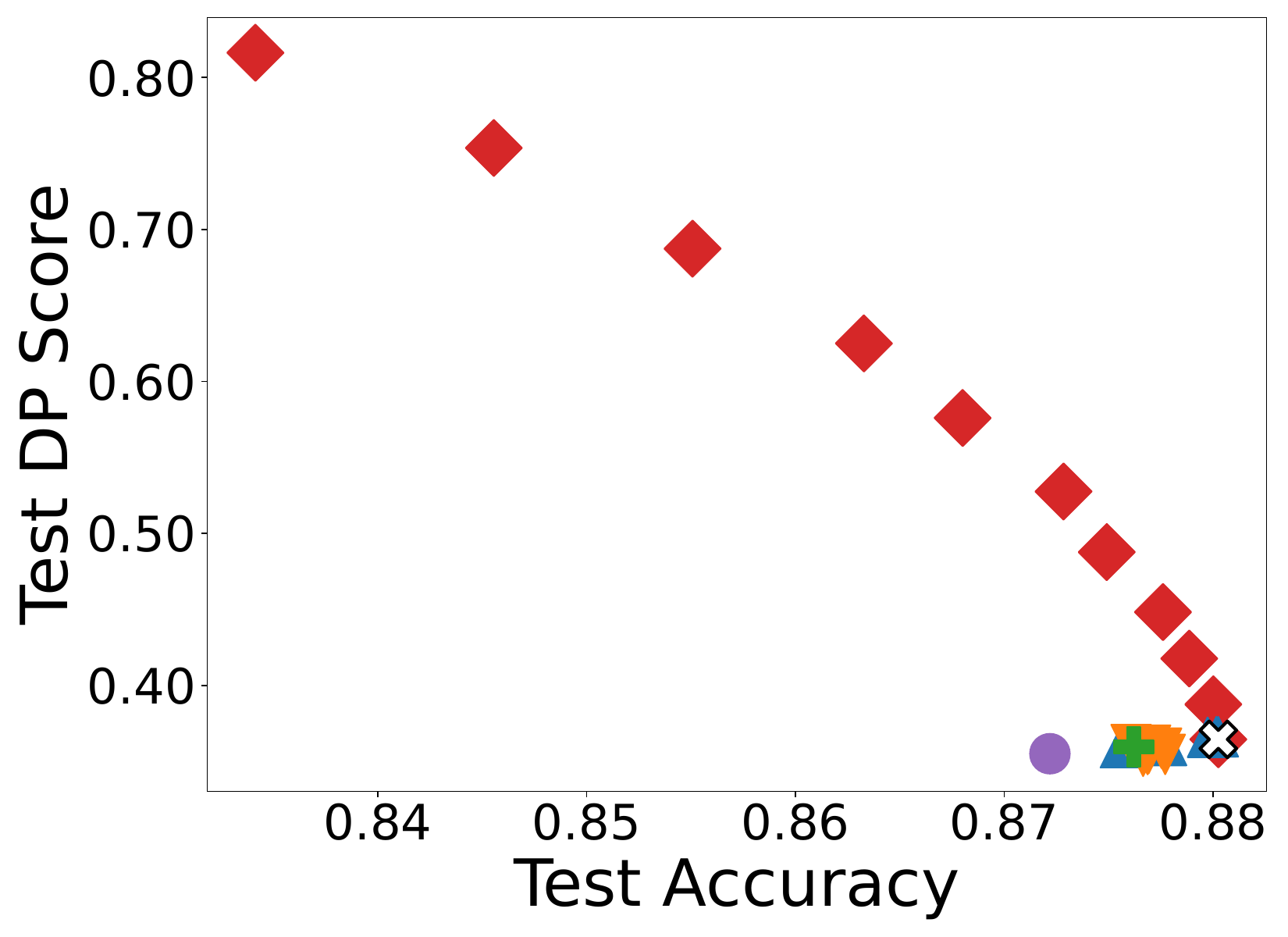}
    \vspace{-0.6cm}
     \caption{Income (DP)}
     \label{fig:compasdp}
  \end{subfigure}
\begin{subfigure}{0.246\textwidth}
    \includegraphics[width=\columnwidth]{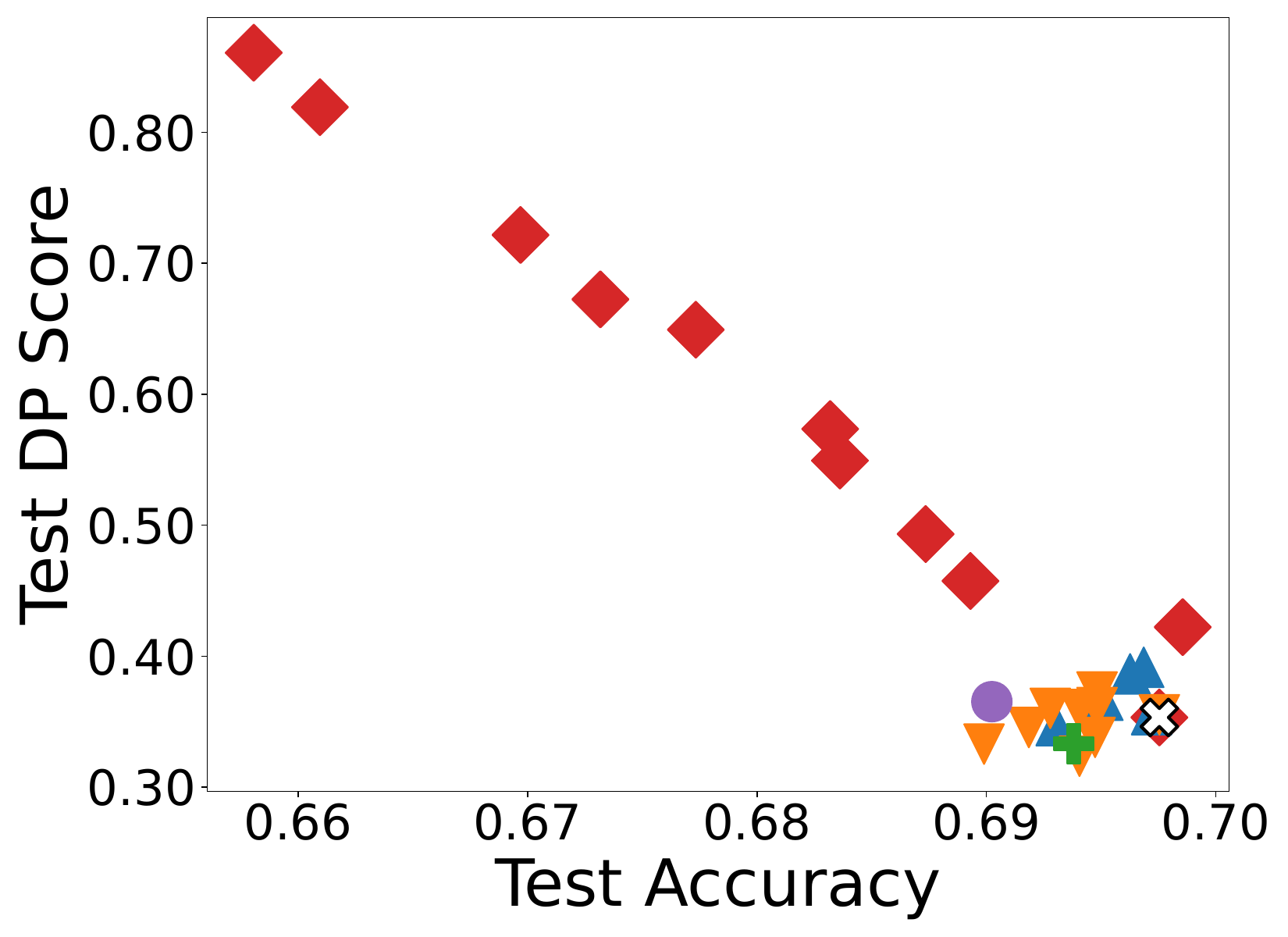}
    \vspace{-0.6cm}
     \caption{COMPAS (DP)}
     \label{fig:celebdp}
  \end{subfigure}
  \begin{subfigure}{0.246\textwidth}
     \includegraphics[width=\columnwidth]{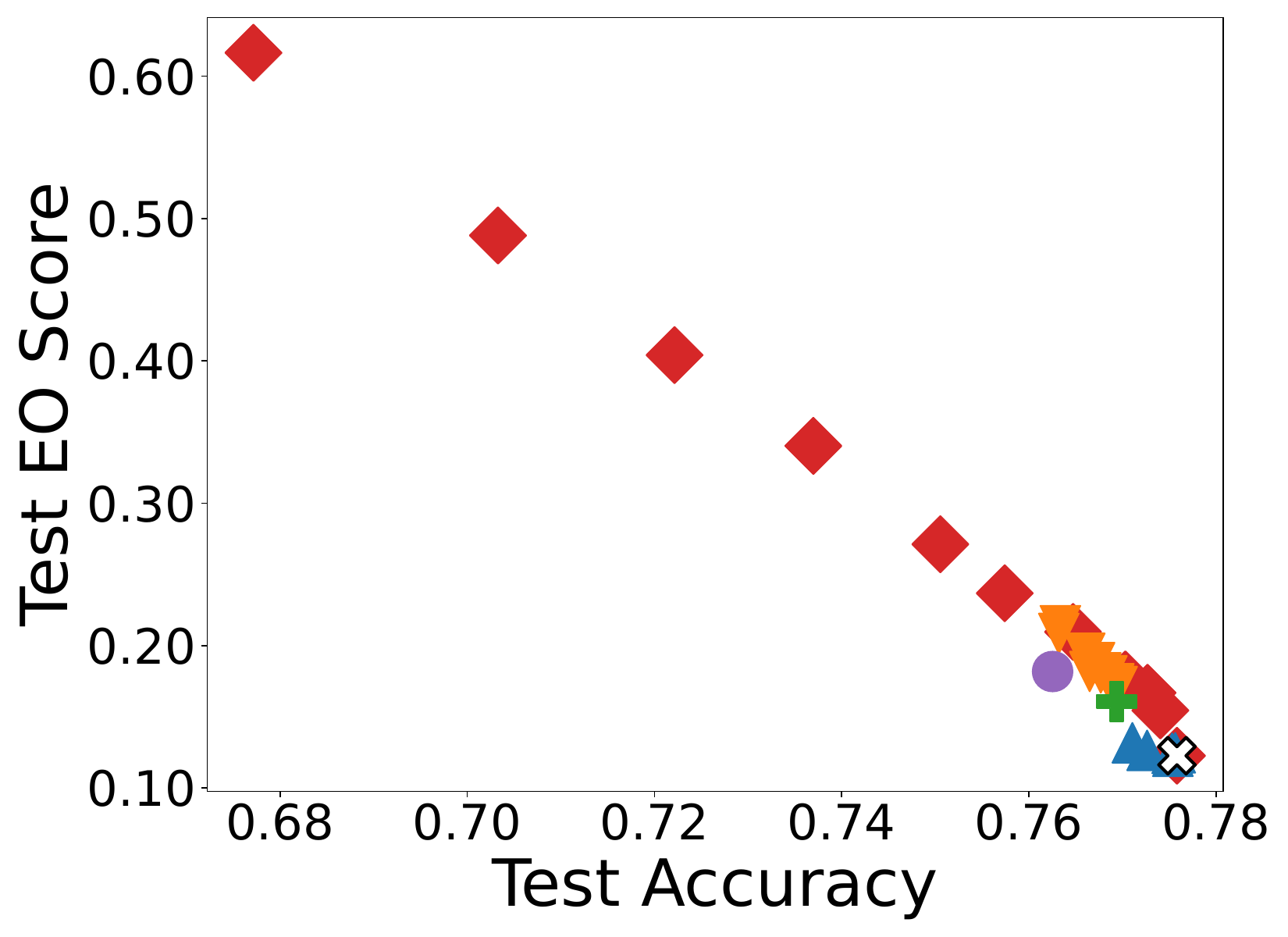}
     \vspace{-0.6cm}
     \caption{TravelTime (EO)}
     \label{fig:acsed}
  \end{subfigure}
  \begin{subfigure}{0.246\textwidth}
    \includegraphics[width=\columnwidth]{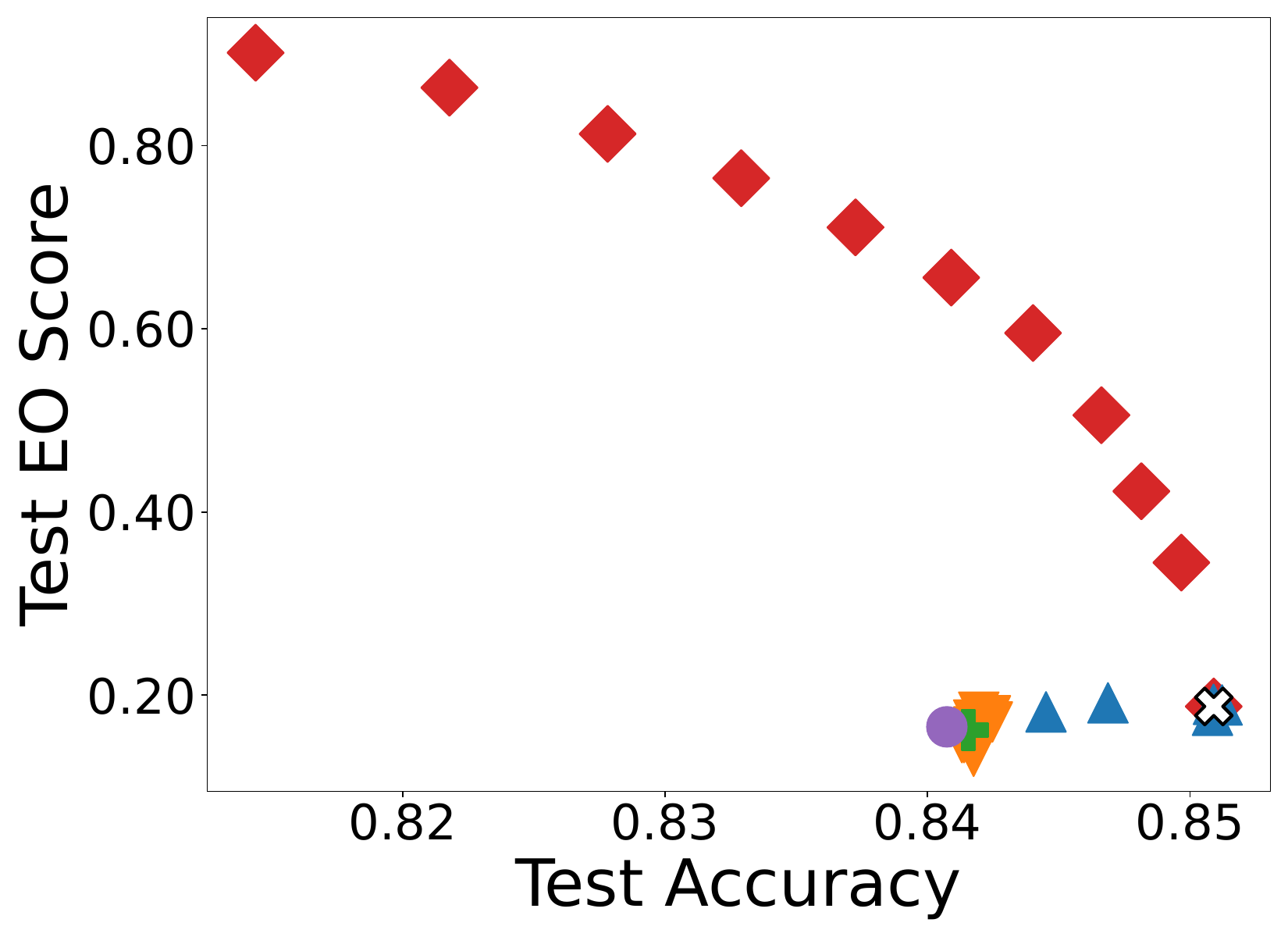}
    \vspace{-0.6cm}
     \caption{Employ (EO)}
     \label{fig:adulted}
  \end{subfigure} 
  \begin{subfigure}{0.246\textwidth}
    \includegraphics[width=\columnwidth]{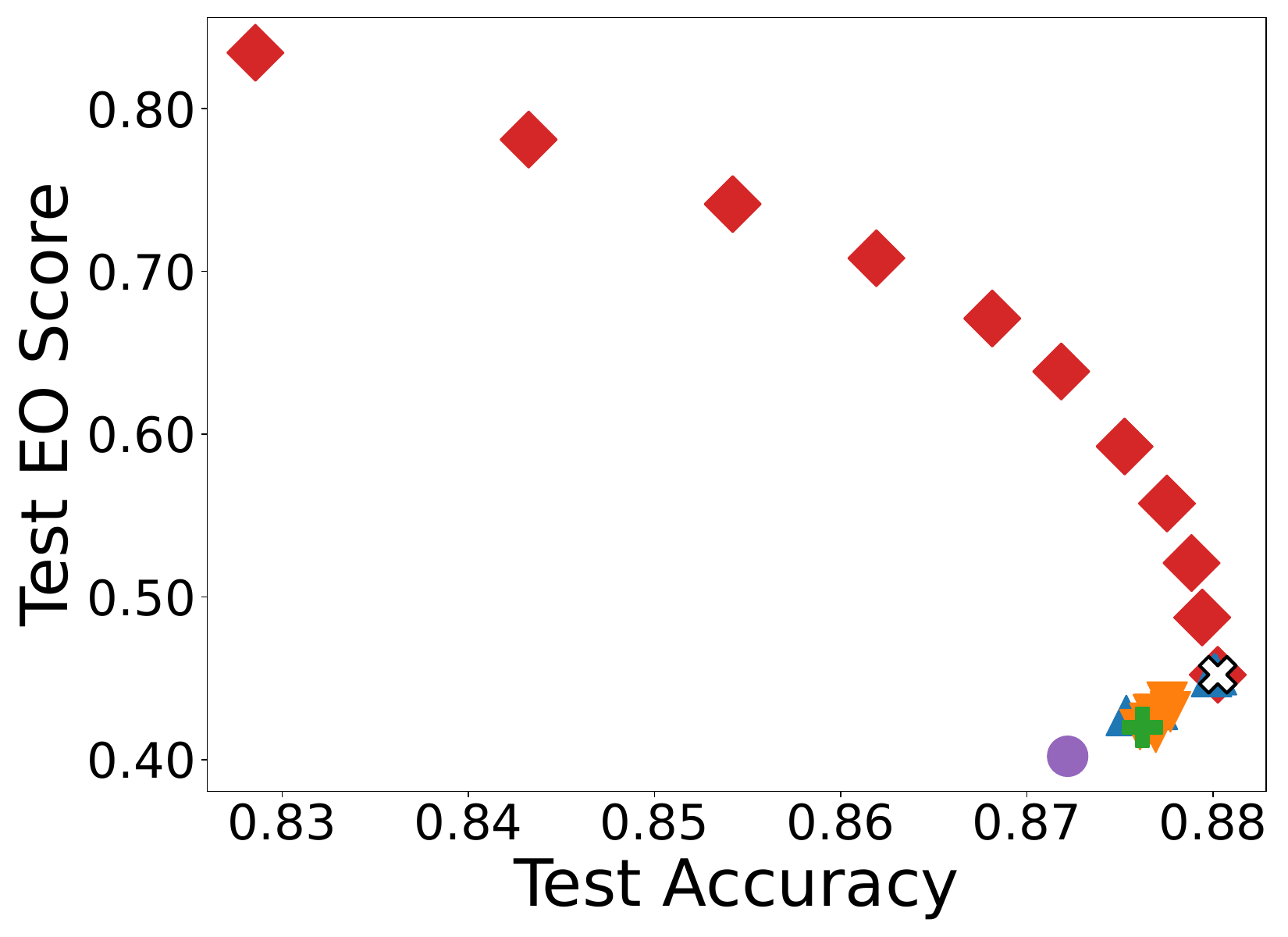}
    \vspace{-0.6cm}
     \caption{Income (EO)}
     \label{fig:compased}
  \end{subfigure}
\begin{subfigure}{0.246\textwidth}
    \includegraphics[width=\columnwidth]{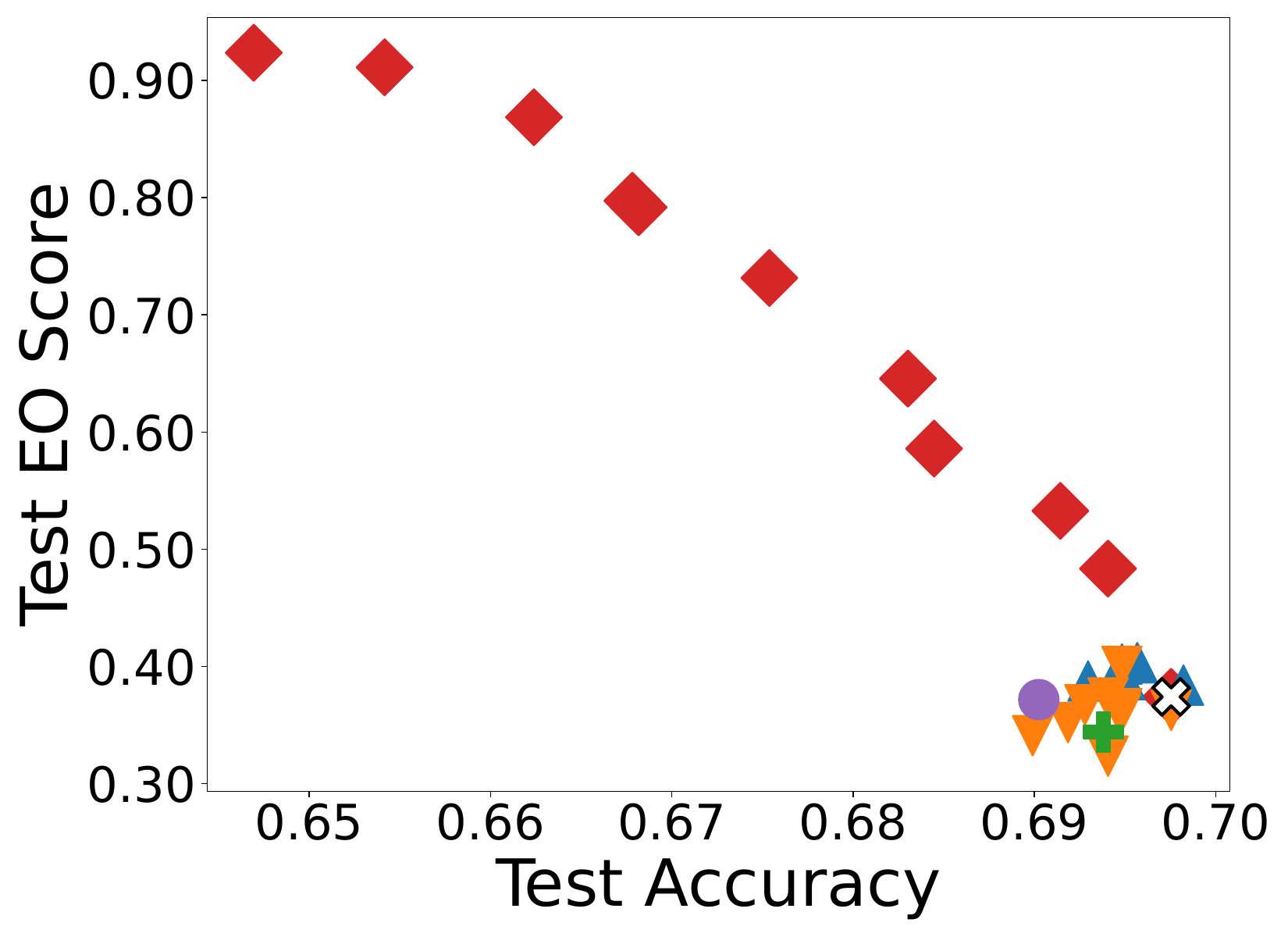}
    \vspace{-0.6cm}
     \caption{COMPAS (EO)}
     \label{fig:celebed}
  \end{subfigure}
  \vspace{-0.3cm}
     \caption{Accuracy-fairness trade-off results on the four datasets and two fairness measures. In addition to the baselines, we add the result of model training without labeling any additional data and call it ``Original.'' As a result, only \method{} significantly improves fairness and shows clear accuracy and fairness trade-offs. Note that \method{} using $\lambda=0$ is the same as \entropy{}.
     }
 \label{fig:tradeoffs}
 \vspace{-0.35cm}
\end{figure*}

\begin{itemize}[leftmargin=*]
\item TravelTime~\cite{ding2021retiring}: Used to predict whether an employee has a commute to work that is longer than 20 minutes. We use \texttt{gender} as the sensitive attribute.
\item Employ~\cite{ding2021retiring}: Used to predict whether an individual is employed. We use \texttt{disability} as the sensitive attribute.
\item Income~\cite{ding2021retiring}: Used to predict whether an individual's income exceeds \$50K per year. We use \texttt{race} as the sensitive attribute and consider three distinct groups: White, Asian, and Others.
\item COMPAS~\cite{machinebias}: Used to predict an individual's criminal recidivism risk. We use \texttt{gender} as the sensitive attribute.
\end{itemize}

For each dataset, we split the dataset into training, test, unlabeled, and validation sets, as shown in Table~\ref{tbl:parameters}. In addition, we use a total labeling budget of $B=4K$ except for the COMPAS dataset where $B=200$ is already enough to obtain high fairness. See our technical report~\cite{techreport} for detailed configurations. 



\paragraph{Fairness evaluation} We consider five group fairness measures in Section~\ref{sec:preliminaries}. To quantify fairness, we define a fairness score as one minus the maximum fairness disparity~\cite{bird2020fairlearn} across any sensitive groups on the test set. A higher score indicates better fairness, and we provide the detailed equations in our technical report~\cite{techreport}.



\paragraph{Parameters} We assume a batch AL setup and use different default batch sizes for the datasets as shown in Table~\ref{tbl:parameters}.  We use a default policy set of [$r = 0.3$, $r = 0.4$, $r = 0.5$, $r = 0.6$, $r = 0.7$]. We investigate the impact of batch sizes and the number of policies in \Cref{exp:batchsize} and \Cref{exp:numpolicies}, respectively.

\paragraph{Baselines Compared} We compare \method{} with existing fair AL and AL algorithms.
\begin{itemize}[leftmargin=*]
\item \entropy{}~\cite{entropy2001shannon}: A standard AL algorithm that selects the most uncertain samples based on entropy.
\item \random{}: A randomized algorithm that uniformly selects unlabeled data samples.
\item \fal{}~\cite{DBLP:journals/eswa/AnahidehAT22}: The first fairness-aware AL algorithm that optimizes both group fairness and accuracy. \fal{} selects the top $m$ points with the highest entropy value and then chooses samples that also have the maximum expected reduction in unfairness. A higher $m$ favors better fairness.
\item \decouple{}~\cite{cao2022decouple}: A disagreement-based fairness-aware AL algorithm. \decouple{} selects samples for which the decoupled models, trained separately on different sensitive groups, provide different predictions. \decouple{}'s primary goal is to improve DP, but we also consider other fairness measures for a detailed comparison.
\end{itemize}


For the fair AL baselines, there are hyperparameters that can control an accuracy-fairness trade-off. We start with the default hyperparameters as described in the original papers, and tune them to provide the best results (see our technical report~\cite{techreport} for details).



\paragraph{Model Setup}
We use logistic regression (LR) and neural network (NN) models. \rev{For NN, we use a multi-layer perceptron with one hidden layer consisting of 10 nodes. We tune the model hyperparameters such that the trained model has the highest validation accuracy. More detailed settings are in our technical report~\cite{techreport}}.


\subsection{Accuracy and Fairness} 
\label{exp:accuracyandfairness}

We compare the accuracy and fairness results of \method{} with the other baselines using the four datasets. In the main experiments, we use demographic parity (DP) and equal opportunity (EO). The results for other fairness metrics are similar and can be found in our technical report~\cite{techreport}. Figure~\ref{fig:tradeoffs} shows the trade-off results with logistic regression models where the x-axis is the accuracy, and the y-axis is the fairness score on the test set. {\em Original} is where we train a model on the original data without performing labeling. For \fal{} and \decouple{}, we employ 5 and 9 different sets of hyperparameters, respectively. For \method{}, we use 11 different $\lambda$ values ranging from 0.0 to 1.0. \rev{As a result, \method{} shows the best accuracy and fairness trade-off compared to the baselines, which have noisy and even overlapping results for different hyperparameters. Concretely, \method{} improves the fairness score by up to 0.81 with up to 0.12 decrease in test accuracy. Notice that only \method{} is able to obtain high fairness when needed, whereas the baselines cannot. If the accuracy needs to be improved more than fairness, then one can simply lower the blending parameter $\lambda$ so that AL is invoked more frequently.} The results for a neural network are in our technical report~\cite{techreport}, and the results are similar to Figure~\ref{fig:tradeoffs} where \method{} shows a better trade-off than the baselines.





Table~\ref{tbl:detailcomparison} provides a more detailed comparison with the fair AL algorithms. We report the maximum fairness score that each method can achieve when using the entire labeling budget. Note that \method{} is the only method where fairness actually improves, while the baselines do not show significant improvements in fairness compared to $Original$ and, in some cases, even have worse fairness results. For all the scenarios considered, \method{}'s maximum fairness score is 1.8--4.5x higher than the second-best results. The baselines do not perform well because they do not attempt to predict the actual label values and use them even if they have undesired labels. Thus, postponing undesired labels is critical, and it is important to find samples that are informative for fairness.


\begin{table}[t]
\small
  \centering
  \begin{tabular}{ccc|ccc}
    \toprule
    {\bf Datasets} & {\bf Fairness} & \multicolumn{4}{c}{\bf Max. Fairness Score} \\
    \midrule
    & & {\bf Original} & {\bf FAL} & $\mathbf{D}$-$\mathbf{FA^{2}L}$ & {\bf \method{}} \\
    \midrule
    \multirow{2}{*}{TravelTime} & DP & 0.212 & 0.160 & 0.237 & {\bf 0.966} \\
    \cmidrule{2-6}
    & EO & 0.182 & 0.132 & 0.214 & {\bf 0.616} \\
    \midrule
    \multirow{2}{*}{Employ} & DP & 0.248 & 0.275 & 0.252 & {\bf 0.645} \\
    \cmidrule{2-6}
    & EO & 0.165 & 0.192 & 0.181 & {\bf 0.901} \\
    \midrule
    \multirow{2}{*}{Income} & DP & 0.355  & 0.366 & 0.361 & {\bf 0.816} \\
    \cmidrule{2-6}
    & EO & 0.402 & 0.453 & 0.435 & {\bf 0.834} \\
    \midrule
    \multirow{2}{*}{COMPAS} & DP & 0.365 & 0.392 & 0.373  & {\bf 0.861} \\
    \cmidrule{2-6}
    & EO & 0.372 & 0.403 & 0.400 & {\bf 0.924} \\
    \bottomrule
  \end{tabular}
  \caption{Detailed fairness comparison of methods by tuning their hyperparameters to achieve the highest fairness scores.}
  \label{tbl:detailcomparison}
  \vspace{-0.5cm}
\end{table}

\subsection{Running Time}
\label{exp:scalability}
We compare the running time of \method{} with the baselines in Table~\ref{tbl:runtimecomparison}. For each method, we show the average end-to-end running time for all experiments in Figure~\ref{fig:tradeoffs} where the target fairness metric is DP. The end-to-end labeling process consists of model training, identifying target groups, and selecting samples. While the first two steps take similar amounts of time, the sample selection is where there are time differences. As expected, \random{} is the fastest since there is no cost for selecting samples. For all the datasets, \method{} is 1.4--10x faster than the other fair AL algorithms because \method{} does not require additional model trainings for sample selection. In contrast, \fal{} computes the expected fairness reduction over all possible labels for the top $m$ uncertain unlabeled samples, which requires $2 \times m$ additional model trainings. In addition, \decouple{} retrains the model for each sensitive group to find samples that receive conflicting predictions. 

\begin{table}[t]
\small
  \centering
  \begin{tabular}{ccc|ccc}
    \toprule
    {\bf Datasets} & \multicolumn{5}{c}{\bf Avg. Running time (sec)} \\
    \midrule
    & {\bf Entropy} & {\bf Random} & {\bf FAL} & $\mathbf{D}$-$\mathbf{FA^{2}L}$ & {\bf \method{}} \\
    \midrule
    TravelTime & 139 & 91 & 1,420 & 179 & {\bf 126} \\
    Employ & 114 & 76 & 1,411 & 140 & {\bf 98}\\
    Income & 244 & 149 & 1,965 & 290 & {\bf 205}  \\
    COMPAS & 6.1 & 5.5 & 153 & 12 & {\bf 5.9}  \\
    \bottomrule
  \end{tabular}
  \caption{Running time comparison of all methods on the four datasets using DP fairness. For each method, we show the average running time for all experiments in Figure~\ref{fig:tradeoffs}.}
  \label{tbl:runtimecomparison}
  \vspace{-0.8cm}
\end{table}

Another interesting observation is that \method{} even performs better than \entropy{}. While \entropy{} needs to calculate entropy for all unlabeled samples, \method{} computes the predicted probability for the target group only. For example, if the target subgroup is \texttt{(attribute=female, label=positive)}, then \method{} focuses on the female group instead of the entire dataset. This result indicates that the cost of managing the MAB is not significant compared to the overall labeling process, and that \method{} is practically efficient.


\subsection{Automatic Policy Search} \label{exp:policysearch}
\begin{figure}[t]
  \centering
  \begin{subfigure}{0.495\columnwidth}
     \includegraphics[width=\columnwidth]{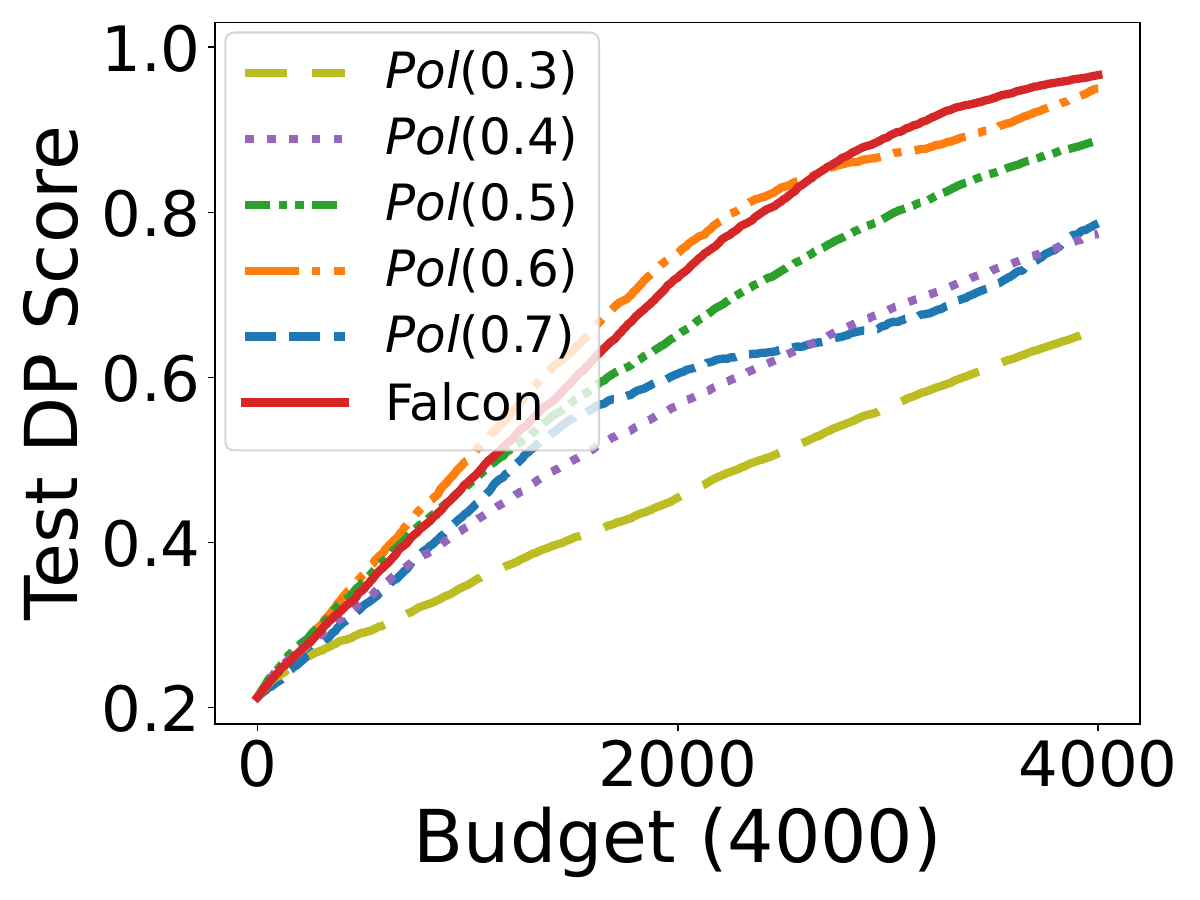}
      \vspace{-0.5cm}
     \caption{Test DP Score} \label{fig:policycomparisondp}
  \end{subfigure}
  \begin{subfigure}{0.495\columnwidth}
    \includegraphics[width=\columnwidth]{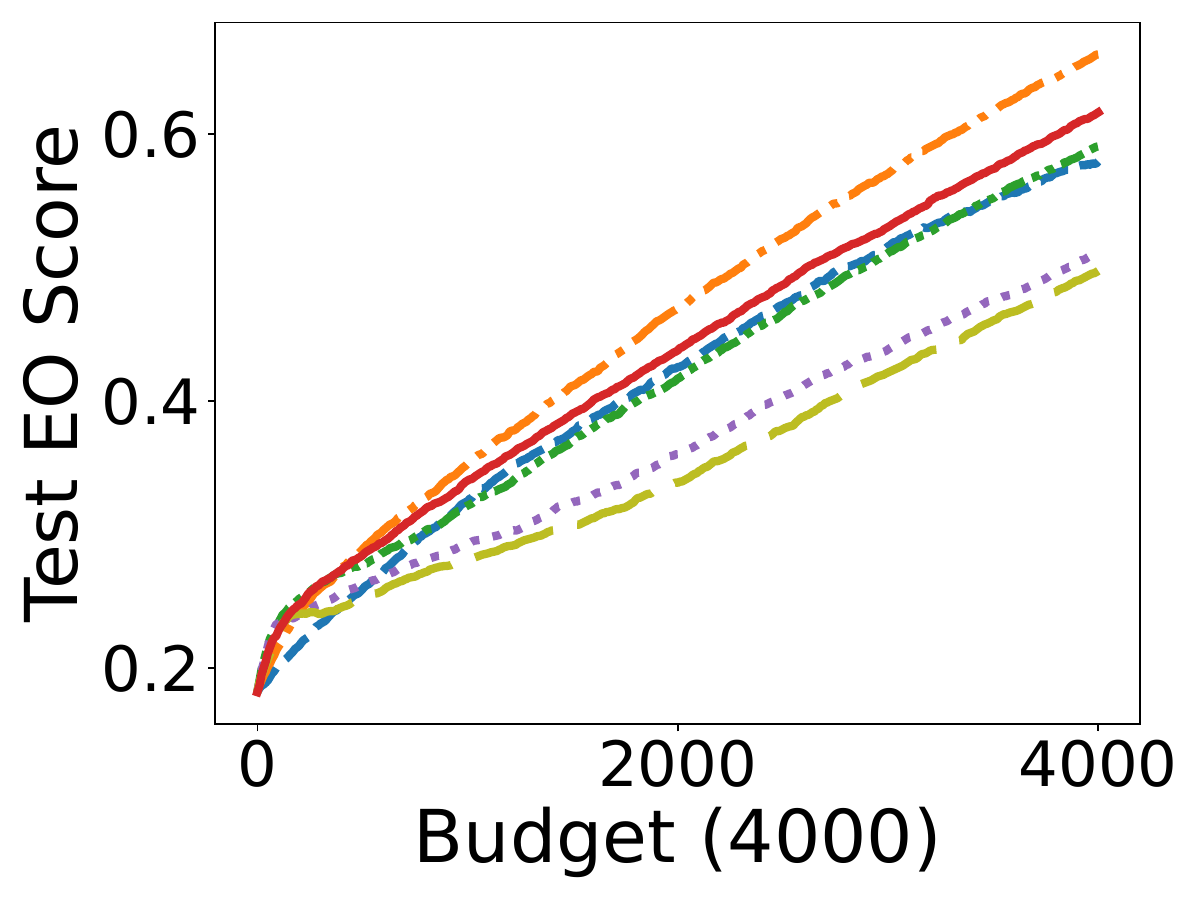}
     \vspace{-0.5cm}
     \caption{Test EO score} \label{fig:policycomparisoneo}
  \end{subfigure} 
       \vspace{-0.6cm}
     \caption{Fairness comparison of \method{} against a set of single policy baselines on the TravelTime dataset.}
     \label{fig:policycomparison}
     \vspace{-0.2cm}
\end{figure}

We compare \method{} with a set of single policy baselines to show its ability to identify the most efficient policy during the labeling process. Here a baseline method $Pol(c)$ for policy $r=c$ randomly selects a target group (if there are more than one) for each round and selects a sample whose probability for the desirable label is closest to $1-c$. Figure~\ref{fig:policycomparison} makes a comparison in terms of improved fairness on the TravelTime dataset where we use DP and EO. As a result, \method{} performs the best when using DP and the second best when using EO. \rev{Achieving the second-best performance is reasonable because \method{} has no prior knowledge of optimal policy and needs to allocate its budget to explore all policies}.

\begin{figure}[t]
  \centering
  \begin{subfigure}{0.495\columnwidth}
     \includegraphics[width=\columnwidth]{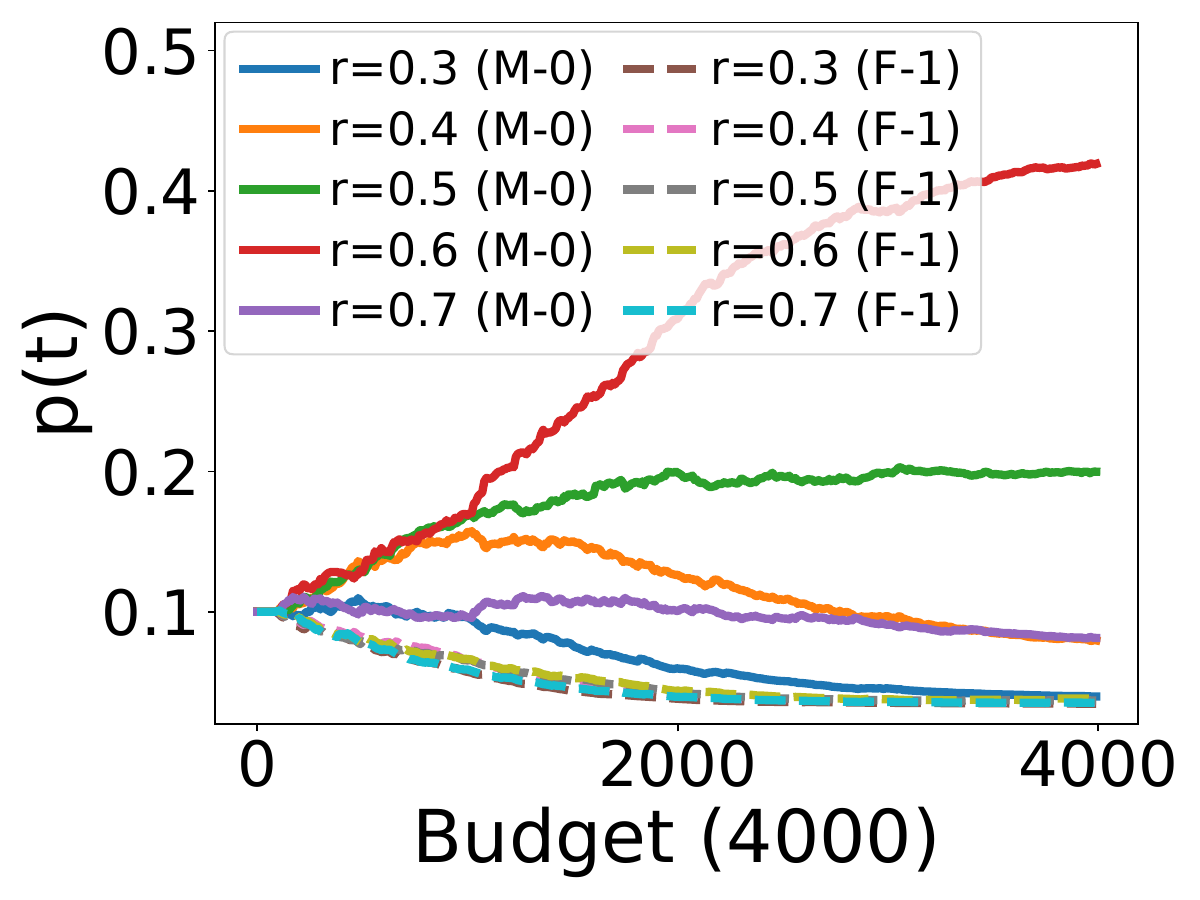}
     \vspace{-0.4cm}
     \caption{Policy Selection Probability} \label{fig:policyfairness}
  \end{subfigure}
    \begin{subfigure}{0.495\columnwidth}
    \includegraphics[width=\columnwidth]{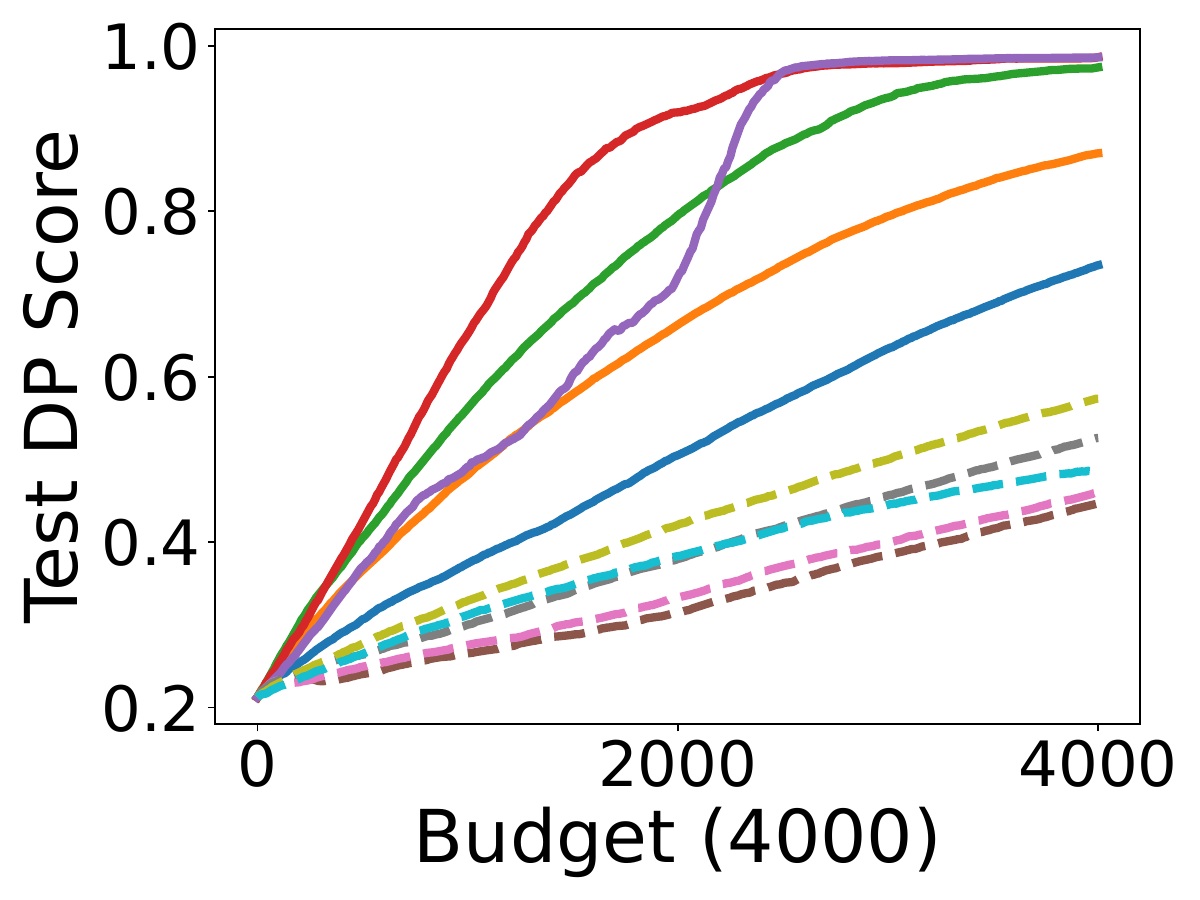}
    \vspace{-0.4cm}
     \caption{All Single Policies} 
     \label{fig:eachpolicyfairness}
  \end{subfigure} 
       \vspace{-0.6cm}
     \caption{A detailed analysis for Figure~\ref{fig:policycomparisondp}. (a) \method{} increases the selection probability of $r=0.6$ for the subgroup $(M-0)$, where we denote the sensitive attribute (Male or Female) and label of the target subgroup in parentheses. (b) Fairness improvements for all single policies. The policy $r=0.6$ for $(M-0)$ is the most effective in improving fairness.
     }
     \label{fig:policyselectiondetails}
     \vspace{-0.45cm}
\end{figure}

We analyze Figure~\ref{fig:policycomparisondp} in more detail where we show how \method{} updates its policies. Figure~\ref{fig:policyfairness} shows the selection probabilities of the policies over labeling iterations where the target group for each policy is indicated in parentheses. Here, we have two target subgroups, \texttt{(attribute=female, label=positive)} and \texttt{(attribute=male, label=negative)}, denoted as $(F-1)$ and $(M-0)$, respectively. Initially, \method{} randomly selects a policy, but starts to update the selection probabilities based on the obtained rewards. In the end, \method{} increases the probability for $r=0.6$ for $(M-0)$ the most. Figure~\ref{fig:eachpolicyfairness} shows the performance of all individual policies. We observe that $r=0.6$ for the subgroup $(M-0)$ is the most effective policy, which is consistent with \method{}'s findings. In comparison, the baseline $Pol(c)$ has a fundamental limitation as it relies on a fixed $r=c$ value for every target group.

We also perform the above experiments on the Employ dataset, and the key trends are similar where \method{} correctly updates the policies and achieves the best or second-best performance among the policies (see our technical report~\cite{techreport}).

\subsection{Varying Batch Size} \label{exp:batchsize}
We evaluate \method{} when varying the batch size $b$ from 1 to 100 when using DP fairness. Table~\ref{tbl:batchsize} shows the DP scores of different batch sizes for the four datasets. The overall trend is that a larger batch size results in less runtime because we can reduce the number of sample selection iterations, and the reward of each batch becomes more substantial. At the same time, a batch size that is too large makes it difficult to find the optimal policy due to fewer chances to update the MAB given a limited labeling budget. In our results, the best batch size depends on the datasets. For the Employ and Income datasets, a batch size of 10 is the best. For the TravelTime and COMPAS datasets, a batch size of 1 is the best where each sample has enough impact on fairness improvement, so the MAB can be updated more frequently with a small batch size. For TravelTime, however, we use a batch size of 10 in the other experiments because \method{} runs much faster without sacrificing fairness significantly.

\begin{table}[t]
\small
  \centering
  \begin{tabular}{cccccc}
    \toprule
    {\bf Datasets} & \multicolumn{5}{c}{\bf Test DP Score} \\
    \midrule
    & $\mathbf{b=1}$ & $\mathbf{b=10}$ & $\mathbf{b=20}$ & $\mathbf{b=50}$ & $\mathbf{b=100}$\\
    \midrule
    TravelTime & {\bf 0.970} & 0.966 & 0.964 & 0.939 & 0.928\\
    Employ & 0.588 & {\bf 0.645} & 0.639 & 0.631 & 0.620 \\
    Income & 0.808 & {\bf 0.816} & 0.815 & 0.807 &  0.803 \\
    COMPAS & {\bf 0.861} & 0.840 & 0.812 & 0.828 & 0.830 \\
    \bottomrule
  \end{tabular}
  \caption{Batch size impact on \method{}.
  }
  \label{tbl:batchsize}
  \vspace{-0.6cm}
\end{table}

\subsection{Varying Policies} \label{exp:numpolicies}
We vary the number of policies ($|P|$) by adjusting the spacing between the $r$ values of the two nearest policies. As discussed in Section~\ref{sec:algorithm}, we set the lower and upper bounds of the $r$ value to $0.3$ and $0.7$, respectively, to ensure that the policy set does not include extreme policies for better results. So if we use two policies, the policy set is $[r=0.3, r=0.7]$, and for five policies, it is $[r=0.3, r=0.4, r=0.5, r=0.6, r=0.7]$, which is our default set.

\begin{table}[t]
\small
  \centering
  \begin{tabular}{cccccc}
    \toprule
    {\bf Datasets} & {\bf Fairness} & \multicolumn{4}{c}{\bf Fairness Score} \\
    \midrule
    & & $\mathbf{|P|=2}$ & $\mathbf{|P|=3}$ & $\mathbf{|P|=5}$ & $\mathbf{|P|=9}$ \\
    \midrule
    \multirow{2}{*}{TravelTime} & DP & 0.956 & {\bf 0.968} & 0.966 & 0.944\\
    \cmidrule{2-6}
    & EO & 0.583 & 0.610 & {\bf 0.616} & 0.606 \\
    \midrule
    \multirow{2}{*}{Employ} & DP & 0.633 & {\bf 0.645} & {\bf 0.645} & 0.638 \\
    \cmidrule{2-6}
    & EO & 0.896 & 0.901 & 0.901 & {\bf 0.902} \\
    \bottomrule
  \end{tabular}
  \caption{Varying the number of policies.}
  \label{tbl:varyingpolicy}
  \vspace{-0.8cm}
\end{table}

Table~\ref{tbl:varyingpolicy} shows the fairness results when varying the number of policies in the range of $[2, 9]$ on the TravelTime and Employ datasets. As a result, using $3$ or $5$ policies usually yields the best performance compared to other cases. This result is expected because increasing policies initially helps to find good ones, but having too many makes it more difficult to find the good ones using a limited labeling budget. For the Employ dataset and EO, there is no significant difference between the last three options. 

\rev{We perform additional experiments using more diverse policy sets to further investigate their impact on \method{} in our technical report~\cite{techreport}. As a result, the key trends are similar to Table~\ref{tbl:varyingpolicy} where our default set outperforms others.}

\subsection{Ablation Study} \label{exp:ablation}
In Table~\ref{tbl:ablation}, we perform an ablation study to investigate the effectiveness of each component in \method{} using the TravelTime and Employ datasets. We consider the ablation scenarios of removing reward propagation, reward normalization, MAB, and trial-and-error, in that order cumulatively. 
As a result, each ablation scenario leads to worse fairness. Not propagating rewards worsens performance by reducing the chance to find the best policy. Not normalizing the rewards leads to fairness improvements that are not large enough to make a difference in the policy selection. Not using MAB is equivalent to using the simple trial-and-error algorithm in Section~\ref{sec:handlingunknown} where we can no longer search policies dynamically. \rev{The benefit of using MABs is relatively smaller than that of trial-and-error, but is still significant as shown in Figure~\ref{fig:policycomparisondp} where \method{} can achieve similar fairness as \method{} without MABs while saving up to 25\% of the labeling budget.} Finally, not using the trial-and-error strategy is equivalent to the \entropy{} method, which sometimes performs worse than {\em Original}. Thus, all functionalities are necessary.


\begin{table}[t]
\small
  \centering
  \begin{tabular}{@{\hspace{1pt}}lcccc@{\hspace{1pt}}}
    \toprule
     & \multicolumn{2}{c}{\bf TravelTime} & \multicolumn{2}{c}{\bf Employ}\\ 
    \cmidrule(lr){1-1}\cmidrule(lr){2-3} \cmidrule(lr){4-5} 
    {\bf Method} & {\bf DP} & {\bf EO} & {\bf DP} & {\bf EO} \\
    \midrule
    Original & 0.212 & 0.182 & 0.248 & 0.165 \\
    \midrule
    \method{} w/o trial-and-error \& MAB & 0.138 & 0.123 & 0.274 & 0.188 \\
    \method{} w/o MAB & 0.887 & 0.591 & 0.630 & 0.900 \\
    \method{} w/o reward norm. \& propag. & 0.863 & 0.602 & 0.614 & 0.900 \\
    \method{} w/o reward propag. & 0.959 & 0.611 &  0.635 & 0.899 \\
    \textbf{\method{}} & {\bf 0.966} & {\bf 0.616} & {\bf 0.645} & {\bf 0.901} \\
    \bottomrule
  \end{tabular}
  \caption{Ablation study of \method{}.}
  \label{tbl:ablation}
  \vspace{-0.6cm}
\end{table}




\begin{table}[t]
\small
  \centering
  \begin{tabular}{cccc|c@{\hspace{2.5pt}}c|c}
    \toprule
    {\bf Datasets} & {\bf Fair.} & \multicolumn{5}{c}{\bf Fairness Score} \\
    \midrule
    & & {\bf FAL} & $\mathbf{FAL^+}$ & $\mathbf{D}$-$\mathbf{FA^{2}L}$ & $\mathbf{D}$-$\mathbf{FA^{2}L^+}$ & {\bf \method{}} \\
    \midrule
    \multirow{2}{*}{TravelTime} & DP & 0.160 & 0.957 & 0.237 & 0.867 & {\bf 0.966}\\
    \cmidrule{2-7}
     & EO & 0.132 & 0.300 & 0.214 & 0.352 & {\bf 0.616}\\
    \midrule
    \multirow{2}{*}{Employ} & DP & 0.275 & 0.500 & 0.252 & 0.486 & {\bf 0.645}\\
    \cmidrule{2-7}
    & EO & 0.192 & 0.727 & 0.181 & 0.538 & {\bf 0.901}\\
    \bottomrule
  \end{tabular}
  \caption{Comparison of \method{} against fair AL baselines combined with trial-and-error.}
  \label{tbl:newbaselines-small}
    \vspace{-0.85cm}
\end{table}

\rev{To better understand the effectiveness of trial-and-error labeling, we also extend fair AL baselines with trial-and-error to see how they perform compared to \method{}. Table~\ref{tbl:newbaselines-small} shows the fairness scores of different methods, where we combine \fal{} and \decouple{} with trial-and-error and refer to them as \fal$^{+}$ and \decouple$^{+}$, respectively. As a result, we observe trial-and-error sampling significantly improves the fairness of baselines. However, \method{} still consistently outperforms all baselines, which highlights that the MAB components of \method{} are also important for optimizing fairness. The results for other datasets are similar and shown in our technical report~\cite{techreport}.}

\subsection{\rev{Other Adversarial MAB Algorithms}} \label{exp:mabs}
\rev{We now evaluate the empirical performance of EXP3 compared to other adversarial MABs, EXP3-IX~\cite{lattimore2020} and EXP4.P~\cite{DBLP:journals/jmlr/BeygelzimerLLRS11}, which are designed to achieve a high probability regret bound by reducing the variance of EXP3. Table~\ref{tbl:moremabmethods} below shows the fairness results when using different adversarial MAB methods on the TravelTime and Employ datasets. The results show that there is no single best algorithm, and EXP3 empirically works as well as other methods. Even if EXP3 is not always the best, its fairness score is usually very close to the best score. We make similar observations in other datasets as well (see our technical report~\cite{techreport}).}

\begin{table}[t]
\small
  \centering
  \begin{tabular}{ccc|ccc}
    \toprule
    {\bf Datasets} & {\bf Fairness} & \multicolumn{4}{c}{\bf Fairness Score} \\
    \midrule
    & & {\bf Original} & $\mathbf{EXP3}$ & $\mathbf{EXP3}$-$\mathbf{IX}$ & $\mathbf{EXP4.P}$\\
    \midrule
    \multirow{2}{*}{TravelTime} & DP & 0.212 & 0.966 & {\bf 0.970} & 0.936 \\
    \cmidrule{2-6}
     & EO & 0.182 & {\bf 0.616} & 0.610 & 0.609 \\
    \midrule
    \multirow{2}{*}{Employ} & DP & 0.248 & 0.645 & {\bf 0.651} & 0.628 \\
    \cmidrule{2-6}
    & EO & 0.165 & 0.901 & 0.901 & {\bf 0.902} \\
    \bottomrule
  \end{tabular}
  \caption{Fairness of \method{} with other adversarial MABs.}
  \label{tbl:moremabmethods}
  \vspace{-0.8cm}
\end{table}

\section{Related Work}

\paragraph{\rev{Data-centric AI}}
\rev{Data-centric AI techniques boost various ML performances by improving the training data through better data management~\cite{ratner2017snorkel, krishnan2017boostclean, 10.14778/3611540.3611606}. Our work falls into the category of data-centric AI for better model fairness~\cite{pfr2019, salimi2019interventional, DBLP:conf/sigmod/ZhangCAN21, zhangiflipper2023} by proposing a novel data labeling mechanism.}

\paragraph{AI Fairness}

Conventional fairness techniques for fair training can be largely categorized into pre-processing, in-processing, and post-processing techniques. \method{} can be categorized as a pre-processing approach and assumes that data labels are not available. Related works are OmniFair~\cite{DBLP:conf/sigmod/ZhangCAN21} and FairBatch~\cite{DBLP:conf/iclr/Roh0WS21}, which improve group fairness using sample weighting. \rev{In comparison, \method{} focuses on the data labeling problem, which has quite different issues as the ground truth labels are unavailable. In this setup, trial-and-error sampling allows us to apply the subgroup sampling strategy. However, prioritizing only fairness-informative samples results in excessive postponing of undesired samples, wasting the limited labeling budget. To navigate this trade-off, \method{} leverages adversarial MABs, effectively balancing the informativeness of samples and the postpone rate by dynamically selecting various policies.}



\paragraph{Fair Active Learning}

A recent line of work addresses the fair active learning problem. FAL\,\cite{DBLP:journals/eswa/AnahidehAT22} extends conventional active learning to estimate the resulting fairness of selecting a sample. The estimation is a probabilistic analysis assuming that the label can have any of the possible values with equal probability. PANDA\,\cite{DBLP:conf/fat/SharafDN22} uses reinforcement learning for the selection, but this can be expensive. 
Another related line of work is resolving class imbalance~\cite{imbalanceactivelearning1, imbalanceactivelearning2}. However, an implicit assumption here is that there is a fixed underrepresented group whose model accuracy needs to be improved. In contrast, improving fairness is more complicated where the groups that need to be improved may change as we label more data.

\paragraph{Fair Adaptive Sampling}

Another line of research is sampling training data for the purpose of improving minimax fairness\,\cite{DBLP:conf/nips/ShekharFGJ21, abernethy2022ActiveSampling}, which takes samples from the group that has the worst model's accuracy. Here the group is defined only as a combination of sensitive attributes. In comparison, \method{} supports any group fairness measure beyond minimax fairness and addresses the more challenging problem where the target group can be defined using labels. 



\paragraph{Active Learning}

Active learning\,\cite{Lewis:1994:SAT:188490.188495,Seung:1992:QC:130385.130417,DBLP:series/synthesis/2012Settles,Abe:1998:QLS:645527.657478,Settles:2008:AAL:1613715.1613855,Settles:2007:MAL:2981562.2981724,Roy:2001:TOA:645530.655646} has been studied for decades for efficient labeling. A standard approach is sampling based on uncertainty, e.g., least confidence~\cite{confidence2014wang} or highest entropy~\cite{entropy2001shannon}. Diversity-based methods have also been proposed to find representative samples using a clustering~\cite{10.1145/1015330.1015349} or coreset selection~\cite{sener2018active}. Hybrid approaches~\cite{6751135, ash2020deep} that combine both criteria have been studied as well.
A conceptually close work to our framework is active learning by learning\,\cite{DBLP:conf/aaai/HsuL15}, which selects the best labeling policy among the set of predefined AL algorithms. In comparison, we solve the new problem of improving fairness in active learning by learning the best self-generated policies.


\paragraph{Adversarial MAB}

Adversarial MABs are used to choose policies only based on rewards where the rewards can be any value. As explained in Section~\ref{sec:adversarialmab}, EXP3\,\cite{DBLP:journals/siamcomp/AuerCFS02} chooses arms probabilistically and updates the probabilities based on the rewards. \rev{EXP3-IX~\cite{lattimore2020} reduces the variance of EXP3 at the price of introducing bias. EXP4.P\,\cite{DBLP:journals/jmlr/BeygelzimerLLRS11} improves the performance by accepting expert advice on which arms are more promising.} \rev{MABs have recently been widely used in data-centric AI works like data charging~\cite{DBLP:journals/pvldb/ChaiLTLL22} and entity augmentation~\cite{10.14778/3611479.3611535}, as well as in visualization recommendation~\cite{vartak2015seedb} and video database management systems~\cite{10.14778/3598581.3598599}.} \method{} can be compatible with any adversarial MAB algorithm.




\section{Conclusion}

We propose the fair active learning framework \method{}, which selects labels to improve model fairness and accuracy by learning policies. \method{} determines which groups of data need to be labeled first for better fairness and uses a trial-and-error strategy to find labeled data of the groups. \method{} also learns policies to determine how much risk to take in selecting fairness-informative samples, which can be done with adversarial MAB methods. The sample selection for fairness is blended with active learning methods to also improve accuracy. \method{} is efficient and requires much fewer model trainings than other fair active learning approaches. Experiments show how \method{} drastically outperforms the state-of-the-art fair active learning baselines on real benchmark datasets. \method{} is the only method that supports a proper trade-off between fairness and accuracy where its maximum fairness score is 1.8--4.5x higher than the second-best results while being more efficient. 

\begin{acks}
This work was supported by a Google Research Award, by the Institute of Information \& Communications Technology Planning \& Evaluation(IITP) grant funded by the Korea government(MSIT) (No.\@ 2022-0-00157, Robust, Fair, Extensible Data-Centric Continual Learning), and by the National Research Foundation of Korea(NRF) grant funded by the Korea government(MSIT) (No.\@ NRF-2022R1A2C2004382). 
\end{acks}

\bibliographystyle{ACM-Reference-Format}
\bibliography{main}


\begin{thebibliography}{70}


\ifx \showCODEN    \undefined \def \showCODEN     #1{\unskip}     \fi
\ifx \showDOI      \undefined \def \showDOI       #1{#1}\fi
\ifx \showISBNx    \undefined \def \showISBNx     #1{\unskip}     \fi
\ifx \showISBNxiii \undefined \def \showISBNxiii  #1{\unskip}     \fi
\ifx \showISSN     \undefined \def \showISSN      #1{\unskip}     \fi
\ifx \showLCCN     \undefined \def \showLCCN      #1{\unskip}     \fi
\ifx \shownote     \undefined \def \shownote      #1{#1}          \fi
\ifx \showarticletitle \undefined \def \showarticletitle #1{#1}   \fi
\ifx \showURL      \undefined \def \showURL       {\relax}        \fi
\providecommand\bibfield[2]{#2}
\providecommand\bibinfo[2]{#2}
\providecommand\natexlab[1]{#1}
\providecommand\showeprint[2][]{arXiv:#2}

\bibitem[\protect\citeauthoryear{Abe and Mamitsuka}{Abe and Mamitsuka}{1998}]%
        {Abe:1998:QLS:645527.657478}
\bibfield{author}{\bibinfo{person}{Naoki Abe} {and} \bibinfo{person}{Hiroshi
  Mamitsuka}.} \bibinfo{year}{1998}\natexlab{}.
\newblock \showarticletitle{Query Learning Strategies Using Boosting and
  Bagging}. In \bibinfo{booktitle}{\emph{ICML}}. \bibinfo{address}{San
  Francisco, CA, USA}, \bibinfo{pages}{1--9}.
\newblock
\showISBNx{1-55860-556-8}


\bibitem[\protect\citeauthoryear{Abernethy, Awasthi, Kleindessner, Morgenstern,
  Russell, and Zhang}{Abernethy et~al\mbox{.}}{2022a}]%
        {abernethy2022ActiveSampling}
\bibfield{author}{\bibinfo{person}{Jacob Abernethy}, \bibinfo{person}{Pranjal
  Awasthi}, \bibinfo{person}{Matthäus Kleindessner}, \bibinfo{person}{Jamie
  Morgenstern}, \bibinfo{person}{Chris Russell}, {and} \bibinfo{person}{Jie
  Zhang}.} \bibinfo{year}{2022}\natexlab{a}.
\newblock \showarticletitle{Active Sampling for Min-Max Fairness}. In
  \bibinfo{booktitle}{\emph{ICML}}.
\newblock


\bibitem[\protect\citeauthoryear{Abernethy, Awasthi, Kleindessner, Morgenstern,
  Russell, and Zhang}{Abernethy et~al\mbox{.}}{2022b}]%
        {pmlr-v162-abernethy22a}
\bibfield{author}{\bibinfo{person}{Jacob~D. Abernethy},
  \bibinfo{person}{Pranjal Awasthi}, \bibinfo{person}{Matth{\"{a}}us
  Kleindessner}, \bibinfo{person}{Jamie Morgenstern}, \bibinfo{person}{Chris
  Russell}, {and} \bibinfo{person}{Jie Zhang}.}
  \bibinfo{year}{2022}\natexlab{b}.
\newblock \showarticletitle{Active Sampling for Min-Max Fairness}. In
  \bibinfo{booktitle}{\emph{ICML}}. \bibinfo{pages}{53--65}.
\newblock


\bibitem[\protect\citeauthoryear{Aggarwal, Popescu, and Hudelot}{Aggarwal
  et~al\mbox{.}}{2020}]%
        {imbalanceactivelearning1}
\bibfield{author}{\bibinfo{person}{Umang Aggarwal}, \bibinfo{person}{Adrian
  Popescu}, {and} \bibinfo{person}{Céline Hudelot}.}
  \bibinfo{year}{2020}\natexlab{}.
\newblock \showarticletitle{Active Learning for Imbalanced Datasets}. In
  \bibinfo{booktitle}{\emph{WACV}}. \bibinfo{pages}{1417--1426}.
\newblock


\bibitem[\protect\citeauthoryear{Aggarwal, Popescu, and Hudelot}{Aggarwal
  et~al\mbox{.}}{2021}]%
        {imbalanceactivelearning2}
\bibfield{author}{\bibinfo{person}{Umang Aggarwal}, \bibinfo{person}{Adrian
  Popescu}, {and} \bibinfo{person}{Céline Hudelot}.}
  \bibinfo{year}{2021}\natexlab{}.
\newblock \showarticletitle{Minority Class Oriented Active Learning for
  Imbalanced Datasets}. In \bibinfo{booktitle}{\emph{ICPR}}.
  \bibinfo{pages}{9920--9927}.
\newblock


\bibitem[\protect\citeauthoryear{Anahideh, Asudeh, and
  Thirumuruganathan}{Anahideh et~al\mbox{.}}{2022}]%
        {DBLP:journals/eswa/AnahidehAT22}
\bibfield{author}{\bibinfo{person}{Hadis Anahideh}, \bibinfo{person}{Abolfazl
  Asudeh}, {and} \bibinfo{person}{Saravanan Thirumuruganathan}.}
  \bibinfo{year}{2022}\natexlab{}.
\newblock \showarticletitle{Fair active learning}.
\newblock \bibinfo{journal}{\emph{Expert Systems with Applications}}
  \bibinfo{volume}{199} (\bibinfo{year}{2022}), \bibinfo{pages}{116981}.
\newblock


\bibitem[\protect\citeauthoryear{Angwin, Larson, Mattu, and Kirchner}{Angwin
  et~al\mbox{.}}{2016}]%
        {machinebias}
\bibfield{author}{\bibinfo{person}{J. Angwin}, \bibinfo{person}{J. Larson},
  \bibinfo{person}{S. Mattu}, {and} \bibinfo{person}{L. Kirchner}.}
  \bibinfo{year}{2016}\natexlab{}.
\newblock \bibinfo{title}{Machine bias: {T}here's software used across the
  country to predict future criminals. {A}nd its biased against blacks.}
\newblock
\newblock


\bibitem[\protect\citeauthoryear{Ash, Zhang, Krishnamurthy, Langford, and
  Agarwal}{Ash et~al\mbox{.}}{2020}]%
        {ash2020deep}
\bibfield{author}{\bibinfo{person}{Jordan~T. Ash}, \bibinfo{person}{Chicheng
  Zhang}, \bibinfo{person}{Akshay Krishnamurthy}, \bibinfo{person}{John
  Langford}, {and} \bibinfo{person}{Alekh Agarwal}.}
  \bibinfo{year}{2020}\natexlab{}.
\newblock \showarticletitle{Deep Batch Active Learning by Diverse, Uncertain
  Gradient Lower Bounds}. In \bibinfo{booktitle}{\emph{ICLR}}.
\newblock


\bibitem[\protect\citeauthoryear{Auer, Cesa-Bianchi, and Fischer}{Auer
  et~al\mbox{.}}{2002a}]%
        {10.1023/A:1013689704352}
\bibfield{author}{\bibinfo{person}{Peter Auer}, \bibinfo{person}{Nicol\`{o}
  Cesa-Bianchi}, {and} \bibinfo{person}{Paul Fischer}.}
  \bibinfo{year}{2002}\natexlab{a}.
\newblock \showarticletitle{Finite-Time Analysis of the Multiarmed Bandit
  Problem}.
\newblock  \bibinfo{volume}{47}, \bibinfo{number}{2--3} (\bibinfo{year}{2002}),
  \bibinfo{pages}{235--256}.
\newblock
\showISSN{0885-6125}


\bibitem[\protect\citeauthoryear{Auer, Cesa{-}Bianchi, Freund, and
  Schapire}{Auer et~al\mbox{.}}{2002b}]%
        {DBLP:journals/siamcomp/AuerCFS02}
\bibfield{author}{\bibinfo{person}{Peter Auer}, \bibinfo{person}{Nicol{\`{o}}
  Cesa{-}Bianchi}, \bibinfo{person}{Yoav Freund}, {and}
  \bibinfo{person}{Robert~E. Schapire}.} \bibinfo{year}{2002}\natexlab{b}.
\newblock \showarticletitle{The Nonstochastic Multiarmed Bandit Problem}.
\newblock \bibinfo{journal}{\emph{{SIAM} J. Comput.}} \bibinfo{volume}{32},
  \bibinfo{number}{1} (\bibinfo{year}{2002}), \bibinfo{pages}{48--77}.
\newblock


\bibitem[\protect\citeauthoryear{Bang, Kakkar, Chunduri, Mitra, and
  Arulraj}{Bang et~al\mbox{.}}{2023}]%
        {10.14778/3598581.3598599}
\bibfield{author}{\bibinfo{person}{Jaeho Bang}, \bibinfo{person}{Gaurav~Tarlok
  Kakkar}, \bibinfo{person}{Pramod Chunduri}, \bibinfo{person}{Subrata Mitra},
  {and} \bibinfo{person}{Joy Arulraj}.} \bibinfo{year}{2023}\natexlab{}.
\newblock \showarticletitle{Seiden: Revisiting Query Processing in Video
  Database Systems}.
\newblock \bibinfo{journal}{\emph{Proc. VLDB Endow.}} \bibinfo{volume}{16},
  \bibinfo{number}{9} (\bibinfo{date}{may} \bibinfo{year}{2023}),
  \bibinfo{pages}{2289–2301}.
\newblock
\showISSN{2150-8097}


\bibitem[\protect\citeauthoryear{Barocas, Hardt, and Narayanan}{Barocas
  et~al\mbox{.}}{2019}]%
        {barocas-hardt-narayanan}
\bibfield{author}{\bibinfo{person}{Solon Barocas}, \bibinfo{person}{Moritz
  Hardt}, {and} \bibinfo{person}{Arvind Narayanan}.}
  \bibinfo{year}{2019}\natexlab{}.
\newblock \bibinfo{booktitle}{\emph{Fairness and Machine Learning: Limitations
  and Opportunities}}.
\newblock \bibinfo{publisher}{fairmlbook.org}.
\newblock
\newblock
\shownote{\url{http://www.fairmlbook.org}.}


\bibitem[\protect\citeauthoryear{Bellamy, Dey, Hind, Hoffman, Houde, Kannan,
  Lohia, Martino, Mehta, Mojsilovic, Nagar, Ramamurthy, Richards, Saha,
  Sattigeri, Singh, Varshney, and Zhang}{Bellamy et~al\mbox{.}}{2018}]%
        {aif360}
\bibfield{author}{\bibinfo{person}{Rachel K.~E. Bellamy},
  \bibinfo{person}{Kuntal Dey}, \bibinfo{person}{Michael Hind},
  \bibinfo{person}{Samuel~C. Hoffman}, \bibinfo{person}{Stephanie Houde},
  \bibinfo{person}{Kalapriya Kannan}, \bibinfo{person}{Pranay Lohia},
  \bibinfo{person}{Jacquelyn Martino}, \bibinfo{person}{Sameep Mehta},
  \bibinfo{person}{Aleksandra Mojsilovic}, \bibinfo{person}{Seema Nagar},
  \bibinfo{person}{Karthikeyan~Natesan Ramamurthy}, \bibinfo{person}{John
  Richards}, \bibinfo{person}{Diptikalyan Saha}, \bibinfo{person}{Prasanna
  Sattigeri}, \bibinfo{person}{Moninder Singh}, \bibinfo{person}{Kush~R.
  Varshney}, {and} \bibinfo{person}{Yunfeng Zhang}.}
  \bibinfo{year}{2018}\natexlab{}.
\newblock \bibinfo{title}{{AI Fairness} 360: An Extensible Toolkit for
  Detecting, Understanding, and Mitigating Unwanted Algorithmic Bias}.
\newblock
\newblock


\bibitem[\protect\citeauthoryear{Beygelzimer, Langford, Li, Reyzin, and
  Schapire}{Beygelzimer et~al\mbox{.}}{2011}]%
        {DBLP:journals/jmlr/BeygelzimerLLRS11}
\bibfield{author}{\bibinfo{person}{Alina Beygelzimer}, \bibinfo{person}{John
  Langford}, \bibinfo{person}{Lihong Li}, \bibinfo{person}{Lev Reyzin}, {and}
  \bibinfo{person}{Robert~E. Schapire}.} \bibinfo{year}{2011}\natexlab{}.
\newblock \showarticletitle{Contextual Bandit Algorithms with Supervised
  Learning Guarantees}. In \bibinfo{booktitle}{\emph{AISTATS}}
  \emph{(\bibinfo{series}{{JMLR} Proceedings})}, Vol.~\bibinfo{volume}{15}.
  \bibinfo{pages}{19--26}.
\newblock


\bibitem[\protect\citeauthoryear{Bird, Dud{\'i}k, Edgar, Horn, Lutz, Milan,
  Sameki, Wallach, and Walker}{Bird et~al\mbox{.}}{2020}]%
        {bird2020fairlearn}
\bibfield{author}{\bibinfo{person}{Sarah Bird}, \bibinfo{person}{Miro
  Dud{\'i}k}, \bibinfo{person}{Richard Edgar}, \bibinfo{person}{Brandon Horn},
  \bibinfo{person}{Roman Lutz}, \bibinfo{person}{Vanessa Milan},
  \bibinfo{person}{Mehrnoosh Sameki}, \bibinfo{person}{Hanna Wallach}, {and}
  \bibinfo{person}{Kathleen Walker}.} \bibinfo{year}{2020}\natexlab{}.
\newblock \bibinfo{booktitle}{\emph{Fairlearn: A toolkit for assessing and
  improving fairness in {AI}}}.
\newblock \bibinfo{type}{{T}echnical {R}eport} MSR-TR-2020-32.
  \bibinfo{institution}{Microsoft}.
\newblock


\bibitem[\protect\citeauthoryear{Buss, Mousavi, Tokarev, Termehchy, Maier, and
  Lee}{Buss et~al\mbox{.}}{2023}]%
        {10.14778/3611479.3611535}
\bibfield{author}{\bibinfo{person}{Christopher Buss}, \bibinfo{person}{Jasmin
  Mousavi}, \bibinfo{person}{Mikhail Tokarev}, \bibinfo{person}{Arash
  Termehchy}, \bibinfo{person}{David Maier}, {and} \bibinfo{person}{Stefan
  Lee}.} \bibinfo{year}{2023}\natexlab{}.
\newblock \showarticletitle{Effective Entity Augmentation by Querying External
  Data Sources}.
\newblock \bibinfo{journal}{\emph{Proc. VLDB Endow.}} \bibinfo{volume}{16},
  \bibinfo{number}{11} (\bibinfo{date}{jul} \bibinfo{year}{2023}),
  \bibinfo{pages}{3404–3417}.
\newblock
\showISSN{2150-8097}


\bibitem[\protect\citeauthoryear{Cao and Lan}{Cao and Lan}{2022}]%
        {cao2022decouple}
\bibfield{author}{\bibinfo{person}{Yiting Cao} {and} \bibinfo{person}{Chao
  Lan}.} \bibinfo{year}{2022}\natexlab{}.
\newblock \showarticletitle{Fairness-Aware Active Learning for Decoupled
  Model}. In \bibinfo{booktitle}{\emph{IJCNN}}. \bibinfo{pages}{1--9}.
\newblock


\bibitem[\protect\citeauthoryear{Chai, Liu, Tang, Li, and Luo}{Chai
  et~al\mbox{.}}{2022}]%
        {DBLP:journals/pvldb/ChaiLTLL22}
\bibfield{author}{\bibinfo{person}{Chengliang Chai}, \bibinfo{person}{Jiabin
  Liu}, \bibinfo{person}{Nan Tang}, \bibinfo{person}{Guoliang Li}, {and}
  \bibinfo{person}{Yuyu Luo}.} \bibinfo{year}{2022}\natexlab{}.
\newblock \showarticletitle{Selective Data Acquisition in the Wild for Model
  Charging}.
\newblock \bibinfo{journal}{\emph{Proc. {VLDB} Endow.}} \bibinfo{volume}{15},
  \bibinfo{number}{7} (\bibinfo{year}{2022}), \bibinfo{pages}{1466--1478}.
\newblock


\bibitem[\protect\citeauthoryear{Chen, Johansson, and Sontag}{Chen
  et~al\mbox{.}}{2018}]%
        {DBLP:conf/nips/ChenJS18}
\bibfield{author}{\bibinfo{person}{Irene~Y. Chen}, \bibinfo{person}{Fredrik~D.
  Johansson}, {and} \bibinfo{person}{David~A. Sontag}.}
  \bibinfo{year}{2018}\natexlab{}.
\newblock \showarticletitle{Why Is My Classifier Discriminatory?}. In
  \bibinfo{booktitle}{\emph{NeurIPS}}. \bibinfo{pages}{3543--3554}.
\newblock


\bibitem[\protect\citeauthoryear{Chouldechova}{Chouldechova}{2017}]%
        {DBLP:journals/bigdata/Chouldechova17}
\bibfield{author}{\bibinfo{person}{Alexandra Chouldechova}.}
  \bibinfo{year}{2017}\natexlab{}.
\newblock \showarticletitle{Fair Prediction with Disparate Impact: {A} Study of
  Bias in Recidivism Prediction Instruments}.
\newblock \bibinfo{journal}{\emph{Big Data}} \bibinfo{volume}{5},
  \bibinfo{number}{2} (\bibinfo{year}{2017}), \bibinfo{pages}{153--163}.
\newblock


\bibitem[\protect\citeauthoryear{Dastin}{Dastin}{2018}]%
        {dastin2018amazon}
\bibfield{author}{\bibinfo{person}{Jeffrey Dastin}.}
  \bibinfo{year}{2018}\natexlab{}.
\newblock \bibinfo{title}{Amazon scraps secret AI recruiting tool that showed
  bias against women. Reuters (2018)}.
\newblock
\newblock


\bibitem[\protect\citeauthoryear{Ding, Hardt, Miller, and Schmidt}{Ding
  et~al\mbox{.}}{2021}]%
        {ding2021retiring}
\bibfield{author}{\bibinfo{person}{Frances Ding}, \bibinfo{person}{Moritz
  Hardt}, \bibinfo{person}{John Miller}, {and} \bibinfo{person}{Ludwig
  Schmidt}.} \bibinfo{year}{2021}\natexlab{}.
\newblock \showarticletitle{Retiring Adult: New Datasets for Fair Machine
  Learning}. In \bibinfo{booktitle}{\emph{NeurIPS}}.
  \bibinfo{pages}{6478--6490}.
\newblock


\bibitem[\protect\citeauthoryear{Dutta, Wei, Yueksel, Chen, Liu, and
  Varshney}{Dutta et~al\mbox{.}}{2020}]%
        {DBLP:conf/icml/DuttaWYC0V20}
\bibfield{author}{\bibinfo{person}{Sanghamitra Dutta}, \bibinfo{person}{Dennis
  Wei}, \bibinfo{person}{Hazar Yueksel}, \bibinfo{person}{Pin{-}Yu Chen},
  \bibinfo{person}{Sijia Liu}, {and} \bibinfo{person}{Kush~R. Varshney}.}
  \bibinfo{year}{2020}\natexlab{}.
\newblock \showarticletitle{Is There a Trade-Off Between Fairness and Accuracy?
  {A} Perspective Using Mismatched Hypothesis Testing}. In
  \bibinfo{booktitle}{\emph{ICML}}, Vol.~\bibinfo{volume}{119}.
  \bibinfo{pages}{2803--2813}.
\newblock


\bibitem[\protect\citeauthoryear{Elhamifar, Sapiro, Yang, and Sastry}{Elhamifar
  et~al\mbox{.}}{2013}]%
        {6751135}
\bibfield{author}{\bibinfo{person}{Ehsan Elhamifar}, \bibinfo{person}{Guillermo
  Sapiro}, \bibinfo{person}{Allen~Y. Yang}, {and} \bibinfo{person}{S.~Shankar
  Sastry}.} \bibinfo{year}{2013}\natexlab{}.
\newblock \showarticletitle{A Convex Optimization Framework for Active
  Learning}. In \bibinfo{booktitle}{\emph{ICCV}}. \bibinfo{pages}{209--216}.
\newblock


\bibitem[\protect\citeauthoryear{Esfandiari, Karbasi, Mehrabian, and
  Mirrokni}{Esfandiari et~al\mbox{.}}{2021}]%
        {batchbandit}
\bibfield{author}{\bibinfo{person}{Hossein Esfandiari}, \bibinfo{person}{Amin
  Karbasi}, \bibinfo{person}{Abbas Mehrabian}, {and} \bibinfo{person}{Vahab
  Mirrokni}.} \bibinfo{year}{2021}\natexlab{}.
\newblock \showarticletitle{Regret Bounds for Batched Bandits}.
\newblock  \bibinfo{volume}{35}, \bibinfo{number}{8} (\bibinfo{date}{May}
  \bibinfo{year}{2021}), \bibinfo{pages}{7340--7348}.
\newblock


\bibitem[\protect\citeauthoryear{Feldman, Friedler, Moeller, Scheidegger, and
  Venkatasubramanian}{Feldman et~al\mbox{.}}{2015}]%
        {DBLP:conf/kdd/FeldmanFMSV15}
\bibfield{author}{\bibinfo{person}{Michael Feldman},
  \bibinfo{person}{Sorelle~A. Friedler}, \bibinfo{person}{John Moeller},
  \bibinfo{person}{Carlos Scheidegger}, {and} \bibinfo{person}{Suresh
  Venkatasubramanian}.} \bibinfo{year}{2015}\natexlab{}.
\newblock \showarticletitle{Certifying and Removing Disparate Impact}. In
  \bibinfo{booktitle}{\emph{KDD}}. \bibinfo{pages}{259--268}.
\newblock


\bibitem[\protect\citeauthoryear{Grafberger, Guha, Groth, and
  Schelter}{Grafberger et~al\mbox{.}}{2023}]%
        {10.14778/3611540.3611606}
\bibfield{author}{\bibinfo{person}{Stefan Grafberger}, \bibinfo{person}{Shubha
  Guha}, \bibinfo{person}{Paul Groth}, {and} \bibinfo{person}{Sebastian
  Schelter}.} \bibinfo{year}{2023}\natexlab{}.
\newblock \showarticletitle{Mlwhatif: What If You Could Stop Re-Implementing
  Your Machine Learning Pipeline Analyses over and Over?}
\newblock \bibinfo{journal}{\emph{Proc. VLDB Endow.}} \bibinfo{volume}{16},
  \bibinfo{number}{12} (\bibinfo{date}{aug} \bibinfo{year}{2023}),
  \bibinfo{pages}{4002–4005}.
\newblock
\showISSN{2150-8097}


\bibitem[\protect\citeauthoryear{Hardt, Price, and Srebro}{Hardt
  et~al\mbox{.}}{2016}]%
        {DBLP:conf/nips/HardtPNS16}
\bibfield{author}{\bibinfo{person}{Moritz Hardt}, \bibinfo{person}{Eric Price},
  {and} \bibinfo{person}{Nati Srebro}.} \bibinfo{year}{2016}\natexlab{}.
\newblock \showarticletitle{Equality of Opportunity in Supervised Learning}. In
  \bibinfo{booktitle}{\emph{NeurIPS}}. \bibinfo{pages}{3315--3323}.
\newblock


\bibitem[\protect\citeauthoryear{Hsu and Lin}{Hsu and Lin}{2015}]%
        {DBLP:conf/aaai/HsuL15}
\bibfield{author}{\bibinfo{person}{Wei{-}Ning Hsu} {and}
  \bibinfo{person}{Hsuan{-}Tien Lin}.} \bibinfo{year}{2015}\natexlab{}.
\newblock \showarticletitle{Active Learning by Learning}. In
  \bibinfo{booktitle}{\emph{AAAI}}. \bibinfo{pages}{2659--2665}.
\newblock


\bibitem[\protect\citeauthoryear{Ilyas and Rekatsinas}{Ilyas and
  Rekatsinas}{2022}]%
        {DBLP:journals/jdiq/IlyasR22}
\bibfield{author}{\bibinfo{person}{Ihab~F. Ilyas} {and}
  \bibinfo{person}{Theodoros Rekatsinas}.} \bibinfo{year}{2022}\natexlab{}.
\newblock \showarticletitle{Machine Learning and Data Cleaning: Which Serves
  the Other?}
\newblock \bibinfo{journal}{\emph{{ACM} J. Data Inf. Qual.}}
  \bibinfo{volume}{14}, \bibinfo{number}{3} (\bibinfo{year}{2022}),
  \bibinfo{pages}{13:1--13:11}.
\newblock


\bibitem[\protect\citeauthoryear{Iosifidis and Ntoutsi}{Iosifidis and
  Ntoutsi}{2019}]%
        {DBLP:conf/cikm/IosifidisN19}
\bibfield{author}{\bibinfo{person}{Vasileios Iosifidis} {and}
  \bibinfo{person}{Eirini Ntoutsi}.} \bibinfo{year}{2019}\natexlab{}.
\newblock \showarticletitle{AdaFair: Cumulative Fairness Adaptive Boosting}. In
  \bibinfo{booktitle}{\emph{CIKM}}. \bibinfo{publisher}{{ACM}},
  \bibinfo{pages}{781--790}.
\newblock


\bibitem[\protect\citeauthoryear{Kamiran and Calders}{Kamiran and
  Calders}{2011}]%
        {DBLP:journals/kais/KamiranC11}
\bibfield{author}{\bibinfo{person}{Faisal Kamiran} {and} \bibinfo{person}{Toon
  Calders}.} \bibinfo{year}{2011}\natexlab{}.
\newblock \showarticletitle{Data preprocessing techniques for classification
  without discrimination}.
\newblock \bibinfo{journal}{\emph{Knowl. Inf. Syst.}} \bibinfo{volume}{33},
  \bibinfo{number}{1} (\bibinfo{year}{2011}), \bibinfo{pages}{1--33}.
\newblock


\bibitem[\protect\citeauthoryear{Kim, Vojnovic, and Yun}{Kim
  et~al\mbox{.}}{2022}]%
        {pmlr-v162-kim22j}
\bibfield{author}{\bibinfo{person}{Jung-Hun Kim}, \bibinfo{person}{Milan
  Vojnovic}, {and} \bibinfo{person}{Se-Young Yun}.}
  \bibinfo{year}{2022}\natexlab{}.
\newblock \showarticletitle{Rotting Infinitely Many-Armed Bandits}. In
  \bibinfo{booktitle}{\emph{ICML}} \emph{(\bibinfo{series}{Proceedings of
  Machine Learning Research})}, Vol.~\bibinfo{volume}{162}.
  \bibinfo{publisher}{PMLR}, \bibinfo{pages}{11229--11254}.
\newblock


\bibitem[\protect\citeauthoryear{Krishnan, Franklin, Goldberg, and Wu}{Krishnan
  et~al\mbox{.}}{2017}]%
        {krishnan2017boostclean}
\bibfield{author}{\bibinfo{person}{Sanjay Krishnan}, \bibinfo{person}{Michael~J
  Franklin}, \bibinfo{person}{Ken Goldberg}, {and} \bibinfo{person}{Eugene
  Wu}.} \bibinfo{year}{2017}\natexlab{}.
\newblock \showarticletitle{Boostclean: Automated error detection and repair
  for machine learning}.
\newblock \bibinfo{journal}{\emph{arXiv preprint arXiv:1711.01299}}
  (\bibinfo{year}{2017}).
\newblock


\bibitem[\protect\citeauthoryear{Lahoti, Gummadi, and Weikum}{Lahoti
  et~al\mbox{.}}{2019}]%
        {pfr2019}
\bibfield{author}{\bibinfo{person}{Preethi Lahoti}, \bibinfo{person}{Krishna~P.
  Gummadi}, {and} \bibinfo{person}{Gerhard Weikum}.}
  \bibinfo{year}{2019}\natexlab{}.
\newblock \showarticletitle{Operationalizing Individual Fairness with Pairwise
  Fair Representations}.
\newblock \bibinfo{journal}{\emph{Proc. VLDB Endow.}} \bibinfo{volume}{13},
  \bibinfo{number}{4} (\bibinfo{date}{Dec.} \bibinfo{year}{2019}),
  \bibinfo{pages}{506–518}.
\newblock


\bibitem[\protect\citeauthoryear{Lattimore and Szepesvári}{Lattimore and
  Szepesvári}{2020}]%
        {lattimore2020}
\bibfield{author}{\bibinfo{person}{Tor Lattimore} {and} \bibinfo{person}{Csaba
  Szepesvári}.} \bibinfo{year}{2020}\natexlab{}.
\newblock \bibinfo{booktitle}{\emph{Bandit Algorithms}}.
\newblock \bibinfo{publisher}{Cambridge University Press}.
\newblock


\bibitem[\protect\citeauthoryear{Lee}{Lee}{2013}]%
        {Lee2013PseudoLabelT}
\bibfield{author}{\bibinfo{person}{Dong-Hyun Lee}.}
  \bibinfo{year}{2013}\natexlab{}.
\newblock \showarticletitle{Pseudo-Label : The Simple and Efficient
  Semi-Supervised Learning Method for Deep Neural Networks}.
\newblock


\bibitem[\protect\citeauthoryear{Levine, Crammer, and Mannor}{Levine
  et~al\mbox{.}}{2017}]%
        {rottingbandit}
\bibfield{author}{\bibinfo{person}{Nir Levine}, \bibinfo{person}{Koby Crammer},
  {and} \bibinfo{person}{Shie Mannor}.} \bibinfo{year}{2017}\natexlab{}.
\newblock \showarticletitle{Rotting Bandits}. In
  \bibinfo{booktitle}{\emph{NeurIPS}}. \bibinfo{pages}{3074--3083}.
\newblock


\bibitem[\protect\citeauthoryear{Lewis and Gale}{Lewis and Gale}{1994}]%
        {Lewis:1994:SAT:188490.188495}
\bibfield{author}{\bibinfo{person}{David~D. Lewis} {and}
  \bibinfo{person}{William~A. Gale}.} \bibinfo{year}{1994}\natexlab{}.
\newblock \showarticletitle{A Sequential Algorithm for Training Text
  Classifiers}. In \bibinfo{booktitle}{\emph{SIGIR}}. \bibinfo{pages}{3--12}.
\newblock


\bibitem[\protect\citeauthoryear{Liu, Zhou, and Rekatsinas}{Liu
  et~al\mbox{.}}{2022}]%
        {DBLP:journals/vldb/LiuZR22}
\bibfield{author}{\bibinfo{person}{Zifan Liu}, \bibinfo{person}{Zhechun Zhou},
  {and} \bibinfo{person}{Theodoros Rekatsinas}.}
  \bibinfo{year}{2022}\natexlab{}.
\newblock \showarticletitle{Picket: guarding against corrupted data in tabular
  data during learning and inference}.
\newblock \bibinfo{journal}{\emph{{VLDB} J.}} \bibinfo{volume}{31},
  \bibinfo{number}{5} (\bibinfo{year}{2022}), \bibinfo{pages}{927--955}.
\newblock


\bibitem[\protect\citeauthoryear{Menon and Williamson}{Menon and
  Williamson}{2018}]%
        {DBLP:conf/fat/MenonW18}
\bibfield{author}{\bibinfo{person}{Aditya~Krishna Menon} {and}
  \bibinfo{person}{Robert~C. Williamson}.} \bibinfo{year}{2018}\natexlab{}.
\newblock \showarticletitle{The cost of fairness in binary classification}. In
  \bibinfo{booktitle}{\emph{FAT}}, Vol.~\bibinfo{volume}{81}.
  \bibinfo{pages}{107--118}.
\newblock


\bibitem[\protect\citeauthoryear{Nguyen and Smeulders}{Nguyen and
  Smeulders}{2004}]%
        {10.1145/1015330.1015349}
\bibfield{author}{\bibinfo{person}{Hieu~Tat Nguyen} {and}
  \bibinfo{person}{Arnold W.~M. Smeulders}.} \bibinfo{year}{2004}\natexlab{}.
\newblock \showarticletitle{Active learning using pre-clustering}. In
  \bibinfo{booktitle}{\emph{ICML}}.
\newblock


\bibitem[\protect\citeauthoryear{Pedregosa, Varoquaux, Gramfort, Michel,
  Thirion, Grisel, Blondel, Prettenhofer, Weiss, Dubourg, Vanderplas, Passos,
  Cournapeau, Brucher, Perrot, and Duchesnay}{Pedregosa et~al\mbox{.}}{2011}]%
        {scikit-learn}
\bibfield{author}{\bibinfo{person}{F. Pedregosa}, \bibinfo{person}{G.
  Varoquaux}, \bibinfo{person}{A. Gramfort}, \bibinfo{person}{V. Michel},
  \bibinfo{person}{B. Thirion}, \bibinfo{person}{O. Grisel},
  \bibinfo{person}{M. Blondel}, \bibinfo{person}{P. Prettenhofer},
  \bibinfo{person}{R. Weiss}, \bibinfo{person}{V. Dubourg}, \bibinfo{person}{J.
  Vanderplas}, \bibinfo{person}{A. Passos}, \bibinfo{person}{D. Cournapeau},
  \bibinfo{person}{M. Brucher}, \bibinfo{person}{M. Perrot}, {and}
  \bibinfo{person}{E. Duchesnay}.} \bibinfo{year}{2011}\natexlab{}.
\newblock \showarticletitle{Scikit-learn: Machine Learning in {P}ython}.
\newblock \bibinfo{journal}{\emph{JMLR}}  \bibinfo{volume}{12}
  (\bibinfo{year}{2011}), \bibinfo{pages}{2825--2830}.
\newblock


\bibitem[\protect\citeauthoryear{Ratner, Bach, Ehrenberg, Fries, Wu, and
  R{\'e}}{Ratner et~al\mbox{.}}{2017}]%
        {ratner2017snorkel}
\bibfield{author}{\bibinfo{person}{Alexander Ratner},
  \bibinfo{person}{Stephen~H Bach}, \bibinfo{person}{Henry Ehrenberg},
  \bibinfo{person}{Jason Fries}, \bibinfo{person}{Sen Wu}, {and}
  \bibinfo{person}{Christopher R{\'e}}.} \bibinfo{year}{2017}\natexlab{}.
\newblock \showarticletitle{Snorkel: Rapid training data creation with weak
  supervision}.
\newblock \bibinfo{journal}{\emph{Proc. VLDB Endow.}} \bibinfo{volume}{11},
  \bibinfo{number}{3}, \bibinfo{pages}{269}.
\newblock


\bibitem[\protect\citeauthoryear{Roh, Lee, Whang, and Suh}{Roh
  et~al\mbox{.}}{2021a}]%
        {DBLP:conf/nips/RohLWS21}
\bibfield{author}{\bibinfo{person}{Yuji Roh}, \bibinfo{person}{Kangwook Lee},
  \bibinfo{person}{Steven Whang}, {and} \bibinfo{person}{Changho Suh}.}
  \bibinfo{year}{2021}\natexlab{a}.
\newblock \showarticletitle{Sample Selection for Fair and Robust Training}. In
  \bibinfo{booktitle}{\emph{NeurIPS}}. \bibinfo{pages}{815--827}.
\newblock


\bibitem[\protect\citeauthoryear{Roh, Lee, Whang, and Suh}{Roh
  et~al\mbox{.}}{2021b}]%
        {DBLP:conf/iclr/Roh0WS21}
\bibfield{author}{\bibinfo{person}{Yuji Roh}, \bibinfo{person}{Kangwook Lee},
  \bibinfo{person}{Steven~Euijong Whang}, {and} \bibinfo{person}{Changho Suh}.}
  \bibinfo{year}{2021}\natexlab{b}.
\newblock \showarticletitle{FairBatch: Batch Selection for Model Fairness}. In
  \bibinfo{booktitle}{\emph{ICLR}}.
\newblock


\bibitem[\protect\citeauthoryear{Roth and Small}{Roth and Small}{2006}]%
        {margin2006roth}
\bibfield{author}{\bibinfo{person}{Dan Roth} {and} \bibinfo{person}{Kevin
  Small}.} \bibinfo{year}{2006}\natexlab{}.
\newblock \showarticletitle{Margin-Based Active Learning for Structured Output
  Spaces}. In \bibinfo{booktitle}{\emph{ECML}}, Vol.~\bibinfo{volume}{4212}.
  \bibinfo{publisher}{Springer}, \bibinfo{pages}{413--424}.
\newblock


\bibitem[\protect\citeauthoryear{Roy and McCallum}{Roy and McCallum}{2001}]%
        {Roy:2001:TOA:645530.655646}
\bibfield{author}{\bibinfo{person}{Nicholas Roy} {and} \bibinfo{person}{Andrew
  McCallum}.} \bibinfo{year}{2001}\natexlab{}.
\newblock \showarticletitle{Toward Optimal Active Learning Through Sampling
  Estimation of Error Reduction}. In \bibinfo{booktitle}{\emph{ICML}}.
  \bibinfo{publisher}{Morgan Kaufmann Publishers Inc.}, \bibinfo{address}{San
  Francisco, CA, USA}, \bibinfo{pages}{441--448}.
\newblock
\showISBNx{1-55860-778-1}


\bibitem[\protect\citeauthoryear{Salimi, Rodriguez, Howe, and Suciu}{Salimi
  et~al\mbox{.}}{2019}]%
        {salimi2019interventional}
\bibfield{author}{\bibinfo{person}{Babak Salimi}, \bibinfo{person}{Luke
  Rodriguez}, \bibinfo{person}{Bill Howe}, {and} \bibinfo{person}{Dan Suciu}.}
  \bibinfo{year}{2019}\natexlab{}.
\newblock \showarticletitle{Interventional fairness: Causal database repair for
  algorithmic fairness}. In \bibinfo{booktitle}{\emph{SIGMOD}}.
  \bibinfo{pages}{793--810}.
\newblock


\bibitem[\protect\citeauthoryear{Sener and Savarese}{Sener and
  Savarese}{2018}]%
        {sener2018active}
\bibfield{author}{\bibinfo{person}{Ozan Sener} {and} \bibinfo{person}{Silvio
  Savarese}.} \bibinfo{year}{2018}\natexlab{}.
\newblock \showarticletitle{Active Learning for Convolutional Neural Networks:
  {A} Core-Set Approach}. In \bibinfo{booktitle}{\emph{ICLR}}.
\newblock


\bibitem[\protect\citeauthoryear{Settles}{Settles}{2012}]%
        {DBLP:series/synthesis/2012Settles}
\bibfield{author}{\bibinfo{person}{Burr Settles}.}
  \bibinfo{year}{2012}\natexlab{}.
\newblock \bibinfo{booktitle}{\emph{Active Learning}}.
\newblock \bibinfo{publisher}{Morgan {\&} Claypool Publishers}.
\newblock


\bibitem[\protect\citeauthoryear{Settles and Craven}{Settles and
  Craven}{2008}]%
        {Settles:2008:AAL:1613715.1613855}
\bibfield{author}{\bibinfo{person}{Burr Settles} {and} \bibinfo{person}{Mark
  Craven}.} \bibinfo{year}{2008}\natexlab{}.
\newblock \showarticletitle{An Analysis of Active Learning Strategies for
  Sequence Labeling Tasks}. In \bibinfo{booktitle}{\emph{EMNLP}}.
  \bibinfo{pages}{1070--1079}.
\newblock


\bibitem[\protect\citeauthoryear{Settles, Craven, and Ray}{Settles
  et~al\mbox{.}}{2007}]%
        {Settles:2007:MAL:2981562.2981724}
\bibfield{author}{\bibinfo{person}{Burr Settles}, \bibinfo{person}{Mark
  Craven}, {and} \bibinfo{person}{Soumya Ray}.}
  \bibinfo{year}{2007}\natexlab{}.
\newblock \showarticletitle{Multiple-instance Active Learning}. In
  \bibinfo{booktitle}{\emph{NIPS}} (Vancouver, British Columbia, Canada).
  \bibinfo{publisher}{Curran Associates Inc.}, \bibinfo{address}{USA},
  \bibinfo{pages}{1289--1296}.
\newblock
\showISBNx{978-1-60560-352-0}


\bibitem[\protect\citeauthoryear{Seung, Opper, and Sompolinsky}{Seung
  et~al\mbox{.}}{1992}]%
        {Seung:1992:QC:130385.130417}
\bibfield{author}{\bibinfo{person}{H.~S. Seung}, \bibinfo{person}{M. Opper},
  {and} \bibinfo{person}{H. Sompolinsky}.} \bibinfo{year}{1992}\natexlab{}.
\newblock \showarticletitle{Query by Committee}. In
  \bibinfo{booktitle}{\emph{COLT}} (Pittsburgh, Pennsylvania, USA).
  \bibinfo{publisher}{ACM}, \bibinfo{address}{New York, NY, USA},
  \bibinfo{pages}{287--294}.
\newblock
\showISBNx{0-89791-497-X}


\bibitem[\protect\citeauthoryear{Shannon}{Shannon}{2001}]%
        {entropy2001shannon}
\bibfield{author}{\bibinfo{person}{C.~E. Shannon}.}
  \bibinfo{year}{2001}\natexlab{}.
\newblock \showarticletitle{A Mathematical Theory of Communication}.
\newblock \bibinfo{journal}{\emph{SIGMOBILE Mob. Comput. Commun. Rev.}}
  \bibinfo{volume}{5}, \bibinfo{number}{1} (\bibinfo{date}{jan}
  \bibinfo{year}{2001}), \bibinfo{pages}{3–55}.
\newblock
\showISSN{1559-1662}


\bibitem[\protect\citeauthoryear{Sharaf, III, and Ni}{Sharaf
  et~al\mbox{.}}{2022}]%
        {DBLP:conf/fat/SharafDN22}
\bibfield{author}{\bibinfo{person}{Amr Sharaf},
  \bibinfo{person}{Hal~Daum{\'{e}} III}, {and} \bibinfo{person}{Renkun Ni}.}
  \bibinfo{year}{2022}\natexlab{}.
\newblock \showarticletitle{Promoting Fairness in Learned Models by Learning to
  Active Learn under Parity Constraints}. In \bibinfo{booktitle}{\emph{FAccT}}.
  \bibinfo{publisher}{{ACM}}, \bibinfo{pages}{2149--2156}.
\newblock


\bibitem[\protect\citeauthoryear{Shekhar, Fields, Ghavamzadeh, and
  Javidi}{Shekhar et~al\mbox{.}}{2021}]%
        {DBLP:conf/nips/ShekharFGJ21}
\bibfield{author}{\bibinfo{person}{Shubhanshu Shekhar}, \bibinfo{person}{Greg
  Fields}, \bibinfo{person}{Mohammad Ghavamzadeh}, {and} \bibinfo{person}{Tara
  Javidi}.} \bibinfo{year}{2021}\natexlab{}.
\newblock \showarticletitle{Adaptive Sampling for Minimax Fair Classification}.
  In \bibinfo{booktitle}{\emph{NeurIPS}}. \bibinfo{pages}{24535--24544}.
\newblock


\bibitem[\protect\citeauthoryear{Snoek, Larochelle, and Adams}{Snoek
  et~al\mbox{.}}{2012}]%
        {10.5555/2999325.2999464}
\bibfield{author}{\bibinfo{person}{Jasper Snoek}, \bibinfo{person}{Hugo
  Larochelle}, {and} \bibinfo{person}{Ryan~P. Adams}.}
  \bibinfo{year}{2012}\natexlab{}.
\newblock \showarticletitle{Practical Bayesian Optimization of Machine Learning
  Algorithms}. In \bibinfo{booktitle}{\emph{NeurIPS}}
  \emph{(\bibinfo{series}{NIPS'12})}. \bibinfo{pages}{2951–2959}.
\newblock


\bibitem[\protect\citeauthoryear{Tae and Whang}{Tae and Whang}{2021}]%
        {DBLP:conf/sigmod/TaeW21}
\bibfield{author}{\bibinfo{person}{Ki~Hyun Tae} {and}
  \bibinfo{person}{Steven~Euijong Whang}.} \bibinfo{year}{2021}\natexlab{}.
\newblock \showarticletitle{Slice Tuner: {A} Selective Data Acquisition
  Framework for Accurate and Fair Machine Learning Models}. In
  \bibinfo{booktitle}{\emph{SIGMOD}}. \bibinfo{publisher}{{ACM}},
  \bibinfo{pages}{1771--1783}.
\newblock


\bibitem[\protect\citeauthoryear{Tae, Zhang, Park, Rong, and Whang}{Tae
  et~al\mbox{.}}{2023}]%
        {techreport}
\bibfield{author}{\bibinfo{person}{Ki~Hyun Tae}, \bibinfo{person}{Hantian
  Zhang}, \bibinfo{person}{Jaeyoung Park}, \bibinfo{person}{Kexin Rong}, {and}
  \bibinfo{person}{Steven~Euijong Whang}.} \bibinfo{year}{2023}\natexlab{}.
\newblock \bibinfo{title}{Falcon: Fair Active Learning using MABs}.
\newblock
  \bibinfo{howpublished}{\url{https://github.com/khtae8250/Falcon/blob/main/techreport.pdf}}.
\newblock


\bibitem[\protect\citeauthoryear{Vartak, Rahman, Madden, Parameswaran, and
  Polyzotis}{Vartak et~al\mbox{.}}{2015}]%
        {vartak2015seedb}
\bibfield{author}{\bibinfo{person}{Manasi Vartak}, \bibinfo{person}{Sajjadur
  Rahman}, \bibinfo{person}{Samuel Madden}, \bibinfo{person}{Aditya
  Parameswaran}, {and} \bibinfo{person}{Neoklis Polyzotis}.}
  \bibinfo{year}{2015}\natexlab{}.
\newblock \showarticletitle{Seedb: Efficient data-driven visualization
  recommendations to support visual analytics}.
\newblock \bibinfo{journal}{\emph{Proc. VLDB Endow.}} \bibinfo{volume}{8},
  \bibinfo{number}{13}, \bibinfo{pages}{2182}.
\newblock


\bibitem[\protect\citeauthoryear{Verma and Rubin}{Verma and Rubin}{2018}]%
        {10.1145/3194770.3194776}
\bibfield{author}{\bibinfo{person}{Sahil Verma} {and} \bibinfo{person}{Julia
  Rubin}.} \bibinfo{year}{2018}\natexlab{}.
\newblock \showarticletitle{Fairness definitions explained}. In
  \bibinfo{booktitle}{\emph{FairWare@ICSE}}. \bibinfo{pages}{1--7}.
\newblock


\bibitem[\protect\citeauthoryear{Vermorel and Mohri}{Vermorel and
  Mohri}{2005}]%
        {10.1007/11564096_42}
\bibfield{author}{\bibinfo{person}{Joann{\`e}s Vermorel} {and}
  \bibinfo{person}{Mehryar Mohri}.} \bibinfo{year}{2005}\natexlab{}.
\newblock \showarticletitle{Multi-armed Bandit Algorithms and Empirical
  Evaluation}. In \bibinfo{booktitle}{\emph{ECML}}.
  \bibinfo{publisher}{Springer Berlin Heidelberg}, \bibinfo{pages}{437--448}.
\newblock


\bibitem[\protect\citeauthoryear{Wang and Shang}{Wang and Shang}{2014}]%
        {confidence2014wang}
\bibfield{author}{\bibinfo{person}{Dan Wang} {and} \bibinfo{person}{Yi Shang}.}
  \bibinfo{year}{2014}\natexlab{}.
\newblock \showarticletitle{A new active labeling method for deep learning}. In
  \bibinfo{booktitle}{\emph{IJCNN}}. \bibinfo{pages}{112--119}.
\newblock


\bibitem[\protect\citeauthoryear{Wang, Audibert, and Munos}{Wang
  et~al\mbox{.}}{2008}]%
        {NIPS2008_49ae49a2}
\bibfield{author}{\bibinfo{person}{Yizao Wang}, \bibinfo{person}{Jean-yves
  Audibert}, {and} \bibinfo{person}{R\'{e}mi Munos}.}
  \bibinfo{year}{2008}\natexlab{}.
\newblock \showarticletitle{Algorithms for Infinitely Many-Armed Bandits}. In
  \bibinfo{booktitle}{\emph{NeurIPS}}, Vol.~\bibinfo{volume}{21}.
  \bibinfo{publisher}{Curran Associates, Inc.}
\newblock


\bibitem[\protect\citeauthoryear{Whang, Roh, Song, and Lee}{Whang
  et~al\mbox{.}}{2023}]%
        {whangdata2023}
\bibfield{author}{\bibinfo{person}{Steven~Euijong Whang}, \bibinfo{person}{Yuji
  Roh}, \bibinfo{person}{Hwanjun Song}, {and} \bibinfo{person}{Jae{-}Gil Lee}.}
  \bibinfo{year}{2023}\natexlab{}.
\newblock \showarticletitle{Data Collection and Quality Challenges in Deep
  Learning: {A} Data-Centric {AI} Perspective}.
\newblock \bibinfo{journal}{\emph{{VLDB} J.}} (\bibinfo{year}{2023}).
\newblock


\bibitem[\protect\citeauthoryear{Wu, Su, Chen, and Hsu}{Wu
  et~al\mbox{.}}{2022}]%
        {wu2022fair}
\bibfield{author}{\bibinfo{person}{Tsung-Han Wu}, \bibinfo{person}{Hung-Ting
  Su}, \bibinfo{person}{Shang-Tse Chen}, {and} \bibinfo{person}{Winston~H.
  Hsu}.} \bibinfo{year}{2022}\natexlab{}.
\newblock \bibinfo{title}{Fair Robust Active Learning by Joint Inconsistency}.
\newblock
\newblock
\showeprint[arxiv]{2209.10729}


\bibitem[\protect\citeauthoryear{Zhang, Chu, Asudeh, and Navathe}{Zhang
  et~al\mbox{.}}{2021}]%
        {DBLP:conf/sigmod/ZhangCAN21}
\bibfield{author}{\bibinfo{person}{Hantian Zhang}, \bibinfo{person}{Xu Chu},
  \bibinfo{person}{Abolfazl Asudeh}, {and} \bibinfo{person}{Shamkant~B.
  Navathe}.} \bibinfo{year}{2021}\natexlab{}.
\newblock \showarticletitle{OmniFair: {A} Declarative System for Model-Agnostic
  Group Fairness in Machine Learning}. In \bibinfo{booktitle}{\emph{SIGMOD}}.
  \bibinfo{publisher}{{ACM}}, \bibinfo{pages}{2076--2088}.
\newblock


\bibitem[\protect\citeauthoryear{Zhang, Tae, Park, Chu, and Whang}{Zhang
  et~al\mbox{.}}{2023}]%
        {zhangiflipper2023}
\bibfield{author}{\bibinfo{person}{Hantian Zhang}, \bibinfo{person}{Ki~Hyun
  Tae}, \bibinfo{person}{Jaeyoung Park}, \bibinfo{person}{Xu Chu}, {and}
  \bibinfo{person}{Steven~Euijong Whang}.} \bibinfo{year}{2023}\natexlab{}.
\newblock \showarticletitle{iFlipper: Label Flipping for Individual Fairness}.
\newblock \bibinfo{journal}{\emph{Proc. {ACM} Manag. Data}}
  \bibinfo{volume}{1}, \bibinfo{number}{1} (\bibinfo{year}{2023}),
  \bibinfo{pages}{8:1--8:26}.
\newblock


\bibitem[\protect\citeauthoryear{Zhao and Gordon}{Zhao and Gordon}{2019}]%
        {DBLP:conf/nips/ZhaoG19}
\bibfield{author}{\bibinfo{person}{Han Zhao} {and} \bibinfo{person}{Geoffrey~J.
  Gordon}.} \bibinfo{year}{2019}\natexlab{}.
\newblock \showarticletitle{Inherent Tradeoffs in Learning Fair
  Representations}. In \bibinfo{booktitle}{\emph{NeurIPS}}.
  \bibinfo{pages}{15649--15659}.
\newblock


\end{thebibliography}

\ifthenelse{\boolean{techreport}}{\clearpage 
\newpage
\appendix

\section{Target Groups for Other Metrics.}
We continue from Section~\ref{sec:subgrouplabeling} and explain the target subgroups for Equalized Odds (ED), Predictive Parity (PP) and Equalized Error Rate (EER). For ease of reference, we present the complete results for each fairness metric in Table~\ref{tbl:targetgroups2} again.

\stitle{Target Subgroups for ED.} In addition to the EO case in Example~2, for ED (Equation~\ref{eq:ed}), the goal is to also close the gap between $p(\hat{y}=1|y=0,z=0)$ and $p(\hat{y}=1|y=0,z=1)$. Then, improving $p(\hat{y}=0|y=0,z=1)$ directly decreases $p(\hat{y}=1|y=0,z=1)$ and reduces the disparity. Hence, improving the accuracy on $(y=1, z=0)$ or $(y=0, z=1)$, whichever minimizes the larger gap, will result in fairness improvement.

\stitle{Target Subgroups for PP.} For PP (Equation~\ref{eq:pp}), let us assume that the FDR disparity is dominant, and the goal is to close the gap between $p(y=1|\hat{y}=1, z=0)$ and $p(y=1|\hat{y}=1, z=1)$ where the first term is smaller. We know that 
\begin{align*}
p(y=1 | \hat{y} = 1) &= \frac{p(y=1, \hat{y}=1)}{p(\hat{y}=1)} = \frac{p(y=1)p(y=1|\hat{y}=1)}{p(\hat{y}=1)}
\end{align*}

Then $p(y=1|\hat{y} = 1, z=0) < p(y=1|\hat{y}=1. z=1)$ can be rewritten as $\frac{p(y=1|z=0)p(y=1|\hat{y}=1, z=0)}{p(\hat{y}=1|z=0)} < \frac{p(y=1|z=1)p(y=1|\hat{y}=1, z=1)}{p(\hat{y}=1|z=1)}$. We also know that the denominator term $p(\hat{y}=1)$ can be expressed as $p(y=1)p(\hat{y}=1|y=1) + p(y=0)(1-p(\hat{y}=0|y=0))$. By arranging the terms, we get the following inequality: 
\begin{align*}
\frac{p(y=0|z=1)(1-p(\hat{y}=0|y=0, z=1))}{p(y=1|z=1)p(\hat{y}=1|y=1,z=1)} < \\
\frac{p(y=0|z=0)(1-\underline{p(\hat{y}=0|y=0, z=0))}}{p(y=1|z=0)\underline{p(\hat{y}=1|y=1,z=0)}}
\end{align*}

In this case, we can see that improving $p(\hat{y}=0|y=0, z=0))$ or $p(\hat{y}=1|y=1,z=0)$ results in deceasing the right term. Therefore, both $(y=0, z=0)$ and $(y=1, z=0)$ can be the target subgroups to improve PP. 

Consider the other case where the FOR disparity is larger, and $p(y=1|\hat{y}=0, z=1) < p(y=1|\hat{y}=0, z=1)$. Using a similar approach, we can derive that the target groups are $(y=0, z=1)$ and $(y=1, z=1)$ to close the FOR gap. We summarize the results for all possible scenarios in Table~\ref{tbl:targetgroups2}.


\stitle{Target Subgroups for EER.}
For EER (Equation~\ref{eq:eer}), the goal is to ensure a similar classification error rate across different sensitive groups. Similar to the previous examples, we assume that $p(\hat{y} \ne y|z=0) < p(\hat{y} \ne y|z=1)$, which can also be expressed as $p(\hat{y} = y|z=1) < p(\hat{y} = y|z=0)$. In order to improve $p(\hat{y} = y|z=1)$ in the left term, we need to label data from $(y=1,z=1)$ or $(y=0,z=1)$, as $p(\hat{y} = y|z=1)$ can be written as $p(y=1, z=1)\underline{p(\hat{y} = 1|y=1, z=1)}+p(y=0, z=1)\underline{p(\hat{y}=0|y=0, z=1)}$. On the other hand, if $p(\hat{y} \ne y|z=1)$ is smaller, both $(y=1,z=1)$ and $(y=0,z=1)$ subgroups should be targeted to improve EER.

\paragraph{Trial-and-error for PP and EER} For the target groups for PP and EER in Table~\ref{tbl:targetgroups2}, we observe that an undesirable label with one of the target groups corresponds to another target subgroup. Hence, trial-and-error does not postpone any samples in these cases.

\begin{table}[t]
  \centering
  \begin{tabular}{ccc}
  \toprule
  {\bf Metric} & {\bf Target Subgroups} $\mathbf{(y, z)}$ & \\
  \midrule
  DP & $(1, z^*)$ or $(0, 1-z^*)$ \\
  \midrule
  EO & $(1, z^*)$ \\
  \midrule
  \multirow{2}{*}{ED} & $(0, 1-z^*)$, if FPR gap $\geq$ FNR gap\\
  & $(1, z^*)$, otherwise \\
  \midrule
  \multirow{2}{*}{PP} & $(0, 1-z^*)$ or $(1, 1-z^*)$, if FOR gap $\geq$ FDR gap \\
  & $(0, z^*)$ or $(1, z^*)$, otherwise \\
  \midrule
  EER & $(0, 1-z^*)$ or $(1, 1-z^*)$ \\
  \toprule
  \end{tabular}
  \caption{Target subgroups for each group fairness measure when a sensitive group $z^* \in \{0, 1\}$ has a lower fairness value.
  }
  \label{tbl:targetgroups2}
\end{table}

\section{Implementation Details}

\subsection{Dataset Configurations}
We continue from Section~\ref{sec:setting} and provide more details on the data configurations. Table~\ref{tbl:datasets} shows how we construct the data distributions for the four datasets. We observe that the original distributions of the datasets are not heavily biased. That is, we can achieve a nearly perfect fair classifier using a relatively small labeling budget. Although \method{} also performs well in this scenario, it becomes difficult to compare the performance of different methods because labeling a few samples is sufficient to improve fairness. Hence, we increase the bias by taking a smaller subset of the minority group or a larger subset of the majority group, and then conduct experiments using a larger labeling budget. 

The sensitive groups used for each dataset are as follows:
\begin{itemize}
\item TravelTime: Female and Male.
\item Employ: Disability and Able-bodied.
\item Income: White, Asian, and Others.
\item COMPAS: Female and Male.
\end{itemize}

\begin{table*}[t]
  \centering
  \begin{tabular}{cccccccc}
  \toprule
  {\bf Datasets} & {\bf Sizes} & \multicolumn{2}{c}{\bf Sen. Group 1} & \multicolumn{2}{c}{\bf Sen. Group 2} & \multicolumn{2}{c}{\bf Sen. Group 3} \\
  \midrule
  & & {\bf Label 0} & {\bf Label 1} & {\bf Label 0} & {\bf Label 0} & {\bf Label 0} & {\bf Label 0} \\
  \midrule
  & $|D_{train}|$ & 1,115 & 181 & 441 & 709 & -& -\\
  TravelTime & $|D_{un}|$ & 22,300 & 3,630 & 8,820 & 14,190 & - & -\\
  (gender) & $|D_{test}|$ & 11,150 & 1,815& 4,410 & 7,095 & - & -\\
  & $|D_{val}|$ & 1,115 & 181 & 441 & 709 & -& -\\
  \midrule
  & $|D_{train}|$ & 579 & 81 & 1,673 & 3,292 & - & -\\
  Employ & $|D_{un}|$ & 17,370 & 2,430 & 50,190 & 98,760 & - & -\\
  (disability) & $|D_{test}|$ & 8,685 & 1,215 & 25,095 & 49,380 & - & -\\
  & $|D_{val}|$ & 579 & 81 & 1,673 & 3,292 & - & -\\
  \midrule
  & $|D_{train}|$ & 2,019 & 268 & 169 & 314 & 309 & 109 \\
  Income & $|D_{un}|$ & 40,380 & 5,360 & 3,380 & 6,280 & 6,180 & 2,180 \\
  (race) & $|D_{test}|$ & 20,190 & 2,680 & 1,690 & 3,140 & 3,090 & 1,090 \\
  & $|D_{val}|$ & 2,019 & 268 & 169 & 314 & 309 & 109\\
  \midrule
  & $|D_{train}|$ & 86 & 158 & 37 & 13 & -& -\\
  COMPAS & $|D_{un}|$ & 688 & 1,264 & 300 & 104 & - & -\\
  (gender) & $|D_{test}|$ & 344 & 632 & 150 & 52 & - & -\\
  & $|D_{val}|$ & 86 & 158 & 37 & 13 & - & -\\
  \bottomrule
  \end{tabular}
  \caption{Detailed configurations for the four datasets, with the sensitive attribute in parentheses.
  }
  \label{tbl:datasets}
\end{table*}

\subsection{Baselines}
We continue from Section~\ref{sec:setting} and provide more details on the fair AL algorithms, \fal{}~\cite{DBLP:journals/eswa/AnahidehAT22} and \decouple{}~\cite{cao2022decouple}.

\begin{itemize}
\item \fal{}: The first fairness-aware AL algorithm that optimizes both group fairness and accuracy. \fal{} linearly combines entropy with the expected unfairness reduction, which estimates the expected fairness improvement for each unlabeled sample over all possible labels. This approach, however, requires retraining the model $2 \times |D_{un}|$ times per iteration. In order to improve efficiency, \fal{} computes the reduction in unfairness only for $m$ samples with the highest entropy value, and then chooses top $b$ samples from this subset. As a result, a higher $m$ favors better fairness, but requires more computation time. In our experiments, we set an upper bound for the $m$ value to 64, as it has been reported to provide the best performance\,\cite{cao2022decouple}.
 
\item \decouple{}: A disagreement-based fairness-aware AL algorithm. For a binary-valued sensitive attribute (i.e., $\mathbb{Z} = \{0, 1\}$), a decouple model is a pair of models $(h_0, h_1)$, where each model $h_i$ is trained on a specific sensitive group $z_i$. \decouple{} chooses a sample $x$ that receives significantly different predictions from the decoupled models $h_0$ and $h_1$,  i.e., $|p(h_{0}(x)=1|x) - p(h_{1}(x)=1|x)| > \alpha$ for a predefined hyperparameter $\alpha$. For handling multiple sensitive attributes, we extend \decouple{} to find a sample $x$ that receives conflicting predictions from any two models, i.e., $\text{max}_{z_{i},z_{j}\in \mathbb{Z}} |p(h_{i}(x)=1|x) - p(h_{j}(x)=1|x)| > \alpha$. In our experiments, we vary the $\alpha$ value from 0.1 to 0.9.
\end{itemize}

One technique we do not make a comparison is PANDA~\cite{DBLP:conf/fat/SharafDN22}, which is a meta-learning based algorithm that learns a selection policy that maximizes accuracy and fairness. We exclude this method due to the prohibitively high computational cost of meta-learning, as also noted in~\cite{wu2022fair}. 

\subsection{Fairness Evaluation}
We continue from Section~\ref{sec:setting} and provide more details on the fairness evaluation. We consider five group fairness measures including demographic parity (DP), equal opportunity (EO), equalized odds (ED), predictive parity (PP), and equalized error rate (EER). To quantify fairness, we define a fairness score as one minus the maximum fairness disparity~\cite{bird2020fairlearn} across any sensitive groups on the test set, as described in Section~\ref{sec:preliminaries}. A higher value is better.
\begin{itemize}
\item DP score: $1 - \text{max}_{z_{i},z_{j}\in \mathbb{Z}} |p(\hat{y}=1|z=z_{i}) - p(\hat{y}=1|z=z_{j})|$
\item EO score: $1 - \text{max}_{z_{i},z_{j}\in \mathbb{Z}} |p(\hat{y}=1|y=1, z=z_{i}) - p(\hat{y}=1|y=1, z=z_{j})|$
\item ED score: $1 - \text{max}_{z_{i},z_{j}\in \mathbb{Z}, \text{y}\in\text{\{0,1\}}}|p(\hat{y}=1|y=\text{y}, z=z_{i}) - p(\hat{y}=1|y=\text{y}, z=z_{j})|$
\item PP score: $1 - \text{max}_{z_{i},z_{j}\in \mathbb{Z}, \hat{\text{y}}\in\text{\{0,1\}}} |p(y=1|\hat{y}=\hat{\text{y}}, z=z_{i}) - p(y=1|\hat{y}=\hat{\text{y}}, z=z_{j})|$ 
\item EER score: $1 - \text{max}_{z_{i},z_{j}\in \mathbb{Z}} |p(\hat{y}\ne y|z=z_{i})-p(\hat{y}\ne y|z=z_{j})|$
\end{itemize}

\subsection{Other Experimental Settings}
We continue from Section~\ref{sec:setting} and provide more details on the experimental settings. We implement logistic regression (LR) and neural network (NN) models using Scikit-learn~\cite{scikit-learn} library. \rev{For LR, we set the regularization strength to 1.0. For NN, we use a multi-layer perceptron with one hidden layer consisting of 10 nodes and set the learning rate to 0.0001.} We evaluate all models on a separate test set and repeat the experiments with ten different random seeds. For the NN experiments, we use three random seeds due to the long comparison time, which exceeds 30 hours per seed. 

\begin{figure*}[t]
  \centering
  \begin{subfigure}{0.33\textwidth}
     \includegraphics[width=\columnwidth]{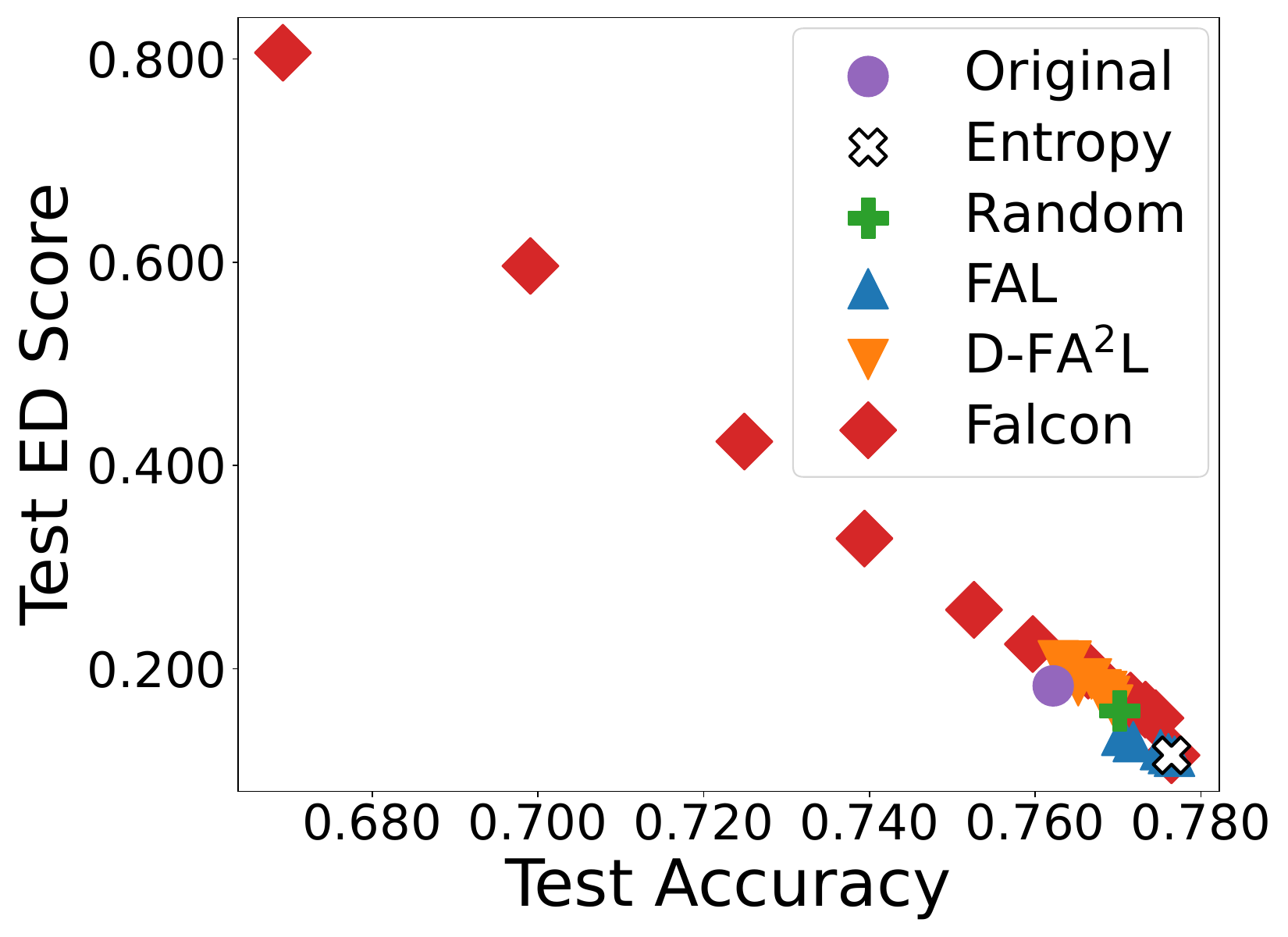}
     \caption{TravelTime (ED)}
     \label{fig:timeed}
  \end{subfigure}
  \begin{subfigure}{0.33\textwidth}
    \includegraphics[width=\columnwidth]{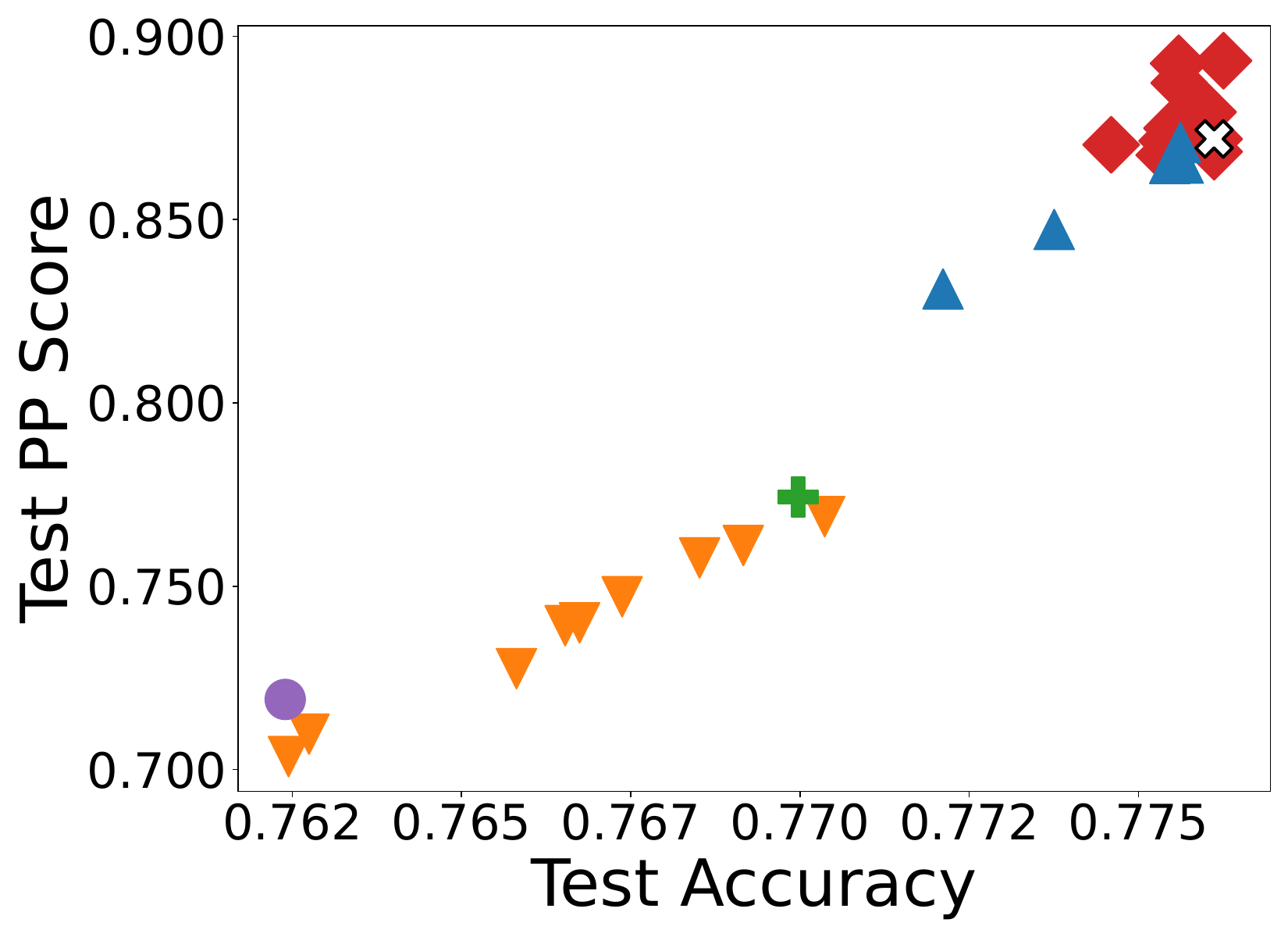}
     \caption{TravelTime (PP)}
     \label{fig:timepp}
  \end{subfigure} 
  \begin{subfigure}{0.33\textwidth}
    \includegraphics[width=\columnwidth]{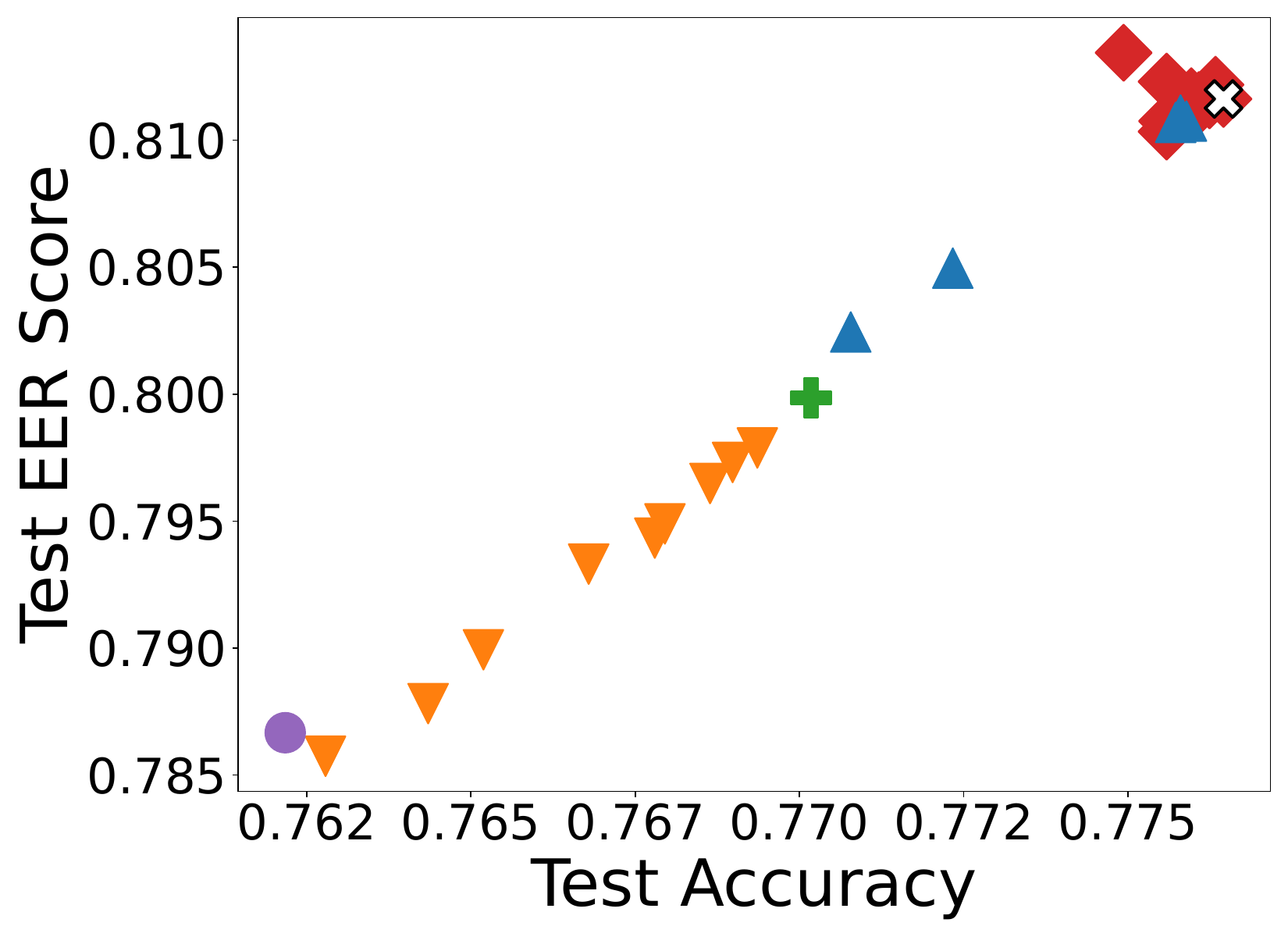}
     \caption{TravelTime (EER)}
     \label{fig:timeeer}
  \end{subfigure}
\begin{subfigure}{0.33\textwidth}
    \includegraphics[width=\columnwidth]{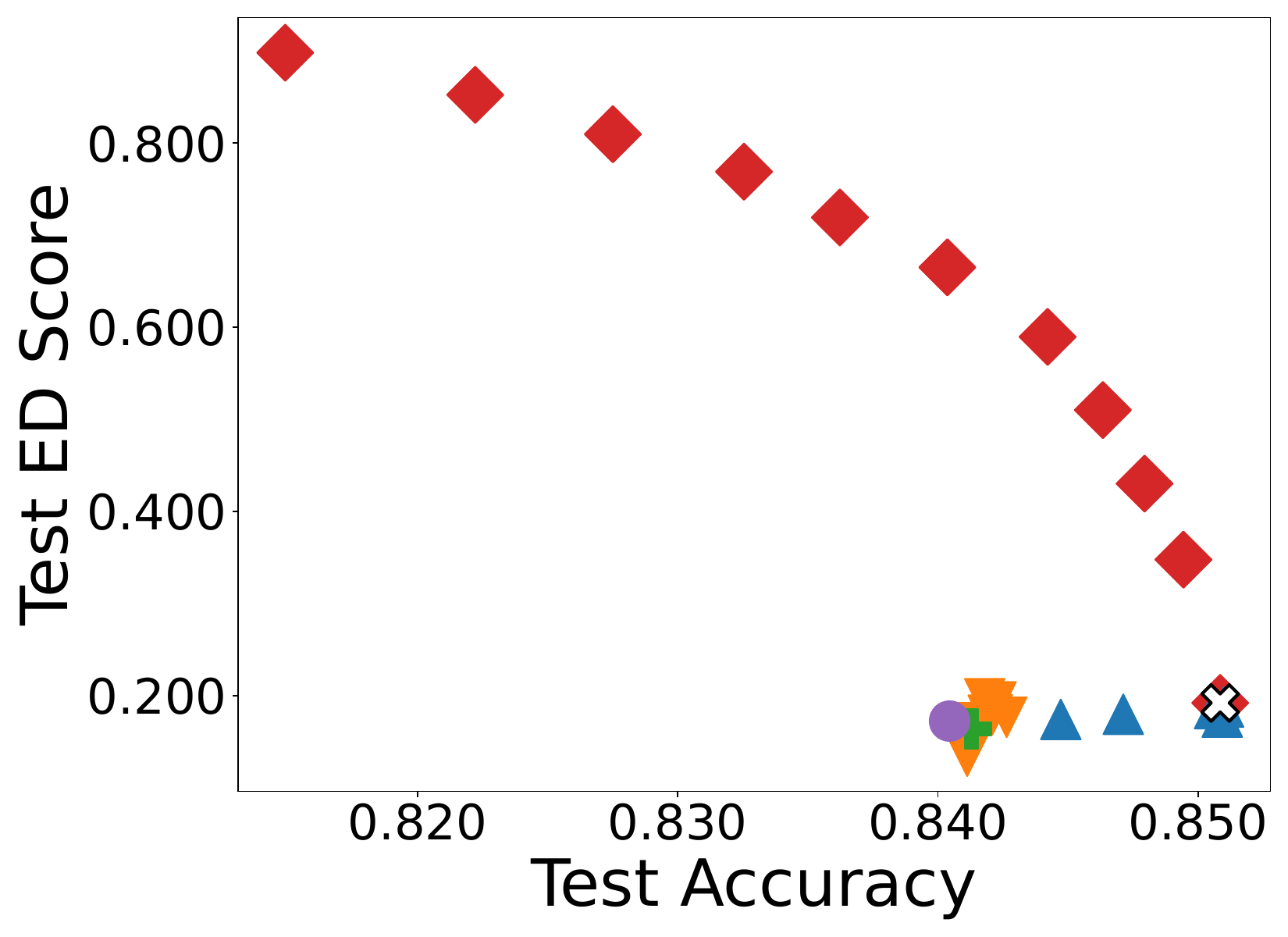}
     \caption{Employ (ED)}
     \label{fig:employed}
  \end{subfigure}
  \begin{subfigure}{0.33\textwidth}
     \includegraphics[width=\columnwidth]{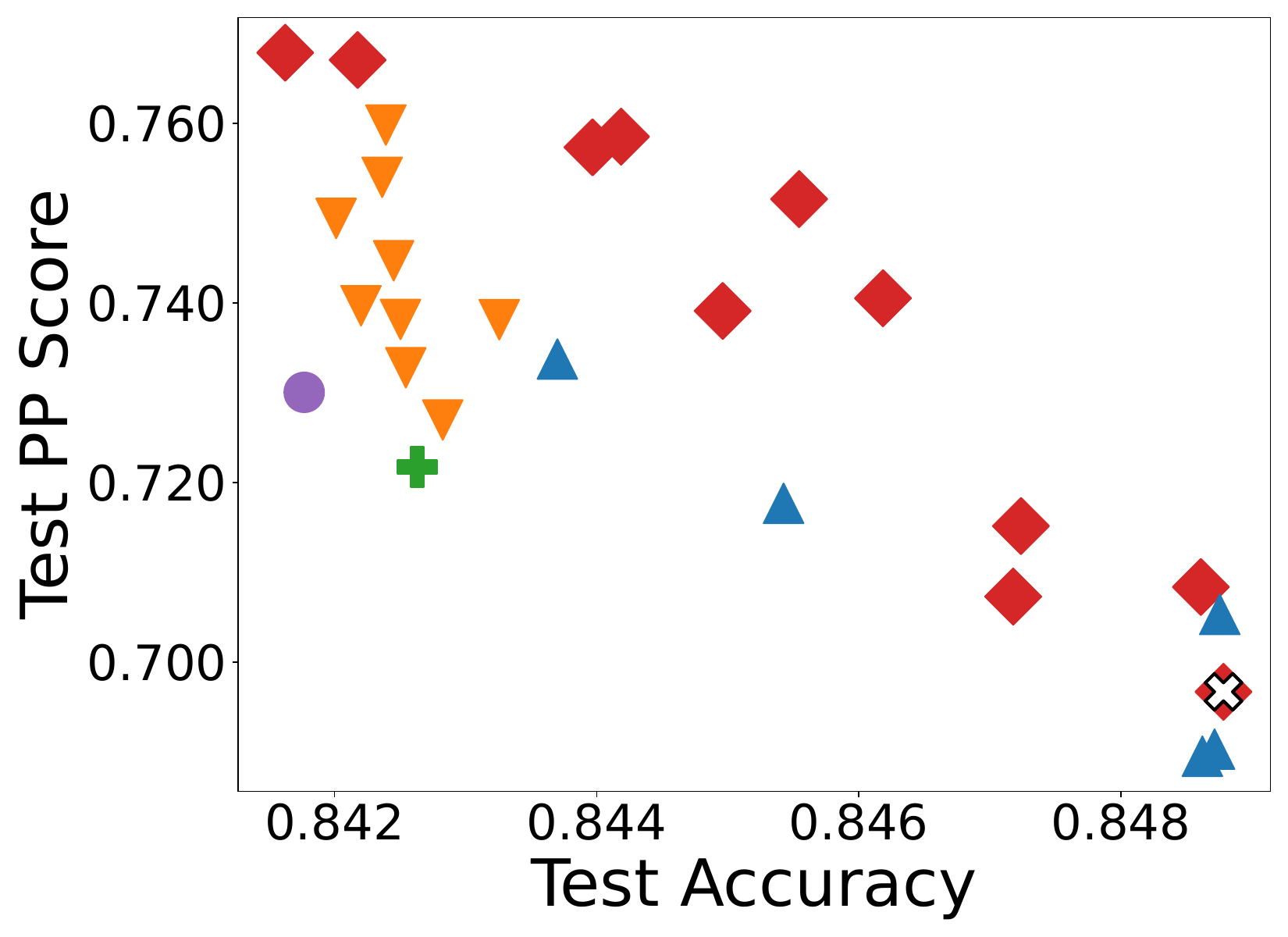}
     \caption{Employ (PP)}
     \label{fig:employpp}
  \end{subfigure}
  \begin{subfigure}{0.33\textwidth}
    \includegraphics[width=\columnwidth]{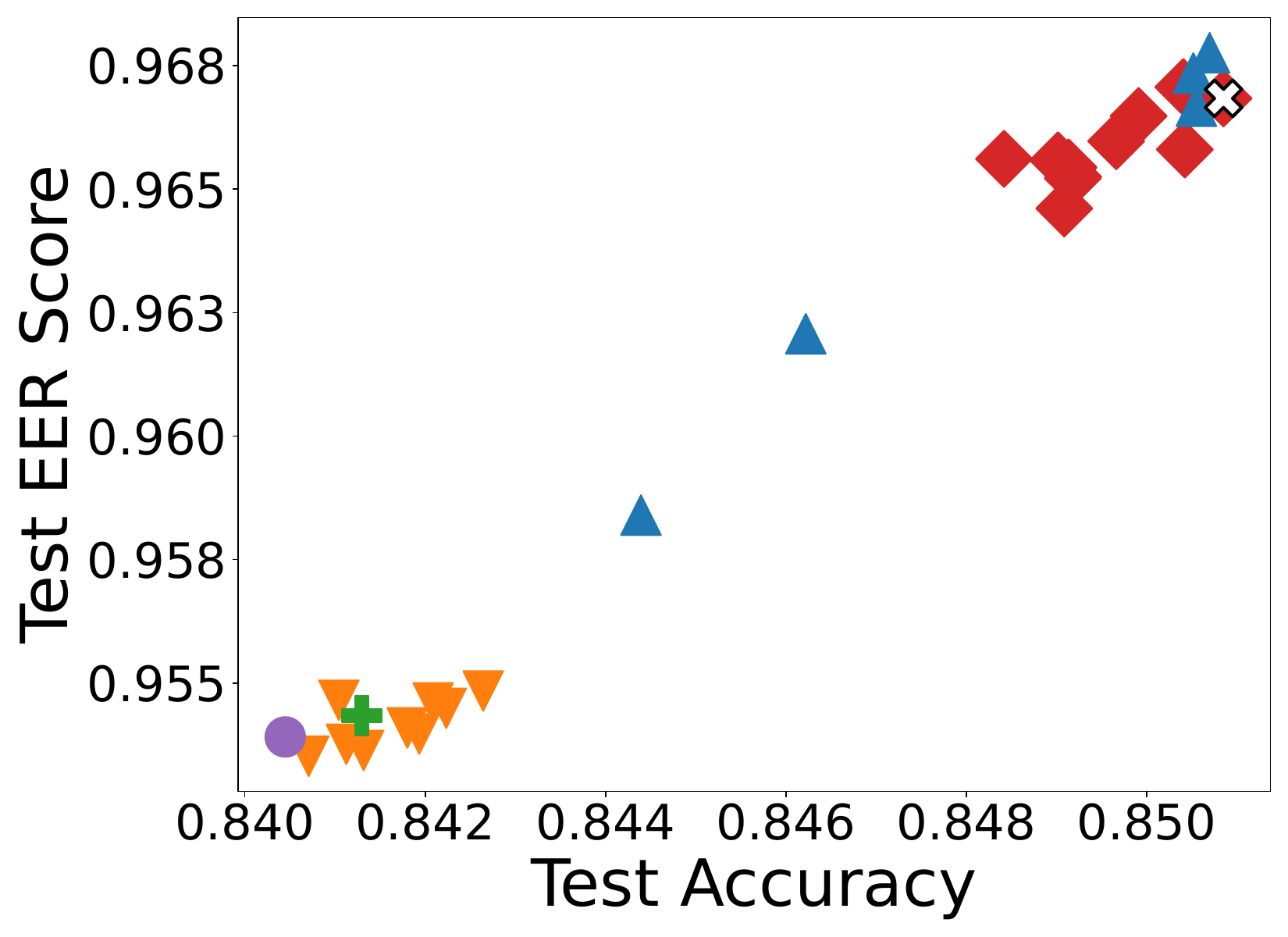}
     \caption{Employ (EER)}
     \label{fig:employeer}
  \end{subfigure} 
     \caption{Accuracy-fairness trade-offs of logistic regression on the two datasets (TravelTime and Employ) using the three fairness measures (ED, PP, and EER). In addition to the baselines, we add the result of model training without labeling any additional data and call it ``Original.'' As a result, \method{} significantly improves fairness compared to the fair AL baselines.  
     }
 \label{fig:tradeoffsothers}
\end{figure*}

\section{Trade-offs for Other Measures}
We continue from Section~\ref{exp:accuracyandfairness} and perform the same experiments using equalized odds (ED), predictive parity (PP), and equalized error rate (EER). Figure~\ref{fig:tradeoffsothers} shows the accuracy-fairness trade-off results on the TravelTime and Employ datasets. The key trends are still similar to Figure~\ref{fig:tradeoffs} where \method{} outperforms the other fair AL baselines in terms of accuracy and fairness. 

Another interesting observation is that we have different shapes of trade-offs for EER (Figure~\ref{fig:timeeer} and Figure~\ref{fig:employeer}) and PP (Figure~\ref{fig:timepp}), where fairness and accuracy can be improved at the same time. This is due to the fact that improving fairness sometimes aligns with improving overall accuracy. For EER, the goal is to equalize the accuracy between different sensitive groups. Hence, we need to label more samples from the group with lower accuracy, which leads to an overall accuracy improvement as well. In addition, the result for PP on the TravelTime dataset (Figure~\ref{fig:timepp}) exhibits a similar pattern because the target groups are sometimes the same as those for EER (detailed conditions are specified in Table~\ref{tbl:targetgroups2}). We also note that \method{} and \entropy{} produce comparable results in these cases, as their underlying objectives are similar, i.e., improving the accuracy of minority groups.

\section{Policy Search for Employ Dataset}
We continue from Section~\ref{exp:policysearch} and show experimental results on the Employ dataset where we use DP and EO as the target fairness measures. In Figure~\ref{fig:policycomparisonemploy}, the key trends are still similar to Figure~\ref{fig:policycomparison} where \method{} achieves the best or second-best fairness improvement among the single policy baselines. In addition, we provide a detailed analysis for Figure~\ref{fig:policycomparisonemploydp} in Figure~\ref{fig:policyfairnessemploy} where we show how \method{} updates the selection probabilities of each policy. Here, the sensitive attribute is \texttt{disability}, and we have two target groups, \texttt{(attribute=disability, label=positive)} and \texttt{(attribute=able-bodied, label=negative)}, denoted as $(D-1)$ and $(A-0)$, respectively. As a result, \method{} increases the selection probability for $r=0.4$ for $(D-1)$ the most. This finding is consistent with Figure~\ref{fig:eachpolicyfairnessemploy}, where it shows that $r=0.4$ for $(D-1)$ and $r=0.5$ for $(D-1)$ are the most effective policies. Hence, we conclude that \method{} correctly updates the MAB to identify the optimal policy among the candidate policy set.

\begin{figure}[t]
  \centering
  \begin{subfigure}{0.495\columnwidth}
     \includegraphics[width=\columnwidth]{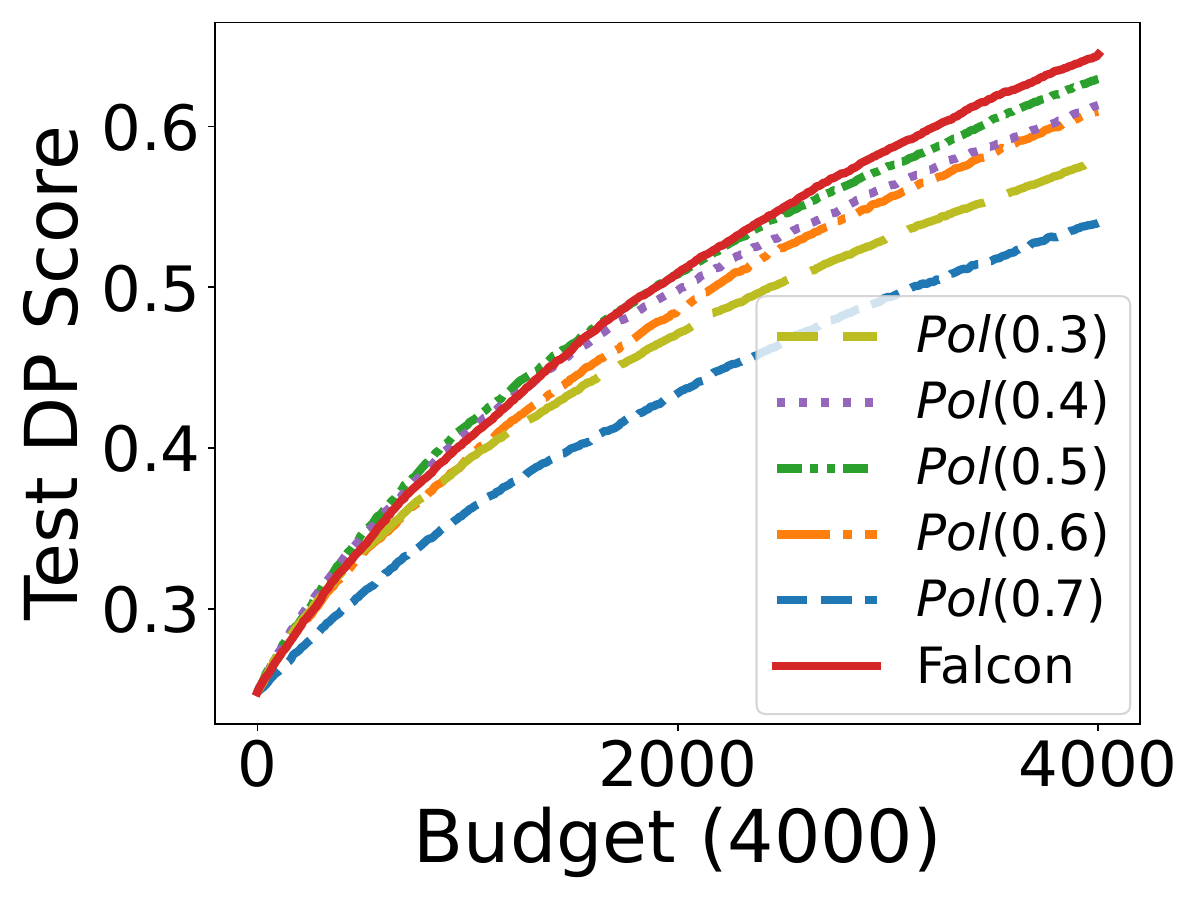}
     \caption{Test DP Score} \label{fig:policycomparisonemploydp}
  \end{subfigure}
  \begin{subfigure}{0.495\columnwidth}
    \includegraphics[width=\columnwidth]{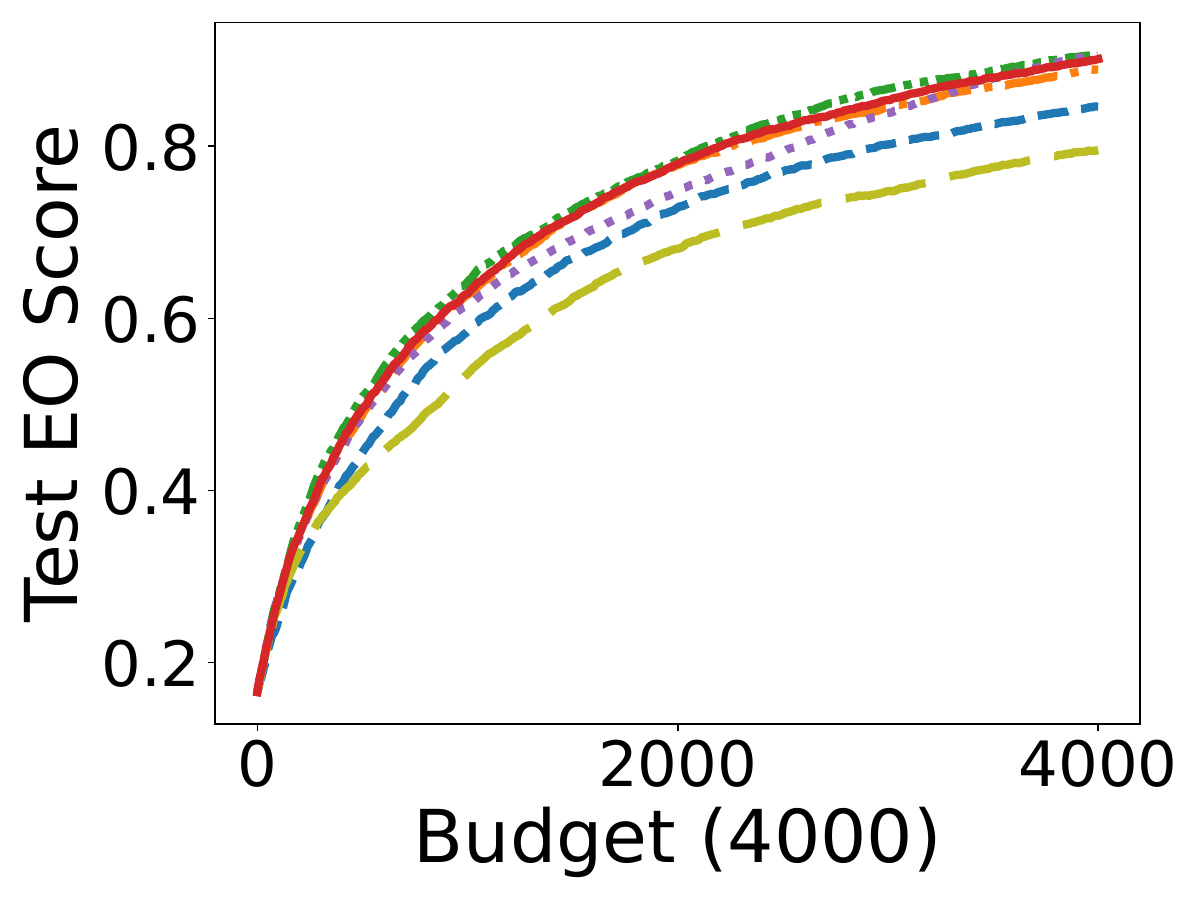}
     \caption{Test EO score} \label{fig:policycomparisonemployeo}
  \end{subfigure} 
     \caption{Fairness comparison of \method{} against a set of single policy baselines on the Employ dataset.}
     \label{fig:policycomparisonemploy}
\end{figure}

\begin{figure}[t]
  \centering
  \begin{subfigure}{0.495\columnwidth}
     \includegraphics[width=\columnwidth]{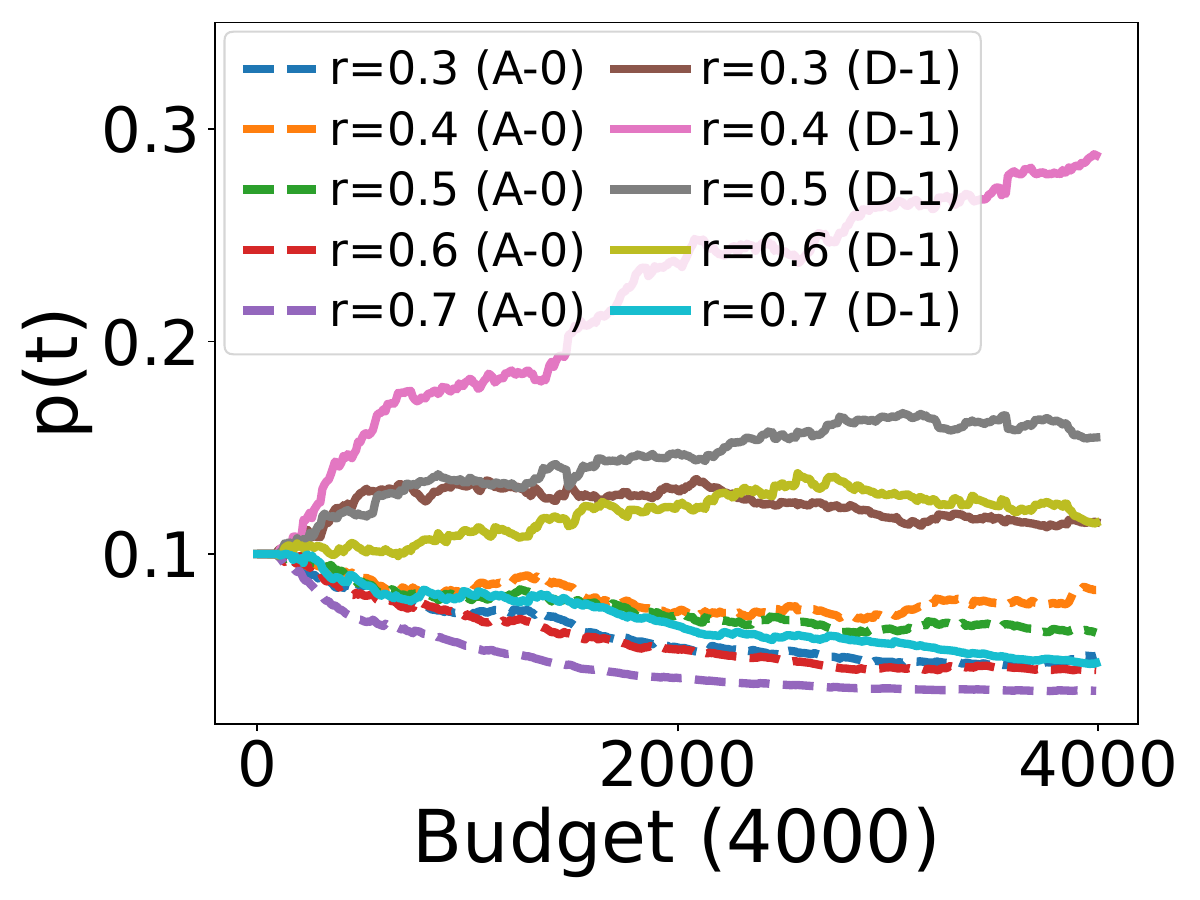}
     \caption{Policy Selection Probability} \label{fig:policyfairnessemploy}
  \end{subfigure}
    \begin{subfigure}{0.495\columnwidth}
    \includegraphics[width=\columnwidth]{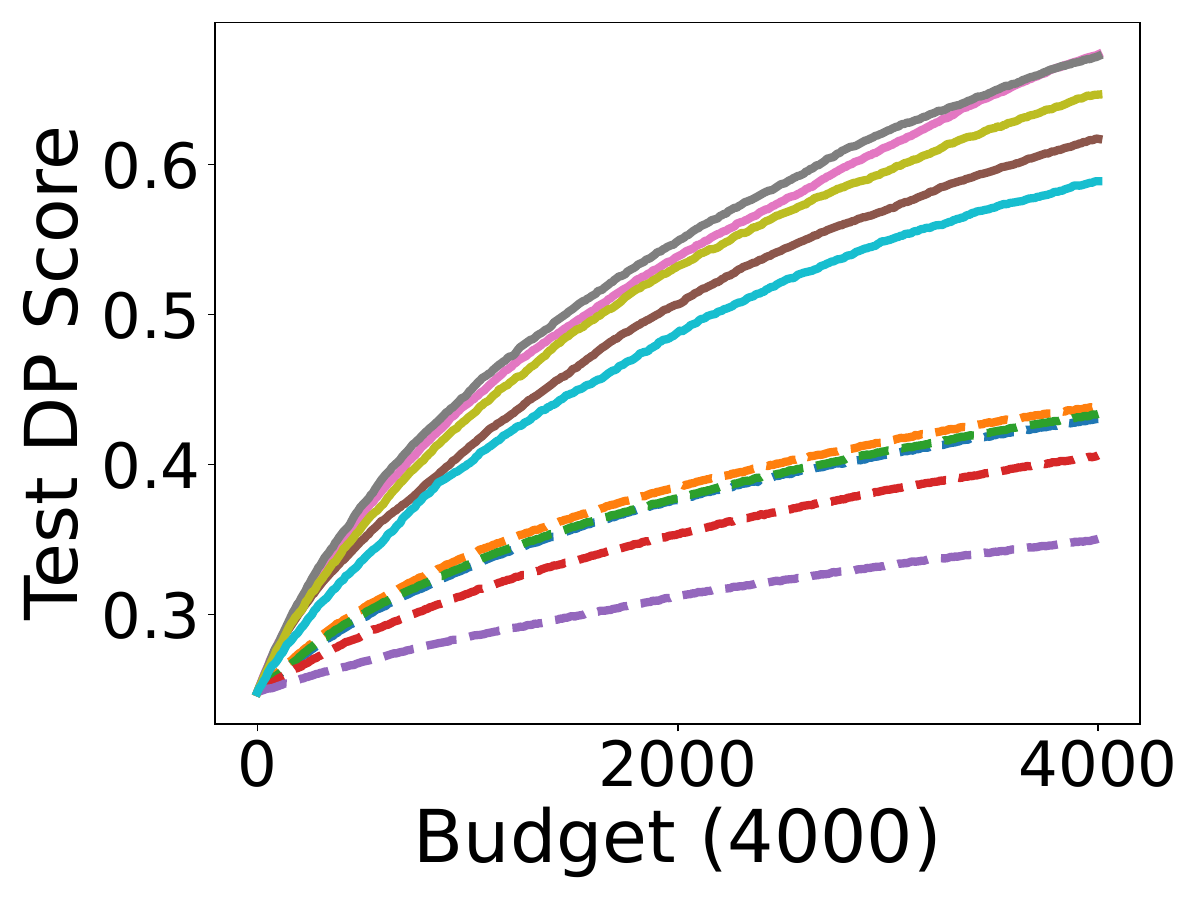}
     \caption{All Single Policies} \label{fig:eachpolicyfairnessemploy}
  \end{subfigure} 
     \caption{A detailed analysis for Figure~\ref{fig:policycomparisonemploydp}. (a) \method{} increases the selection probability of $r=0.4$ for the target group $(D-1)$, where we denote the sensitive attribute (Disability or Able-bodied) and label of the target subgroup in parentheses. (b) Fairness improvements for all single policies. The policies $r=0.4$ for $(D-1)$ and $r=0.5$ for $(D-1)$ are the most effective in improving the DP score.
     }
     \label{fig:policyselectionemploydetails}
\end{figure}

\section{Comparison with Neural Network} \label{exp:nnexperiments}
In Section~\ref{exp:accuracyandfairness}, we compared \method{} with the baselines using logistic regression models. In this section, we perform the same experiments using neural network models. Figure~\ref{fig:tradeoffsnn} is the trade-off results on the TravelTime and Employ datasets when using DP and EO. For \fal{}, we exclude the results for $m=32$ and $m=64$ because the overall running time takes more than 24 hours. The observations are similar to those of Figure~\ref{fig:tradeoffs} where \method{} consistently outperforms the other baselines in terms of fairness and accuracy and provides much cleaner trade-offs. The results clearly demonstrate how \method{} benefits other ML models.

\begin{figure}[t]
  \centering
  \begin{subfigure}{0.495\columnwidth}
     \includegraphics[width=\columnwidth]{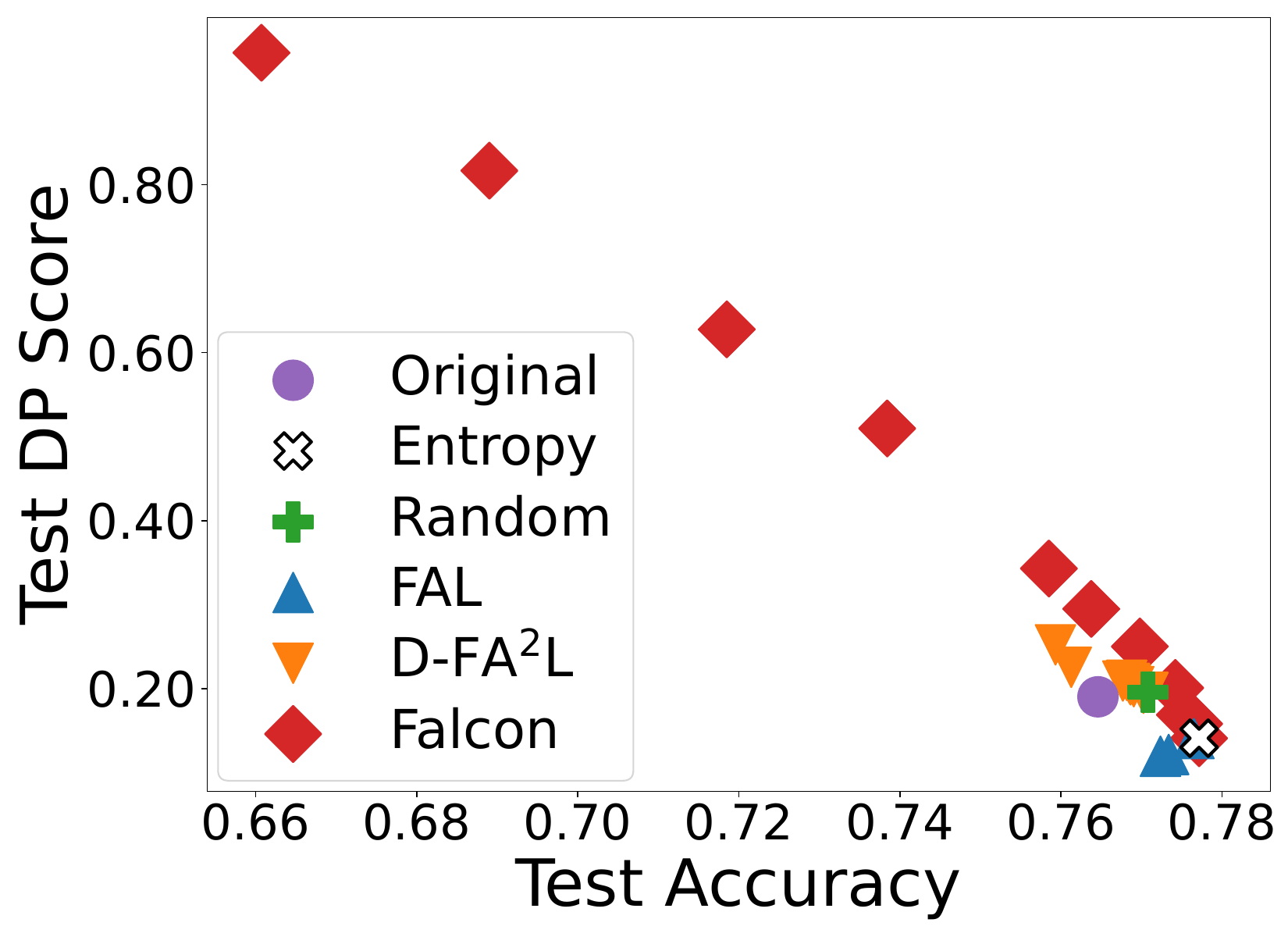}
     \caption{TravelTime (DP)}
     \label{fig:acsdp}
  \end{subfigure}
  \begin{subfigure}{0.495\columnwidth}
    \includegraphics[width=\columnwidth]{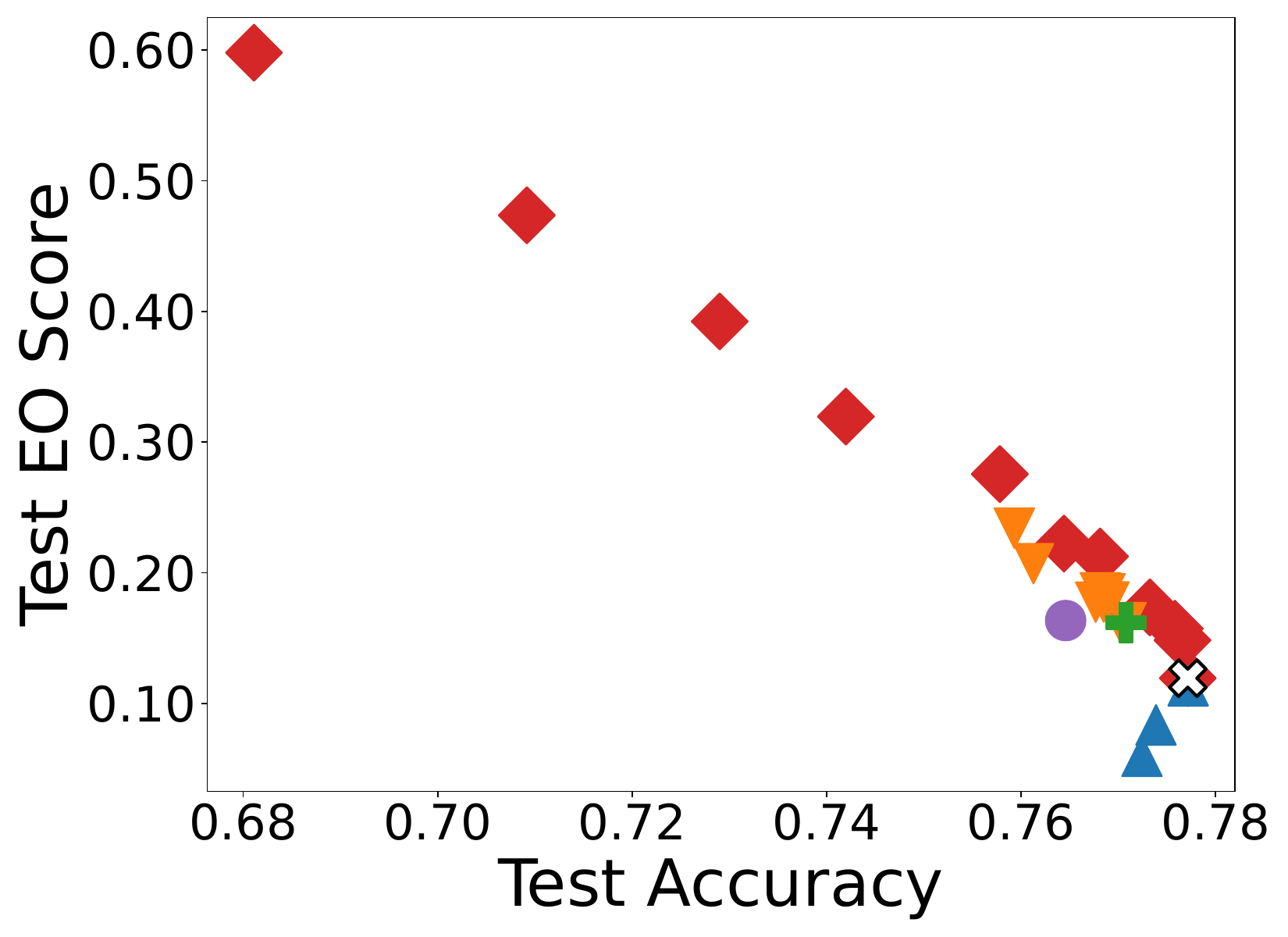}
     \caption{TravelTime (EO)}
     \label{fig:adultdp}
  \end{subfigure} 
  \begin{subfigure}{0.495\columnwidth}
    \includegraphics[width=\columnwidth]{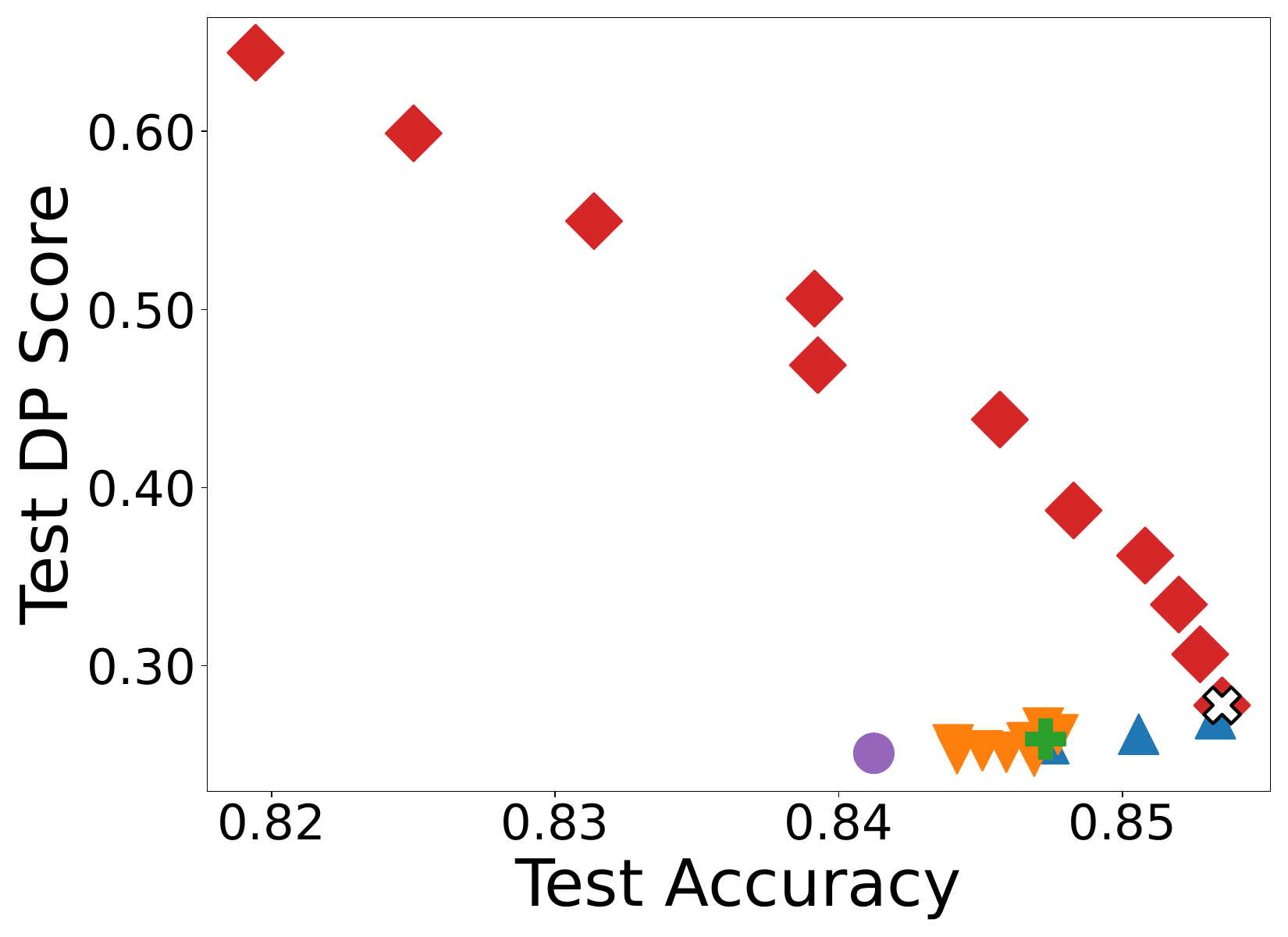}
     \caption{Employ (DP)}
     \label{fig:compasdp}
  \end{subfigure}
\begin{subfigure}{0.495\columnwidth}
    \includegraphics[width=\columnwidth]{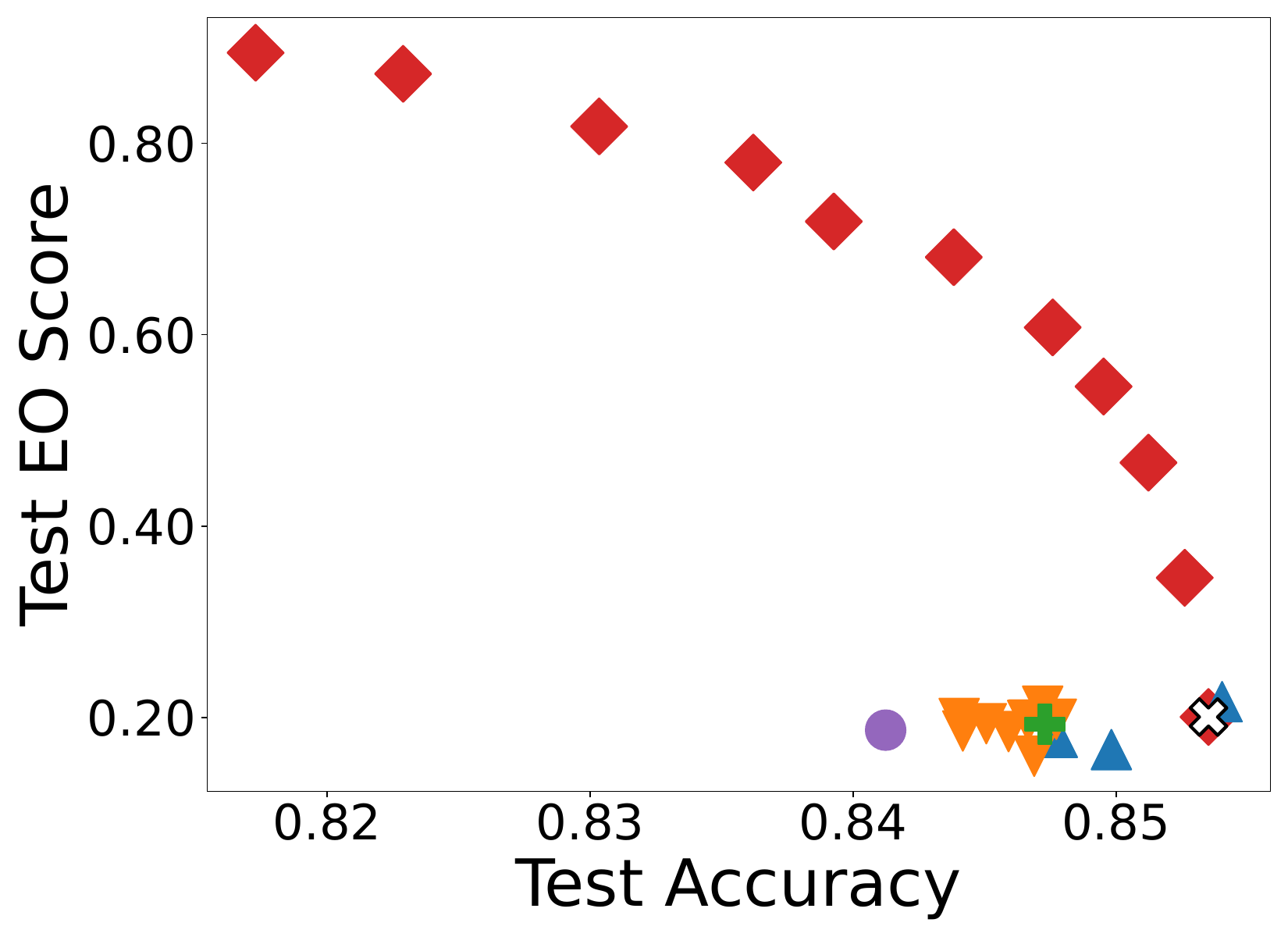}
     \caption{Employ (EO)}
     \label{fig:celebdp}
  \end{subfigure}
     \caption{Accuracy-fairness trade-offs of neural network on the two datasets (TravelTime and Employ) using the two fairness measures (DP and EO). In addition to the baselines, we add the result of model training without labeling any additional data and call it ``Original.'' As a result, only \method{} significantly improves fairness and shows clear accuracy and fairness trade-offs.
     }
 \label{fig:tradeoffsnn}
\end{figure}

\section{More Policy Sets}

\rev{We continue from Section~\ref{exp:numpolicies} and perform additional experiments to investigate the impact of policy sets on the \method{}'s performance. We first considered simpler policy sets with only two policies $[r=0.4, r=0.7]$, $[r=0.3, r=0.8]$ to check the impact of the quality of the policies. We also a policy set that contains extreme policies where we added $r=0.1, 0.2, 0.8, 0.9$ to our default set to check the impact of adding extreme policies. Table~\ref{tbl:morepolicysets} shows the fairness results for different policy sets using the TravelTime and Employ datasets. We first observe that $[r=0.4, r=0.7]$ is better than $[r=0.3, r=0.8]$ because the policies in the former set are closer to the optimal policies for each dataset, which are $r=0.6$ for TravelTime (Figure~\ref{fig:policycomparison}) and $r=0.5$ for Employ (Figure~\ref{fig:policycomparisonemploy}). However, our default policy set outperforms these alternatives in most cases, as it already includes the optimal policies. In addition, the default set performs better than the set with extreme policies in the last row. This is because extreme policies usually yield a worse trade-off between informativeness and postpone rate. Thus, our default policy set offers a balanced selection of diverse policies that are not too extreme.}


\begin{table}[t]
  \small
  \centering
  \begin{tabular}{lcccc}
    \toprule
     & \multicolumn{2}{c}{\bf TravelTime} & \multicolumn{2}{c}{\bf Employ}\\ 
    \cmidrule(lr){1-1}\cmidrule(lr){2-3} \cmidrule(lr){4-5} 
    {\bf Policy Set} & {\bf DP} & {\bf EO} & {\bf DP} & {\bf EO} \\
    \midrule
    $[r=0.4, r=0.7]$ & 0.948 & {\bf 0.622} & 0.644 & 0.899 \\
    $[r=0.3, r=0.8]$ & 0.815 & 0.385 & 0.626 & 0.890 \\
    \midrule
    $[r=0.3, \dots, r=0.7]$ (default) & {\bf 0.966} & 0.616 & {\bf 0.645} & {\bf 0.901} \\
    $[r=0.1, \dots, r=0.9]$ & 0.943 & 0.554 & 0.613 & 0.878 \\
    
    \bottomrule
  \end{tabular}
  \caption{Impact of different policy sets on \method{}.}
  \label{tbl:morepolicysets}
\end{table}

\section{Combining trial-and-error with Baselines for other datasets}
\rev{We continue from Section~\ref{exp:ablation} and perform the same experiments using the Income and COMPAS datasets. In Table~\ref{tbl:newbaselines-others}, the key trends are similar to Table~\ref{tbl:newbaselines-small} where (1) the fair AL baselines can be improved with trial-and-error, but (2) \method{} still outperforms them all, demonstrating the importance of the other \method{} components.}


\begin{table}[t]
\small
  \centering
  \begin{tabular}{cccc|c@{\hspace{2.5pt}}c|c}
    \toprule
    {\bf Datasets} & {\bf Fair.} & \multicolumn{5}{c}{\bf Fairness Score} \\
    \midrule
    & & {\bf FAL} & $\mathbf{FAL^+}$ & $\mathbf{D}$-$\mathbf{FA^{2}L}$ & $\mathbf{D}$-$\mathbf{FA^{2}L^+}$ & {\bf \method{}} \\
    \midrule
    \multirow{2}{*}{Income} & DP & 0.366 & 0.720 & 0.361 & 0.682 & {\bf 0.816}\\
    \cmidrule{2-7}
    & EO & 0.453 & 0.777 & 0.435 & 0.700 & {\bf 0.834}\\
    \midrule
    \multirow{2}{*}{COMPAS} & DP & 0.392 & 0.728 & 0.373 & 0.760 & {\bf 0.861}\\
    \cmidrule{2-7}
    & EO & 0.403 & 0.584 & 0.400 & 0.666 & {\bf 0.924}\\
    \bottomrule
  \end{tabular}
  \caption{Comparison of \method{} against fair AL baselines combined with trial-and-error on the Income and COMPAS datasets.}
  \label{tbl:newbaselines-others}
\end{table}

\section{Adversarial MABs on more datasets}
\rev{We continue from Section~\ref{exp:mabs} and show experimental results on the Income and COMPAS datasets in Table~\ref{tbl:moremabmethods-others}. The key observation is similar to Table~\ref{tbl:moremabmethods} where EXP3 exhibits comparable performance to other adversarial MABs with high probability regret bounds. But, we re-emphasize that \method{} can be compatible with any MAB capable of handling adversarial rewards.}

\begin{table}[t]
\small
  \centering
  \begin{tabular}{ccc|ccc}
    \toprule
    {\bf Datasets} & {\bf Fairness} & \multicolumn{4}{c}{\bf Fairness Score} \\
    \midrule
    & & {\bf Original} & $\mathbf{EXP3}$ & $\mathbf{EXP3}$-$\mathbf{IX}$ & $\mathbf{EXP4.P}$\\
    \midrule
    \multirow{2}{*}{Income} & DP & 0.355 & 0.816 & {\bf 0.820} & 0.801 \\
    \cmidrule{2-6}
    & EO & 0.402 & {\bf 0.834} & 0.829 & {\bf 0.834} \\
    \midrule
    \multirow{2}{*}{COMPAS} & DP & 0.365 & 0.861 & {\bf 0.883} & 0.856 \\
    \cmidrule{2-6}
    & EO & 0.372 & 0.924 & {\bf 0.929} & 0.924 \\
    \bottomrule
  \end{tabular}
  \caption{Fairness results on the Income and COMPAS datasets when using \method{} with other adversarial MABs.}
  \label{tbl:moremabmethods-others}
\end{table}



}{}

\end{document}